%% file: main.tex
\pdfoutput=1
\documentclass[letterpaper]{article} 
\usepackage[preprint]{aaai2027}  
\usepackage[hyphens]{url}  
\usepackage{graphicx} 
\usepackage{natbib}  
\usepackage{caption} 
\usepackage{subcaption}
\usepackage{algorithm}
\usepackage{algorithmic}
\usepackage{multirow}
\usepackage{amsmath}

\usepackage{newfloat}
\usepackage{listings}
\DeclareCaptionStyle{ruled}{labelfont=normalfont,labelsep=colon,strut=off} 
\floatstyle{ruled}
\newfloat{listing}{tb}{lst}{}
\floatname{listing}{Listing}

\usepackage{booktabs}
\usepackage[table]{xcolor}

\title{CUE-Mem: Benchmarking Long-Term User Memory via Implicit Cues \\ in Multimodal Conversations}

\author{
Yulin Hu\textsuperscript{1}, Yanyan Zhao\textsuperscript{1}\corresponding, Zimo Long\textsuperscript{1}, Xing Fu\textsuperscript{1}, Mengtong Ji\textsuperscript{1},\\
Weixiang Zhao\textsuperscript{1}, Yutai Hou\textsuperscript{2}, Qianchao Wang\textsuperscript{2}, Dandan Tu\textsuperscript{2}
}
\affiliations{
\textsuperscript{1}Harbin Institute of Technology\\
\textsuperscript{2}Huawei Technologies Co., Ltd.
}

\newsavebox{\cueexamplebox}
\newsavebox{\cuepromptbox}
\newenvironment{examplebox}{%
  \par\medskip\noindent
  \begin{lrbox}{\cuepromptbox}%
  \begin{minipage}{\dimexpr\linewidth-14pt\relax}\small
  \setlength{\parindent}{0pt}%
  \linespread{0.95}\selectfont
}{%
  \end{minipage}%
  \end{lrbox}%
  \setlength{\fboxsep}{6pt}%
  \setlength{\fboxrule}{0.4pt}%
  \fcolorbox{orange!30}{orange!6}{\usebox{\cuepromptbox}}%
  \par\medskip
}
\newenvironment{mybox}{%
  \par\medskip\noindent
  \begin{lrbox}{\cueexamplebox}%
  \begin{minipage}{0.94\textwidth}\small
}{%
  \end{minipage}%
  \end{lrbox}%
  \setlength{\fboxsep}{8pt}%
  \colorbox{gray!8}{\usebox{\cueexamplebox}}%
  \par\medskip
}
\newcommand{\cueoption}[3]{%
  \begin{minipage}[t]{0.235\linewidth}\centering
  \IfFileExists{#2}{%
    \includegraphics[width=\linewidth,height=2.65cm,keepaspectratio]{#2}%
  }{%
    \fbox{\parbox[c][2.45cm][c]{0.88\linewidth}{\centering\scriptsize
    Option image unavailable}}%
  }\\[0.45ex]
  \scriptsize\textbf{#1}\enspace #3
  \end{minipage}%
}
\newcommand{\cueclue}[3]{%
  \begin{minipage}[t]{0.31\linewidth}\centering
  \IfFileExists{#2}{%
    \includegraphics[width=\linewidth,height=3.1cm,keepaspectratio]{#2}%
  }{%
    \fbox{\parbox[c][2.9cm][c]{0.88\linewidth}{\centering\scriptsize
    Original memory image\\not in transferred snapshot}}%
  }\\[0.45ex]
  \scriptsize\textbf{#1}\enspace #3
  \end{minipage}%
}
\newcommand{\cuecluesmall}[3]{%
  \begin{minipage}[t]{0.19\linewidth}\centering
  \IfFileExists{#2}{%
    \includegraphics[width=\linewidth,height=2.45cm,keepaspectratio]{#2}%
  }{%
    \fbox{\parbox[c][2.25cm][c]{0.86\linewidth}{\centering\scriptsize
    Memory image unavailable}}%
  }\\[0.45ex]
  \scriptsize\textbf{#1}\\[-0.2ex]#3
  \end{minipage}%
}

\newcommand{\projectlink}{\url{https://github.com/yulinlp/CUE-MEM}}

\begin{document}

\maketitle

\begin{abstract}
Long-term memory is essential for multimodal agents that interact with users across sustained conversations. However, user memories are not always explicitly stated: they may also be implied by recurring background objects in images, ambient sounds in audio, or other peripheral multimodal cues. Existing benchmarks largely focus on text-only memory or explicit multimodal evidence, leaving implicit multimodal cues underexplored. We introduce CUE-Mem, a text-image-audio benchmark for evaluating long-term user memory from implicit cues. CUE-Mem contains 2,674 questions across explicit and implicit evidence settings and covers four tasks: Entity Recall, Long Pattern, Personalized Recommendation, and Answer Refusal. Across textualized memory systems, implicit performance remains far below oracle evidence, locating the main bottleneck in preserving and retrieving subtle cues rather than question answerability. Increasing caption detail recovers more of this evidence, but brings uneven gains and rapidly growing token costs, motivating native multimodal access. Yet native access does not uniformly resolve the bottleneck: evidence use depends strongly on the backbone, while multimodal indexing introduces substantial retrieval noise. CUE-Mem provides a testbed for memory systems that selectively retain, retrieve, and use subtle multimodal evidence.\footnote{Open-source repository: \projectlink.}
\end{abstract}


\section{Introduction}

Recent advances in multimodal large language models (MLLMs) have enabled conversational agents to interact with users through text, images, and audio. As these agents evolve into long-term personalized assistants, accumulated conversation histories provide a natural basis for building user memory. Rather than treating each query in isolation, an assistant should gradually develop knowledge of the user's preferences, important entities, habits, living environment, and personal context across sustained interactions. Accordingly, long-term memory systems have explored how to store, retrieve, and update such information over time~\citep{zhong2024memorybank,packer2023memgpt,wang2023augmenting}.

In real-world interactions, user memory is not always built from active disclosure. Shared images and voice messages convey not only the main content of an exchange but also incidental context, such as pet supplies repeatedly appearing at the edge of home selfies or cat meows in the background of voice messages. The notion of \textit{behavioral residue} suggests that people's actions, habits, and living environments leave observable traces in everyday objects and physical spaces~\citep{gosling2002room,gosling2008snoop}. A single trace may be too weak to support a reliable memory, but recurring traces can become meaningful evidence about a user's interests, habits, household, or living environment. We define such signals as \textit{implicit multimodal cues}: recurring visual or acoustic evidence that remains peripheral or in the background of the current exchange, rather than being directly stated or foregrounded. In contrast, \textit{explicit multimodal cues} directly state or foreground user information.

\begin{figure}[t]
\centering
\includegraphics[width=1.0\columnwidth]{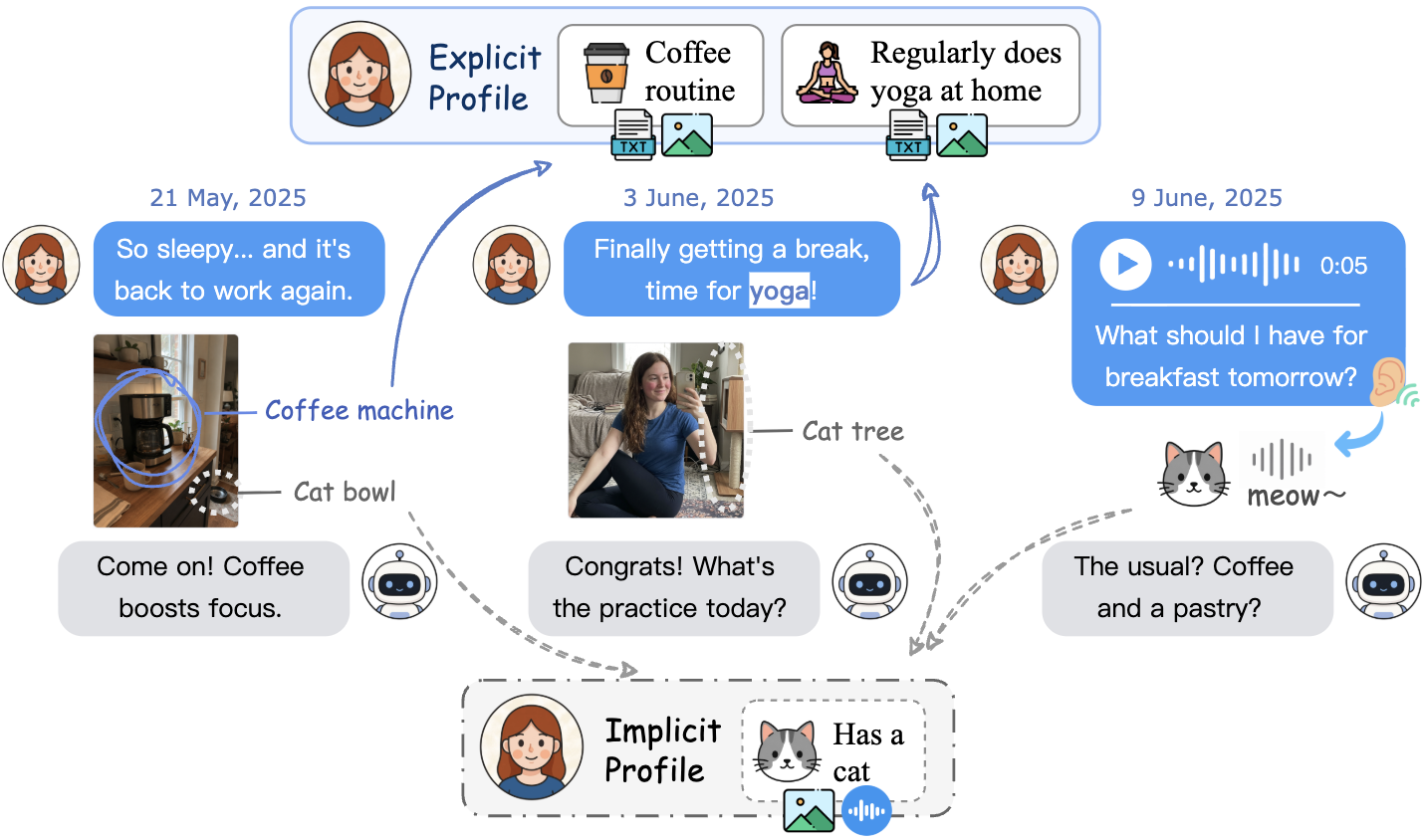}
\caption{Explicit and implicit evidence settings in \textsc{CUE-Mem}. Explicit memories are supported by directly stated or foregrounded information, whereas implicit memories rely on recurring background cues across images and audio. Profile boxes denote target memories.}
\label{fig:intro}
\end{figure}

Despite progress in multimodal memory benchmarks~\citep{bei2026memgallery,ren2026memlens,chai2026smmbench,huang2026m3examMemory}, evaluations still emphasize evidence that is directly stated or foregrounded in the current interaction. Some benchmarks cover multiple modalities, but audio, particularly ambient sound, remains underrepresented. More importantly, few evaluate cues that remain peripheral across interactions and become informative only through recurrence. It remains unclear whether long-term memory systems can form user memories from recurring background evidence across multimodal histories.

To address this gap, we introduce \textsc{CUE-Mem}, a text-image-audio benchmark for evaluating long-term user memory from implicit multimodal cues. It comprises synthetic multi-session conversation histories organized into explicit and implicit evidence conditions. As illustrated in Figure~\ref{fig:intro}, the explicit condition supports user attributes through direct statements or foregrounded multimodal content, whereas the implicit condition embeds \textit{\textbf{c}ontextual \textbf{u}nstated \textbf{e}vidence} in recurring background visual or acoustic cues. The two conditions enable aggregate comparison, but individual examples are not constructed as one-to-one counterparts.

For systematic evaluation, \textsc{CUE-Mem} organizes long-term user memory into four complementary tasks. \textbf{Entity Recall} evaluates whether models can recover user-related entities and attributes from long histories. \textbf{Long Pattern} tests whether models can aggregate recurring evidence into stable memories about preferences, habits, routines, or environments. \textbf{Personalized Recommendation} evaluates whether memories can support personalized assistance. \textbf{Answer Refusal} examines whether models can avoid unsupported memory claims when evidence is absent or insufficient. Each task contains explicit and implicit variants, enabling aggregate comparison between memory supported by direct evidence and memory inferred from weak multimodal cues.

We evaluate representative memory systems under textualized and native multimodal paradigms. Across all evaluated systems, directly stated or foregrounded evidence is consistently easier to use than recurring implicit cues, producing a substantial explicit--implicit gap. Cue-aware textualization narrows this gap, but yields uneven task gains and growing input costs. Native multimodal evidence use can recover lost information, but its benefit depends strongly on the backbone, while multimodal indexing introduces substantial retrieval noise; no setting closes the gap.

Our contributions are threefold:

(1) We introduce \textsc{CUE-Mem}, a text-image-audio benchmark for evaluating long-term user memory from recurring implicit multimodal cues.

(2) We construct explicit and implicit evidence conditions and define four tasks that probe memory recovery, aggregation, application, and calibration.

(3) We benchmark textualized and native multimodal memory systems, revealing a persistent explicit--implicit gap, a preservation--efficiency trade-off in textualization, and substantial retrieval noise in multimodal indexing.

\section{Related Work}

\subsection{Implicit User Memory Benchmarks}

Recent benchmarks study several forms of unstated user information. PrefEval and PersonaMem~\citep{zhao2025prefeval,jiang2025personamem,jiang2025personamemv2} infer preferences or personas from utterances, while DynamicMem~\citep{xie2026dynamicmem} reconstructs evolving profiles from distributed behavioral traces across applications. ImplicitMemBench~\citep{qin2026implicitmembench} instead uses ``implicit memory'' in the cognitive sense of procedural learning and priming. M$^3$Exam~\citep{huang2026m3examMemory} targets unstated roles, relationships, and motivations, whereas \textsc{CUE-Mem} focuses on recurring visual and acoustic cues peripheral to the current exchange.

\subsection{Multimodal Memory Benchmarks}

Long-term multimodal memory benchmarks extend image-text dialogue understanding~\citep{liu2024mmdu,xue2025mmrc} to persistent histories. Mem-Gallery and MemLens~\citep{bei2026memgallery,ren2026memlens} evaluate multi-session image-text memory, Persona-MME~\citep{nie2026personavlm} emphasizes long-term multimodal personalization, and H2HMem~\citep{zhu2026h2hmem} studies dyadic and multi-party interactions. SMMBench~\citep{chai2026smmbench} distributes evidence across heterogeneous sources, whereas MOSAIC~\citep{anonymous2026mosaic} requires multi-hop reasoning across text, image, audio, and video. These benchmarks primarily test foregrounded facts, distributed evidence, or cross-modal chains. As summarized in Table~\ref{tab:benchmark_comparison}, \textsc{CUE-Mem} instead isolates user memories supported by recurring background visual and acoustic cues.

\begin{table}[t]
\centering
\small
\setlength{\tabcolsep}{3.5pt}
\begin{tabular}{lccccc}
\toprule
\textbf{Benchmark} & \textbf{Mod.} & \textbf{A.Turn} & \textbf{A.Sess.} & \textbf{A.MM} & \textbf{Imp.} \\
\midrule
DuLeMon & T & 8.16 & -- & -- & No \\
DialogBench & T & 7.48 & -- & -- & No \\
MemoryBank & T & 3.77 & 10 & -- & No \\
LOCCO & T & 4.77 & 10 & -- & No \\
LongMemEval & T & 5.19 & 35 & -- & No \\
\midrule
LoCoMo & T+I & 10.81 & 13.6 & 3.35 & No \\
Mem-Gallery & T+I & 16.51 & 12 & 74.5 & No \\
MemLens & T+I & $\sim$10 & 44.46 & 45.12 & No \\
SMMBench & T+I & 831 & 1 & 17.81 & No \\
M$^3$Exam & T+I & 12.7 & 15.93 & 7.5 & Sem. \\
Persona-MME & T+I & 142.9 & 1 & 22.7 & No \\
H2HMem & T+I & 22.9 & 12.4 & 52.0 & No \\
MOSAIC & T+I+A & -- & 17.5 & 12.7 & No \\
\midrule
\textbf{\textsc{CUE-Mem}} & \textbf{T+I+A} & \textbf{8.95} & \textbf{32.4} & \textbf{136.2} & \textbf{Yes} \\
\bottomrule
\end{tabular}
\caption{Comparison with representative conversational and multimodal memory benchmarks. T, I, and A denote text, image, and audio. A.Turn denotes turns per session; when session boundaries are unavailable, we use the reported dialogue or conversational-source unit. A.Sess. and A.MM denote sessions and multimodal items per history. MemLens statistics are averaged over its four context-length settings. Imp. distinguishes implicit multimodal cues from semantic (Sem.) inference.}
\label{tab:benchmark_comparison}
\end{table}

\begin{figure*}[t]
\centering
\includegraphics[width=0.98\textwidth]{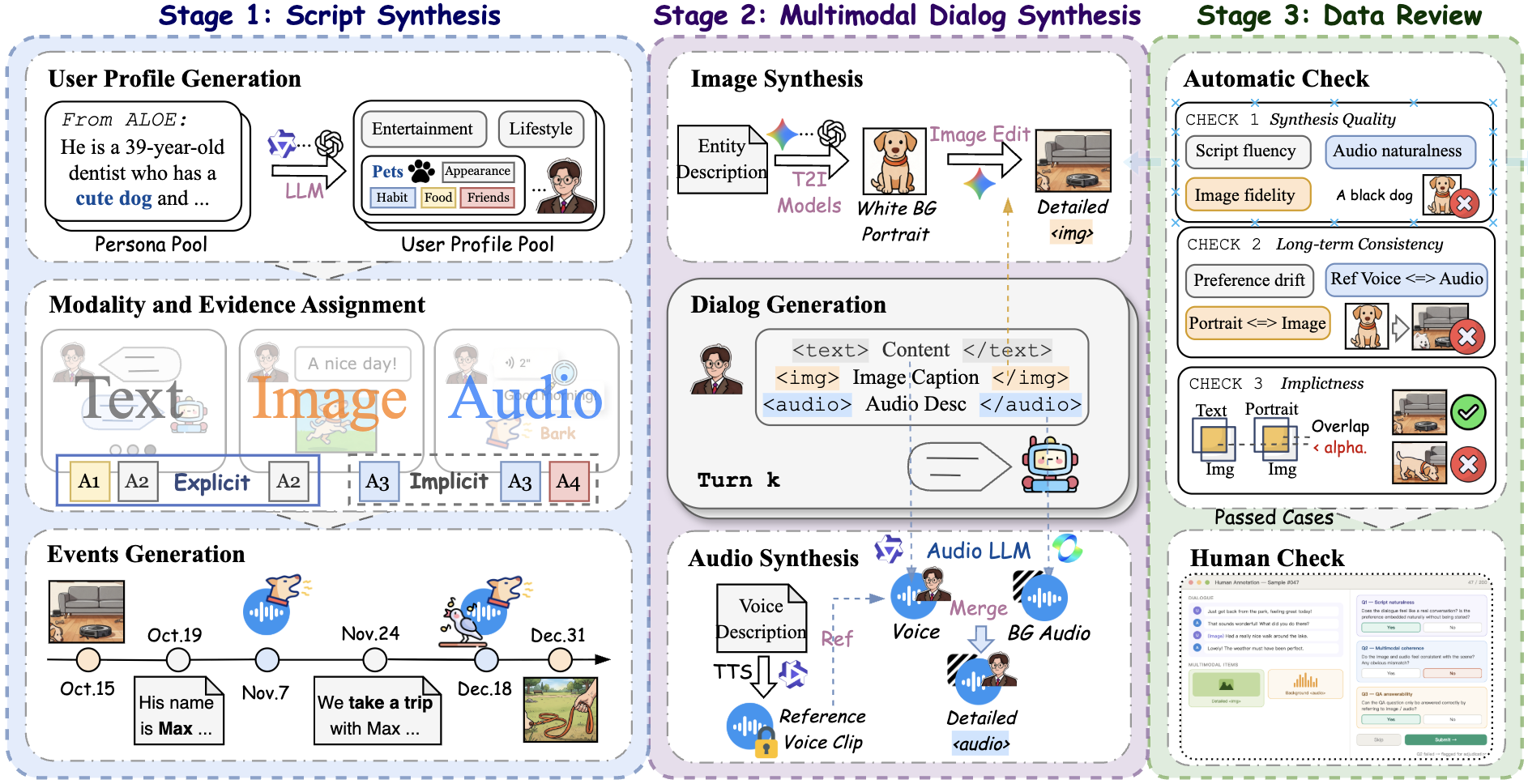}
\caption{Construction pipeline of \textsc{CUE-Mem}. Starting from predefined user profiles, we derive explicit and implicit evidence plans, distribute them across multi-session scripts, and realize them as text-image-audio conversations. Automatic and human checks are applied to ensure synthesis quality, long-term consistency, and the validity of implicit cues.}
\label{fig:pipeline}
\end{figure*}

\section{Dataset Construction}
\label{sec:dataset}

\subsection{Overview}

\textsc{CUE-Mem} consists of synthetic multi-session text-image-audio conversations grounded in structured user profiles. Profile attributes are assigned modalities and evidence types, then instantiated as events, dialogues, images, and audio. As shown in Figure~\ref{fig:pipeline}, \textit{Script Synthesis}, \textit{Multimodal Dialog Synthesis}, and \textit{Data Review} yield controlled histories containing direct disclosures, foregrounded multimodal content, and recurring background visual and acoustic cues.

\begin{figure*}[ht]
\centering
\includegraphics[width=0.98\textwidth]{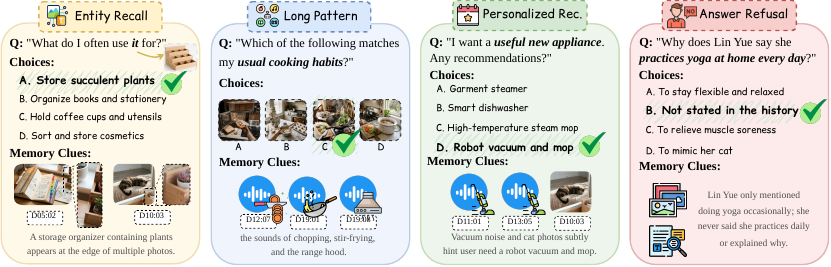}
\caption{
Task examples from \textsc{CUE-Mem}, covering Entity Recall, Long Pattern, Personalized Recommendation, and Answer Refusal. The first three illustrate memory evaluation from implicit multimodal cues, while Answer Refusal tests abstention under insufficient evidence.
}
\label{fig:task_example}
\end{figure*}

\subsection{Stage 1: Script Synthesis}
\label{sec:script_synthesis}

\paragraph{User Profile Generation.}
We initialize each synthetic user with a persona seed from ALOE~\citep{wu-etal-2025-aligning} or PersonaHub~\citep{ge2024scaling}. An LLM expands the seed into a profile covering memory-relevant categories adapted from PrefEval~\citep{zhao2025prefeval}, including entertainment, lifestyle, pets, appearance, habits, food, and social relationships. The profile attributes serve as latent memory ground truth for subsequent conversation and task construction.

\paragraph{Modality and Evidence Assignment.}
Given a structured user profile, LLM agents assign each attribute a modality and evidence type, together with a brief realization plan for later events. To support implicit audio cues, we construct an ambient-sound seed pool from ESC-50~\citep{piczak2015esc} and FSD50K~\citep{fonseca2021fsd50k}. LLM agents first identify 114 candidate categories that can naturally serve as background cues, after which manual review retains 96 usable categories. The resulting pool guides audio-modality assignment and subsequent cue synthesis.

\paragraph{Events Generation.}
We generate a chronological event timeline for each user, with each event specifying its scenario, date, profile attributes, and modalities. Later events are generated conditioned on preceding ones to reduce repetition and maintain continuity. Distributing events across sessions allows implicit cues to recur and tests whether models can accumulate weak multimodal evidence over long histories.

\subsection{Stage 2: Multimodal Dialog Synthesis}
\label{sec:multimodal_dialog_synthesis}

\paragraph{Image Synthesis.}
For each user, we identify recurring entities, including people, pets, objects, and locations, and create profile-grounded descriptions and reusable assets to maintain cross-session consistency. For each image-bearing turn, we synthesize or edit the image according to the event context and evidence type. Explicit visual evidence appears as the foreground subject, whereas implicit cues are placed among distractors in cluttered backgrounds, remaining observable but peripheral and undiscussed in the dialogue.

\paragraph{Audio Synthesis.}
For audio, we construct a reusable voice profile and anchor clip for each user to maintain speaker consistency across sessions. The user's spoken message in each audio-bearing turn is generated conditioned on this anchor. When an acoustic cue is required, we synthesize or retrieve a background track from the audio seed pool and mix it with the speech. In explicit cases, the spoken content directly states the target information; in implicit cases, the ambient cue remains audible but unstated, providing memory evidence beyond the transcript.

\paragraph{Dialog Generation.}
After the image and audio items are prepared, the dialogue generator integrates the event timeline, user profile, modality assignments, and multimodal item descriptions to produce multi-session user--assistant exchanges with text, image, and audio turns at planned positions. The generation process preserves the assigned evidence setting: explicit evidence is verbalized or foregrounded, whereas implicit cues remain in the associated media and are not mentioned in the dialogue text.

\subsection{Stage 3: Data Review}

\paragraph{Automatic Check.}
Before human review, we apply three automatic checks. \textit{CHECK 1: Synthesis Quality} filters unnatural scripts, low-fidelity images, corrupted audio, and cross-modal mismatches. \textit{CHECK 2: Long-term Consistency} verifies cross-session entity consistency by comparing visual entities against reference assets with SigLIP 2~\citep{tschannen2025siglip2} and voices against anchors with ERes2NetV2~\citep{chen2024eres2netv2}. \textit{CHECK 3: Implicitness} requires both text--image overlap and implicit-target--full-image overlap to remain below predefined thresholds, preventing textual leakage and overly salient targets. For implicit audio, the ambient cue must be audible but absent from the speech transcript. Together, these checks remove \textbf{42.80\%} of the initial samples.

\paragraph{Human Check.}
After automatic filtering, \textbf{12} annotators conduct two review rounds totaling approximately \textbf{550} annotator-hours. Three annotators independently assess each sample for \textit{Dialogue naturalness}, \textit{Multimodal coherence}, and \textit{Evidence validity}. Unanimously approved samples are retained; disagreements undergo expert adjudication. The average Fleiss' $\kappa$ across the three dimensions is \textbf{0.73}. Human review removes an additional \textbf{22.10\%} of the remaining samples.

\begin{table*}[t]
\centering
\small
\setlength{\tabcolsep}{2.2pt}
\begin{tabular}{lccccccccccc}
\toprule
\multirow{2.5}{*}{\textbf{Method}}
& \multicolumn{2}{c}{\textbf{Entity Recall}}
& \multicolumn{2}{c}{\textbf{Long Pattern}}
& \multicolumn{2}{c}{\textbf{Personalized Rec.}}
& \multicolumn{2}{c}{\textbf{Answer Refusal}}
& \multicolumn{3}{c}{\textbf{Memory Avg.}} \\
\cmidrule(lr){2-3}
\cmidrule(lr){4-5}
\cmidrule(lr){6-7}
\cmidrule(lr){8-9}
\cmidrule(lr){10-12}
& Explicit & Implicit
& Explicit & Implicit
& Explicit & Implicit
& Explicit & Implicit
& Explicit & Implicit & EI-Gap \\
\midrule

\rowcolor{gray!6}
Human (Full)
& 90.2 & 84.7
& 84.6 & 82.1
& 74.9 & 81.5
& -- & --
& 83.2 & 82.8 & 0.4 \\

\rowcolor{gray!6}
Human (Oracle)
& 93.6 & 88.7
& 92.4 & 94.1
& 86.3 & 76.8
& -- & --
& 90.8 & 86.5 & 4.3 \\

\midrule
\rowcolor{gray!10}
\multicolumn{12}{c}{\textit{Qwen3.6-35B-A3B}} \\
\midrule

\rowcolor{gray!6}
No Context
& 31.7 & 23.8
& 27.7 & 22.6
& 30.5 & 30.8
& 74.9 & 70.9
& 30.0 & 25.7 & 4.2 \\

\rowcolor{gray!6}
Oracle Evidence$^\star$
& 87.3 & 93.8
& 85.0 & 81.2
& 82.6 & 67.3
& -- & --
& 85.0 & 80.8 & 4.2 \\

\midrule

Full Memory
& 80.1 & \textbf{27.0}
& 80.4 & 38.9
& 70.1 & 49.0
& 48.4 & 95.8
& 76.9 & 38.3 & 38.6 \\

NaiveRAG
& 69.9 & 15.6
& 79.1 & 34.1
& 67.5 & \textbf{51.4}
& 62.8 & 97.8
& 72.2 & 33.7 & 38.5 \\

Generative Agents
& 31.2 & 8.7
& 56.2 & 28.4
& 62.3 & \textbf{51.4}
& 77.0 & 98.7
& 49.9 & 29.5 & 20.4 \\

Reflexion
& 80.5 & 26.0
& 81.0 & 38.9
& 68.8 & 50.0
& 49.2 & 96.0
& 76.8 & 38.3 & 38.5 \\

MemGPT
& 80.8 & 26.5
& 79.7 & \textbf{40.9}
& 70.1 & 48.1
& 47.8 & 95.4
& 76.9 & \textbf{38.5} & 38.4 \\

A-Mem
& 64.6 & 14.9
& 75.8 & 31.2
& 62.3 & 50.5
& 66.1 & 96.9
& 67.6 & 32.2 & 35.4 \\

MemoryOS
& 11.5 & 2.7
& 31.4 & 15.4
& 53.9 & 48.1
& 91.0 & \textbf{99.6}
& 32.3 & 22.1 & 10.2 \\

\midrule
\rowcolor{gray!10}
\multicolumn{12}{c}{\textit{GPT-5.4-mini}} \\
\midrule

\rowcolor{gray!6}
No Context
& 32.4 & 27.6
& 29.6 & 25.3
& 26.0 & 30.3
& 51.6 & 43.3
& 29.3 & 27.7 & 1.6 \\

\rowcolor{gray!6}
Oracle Evidence$^\star$
& 87.8 & 93.1
& 82.4 & 76.0
& 80.1 & 66.2
& -- & --
& 83.4 & 78.4 & 5.0 \\

\midrule

Full Memory
& 75.7 & 28.5
& 74.5 & 41.3
& 66.9 & 42.8
& 16.7 & 71.5
& 72.4 & 37.5 & 34.8 \\

NaiveRAG
& 71.7 & 35.9
& 69.9 & 38.0
& 69.5 & 41.3
& 29.0 & 79.5
& 70.4 & 38.4 & 32.0 \\

Generative Agents
& 46.8 & 31.4
& 63.4 & 37.5
& 59.1 & 42.8
& 45.6 & 86.1
& 56.4 & 37.2 & 19.2 \\

Reflexion
& 75.3 & 33.9
& 80.4 & \textbf{46.2}
& 68.8 & \textbf{45.7}
& 18.3 & 71.3
& 74.8 & \textbf{41.9} & 32.8 \\

MemGPT
& 73.3 & 31.2
& 75.8 & 40.4
& 64.3 & 42.8
& 19.4 & 70.6
& 71.1 & 38.1 & 33.0 \\

A-Mem
& 78.6 & \textbf{42.1}
& 76.5 & 38.0
& 67.5 & 44.7
& 39.6 & 83.0
& 74.2 & 41.6 & 32.6 \\

MemoryOS
& 44.1 & 39.1
& 34.6 & 25.0
& 26.0 & 26.9
& 97.8 & \textbf{99.6}
& 34.9 & 30.3 & 4.5 \\

\bottomrule
\end{tabular}
\caption{
Main results of textualized memory systems on CUE-Mem.
Human (Full) reads the full textualized history, whereas Human (Oracle) receives only oracle text.
Oracle Evidence$^\star$ denotes the source text prompts used to synthesize multimodal clues, which provide clean and accurate textual statements of the target information.
Memory Avg. averages Entity Recall, Long Pattern, and Personalized Recommendation.
Best non-oracle implicit accuracies within each backbone are bold.
}
\label{tab:rq1_textualized}
\end{table*}

\section{\textsc{CUE-Mem}}
\label{sec:cue_mem}

\subsection{Task Design}

\textsc{CUE-Mem} evaluates long-term user memory through four multiple-choice tasks: Entity Recall, Long Pattern, Personalized Recommendation, and Answer Refusal, with examples in Figure~\ref{fig:task_example} and Appendix~B.2. Each question has four options with one correct answer. Together, these tasks test whether models can recover user-specific facts, consolidate recurring evidence, use memories for personalized assistance, and avoid unsupported memory claims. Each task includes explicit and implicit instances.

\paragraph{Entity Recall.}
Entity Recall tests whether models can recover concrete user-specific entities or attributes from long-term histories. Questions target facts about pets, personal objects, and the user's living environment.

\paragraph{Long Pattern.}
Long Pattern tests aggregation of recurring evidence into stable user memories. Unlike Entity Recall, which targets specific facts, it integrates evidence across sessions to infer higher-level preferences, habits, routines, or environmental patterns not evident in any session alone.

\paragraph{Personalized Recommendation.}
Personalized Recommendation tests whether models apply relevant user memories to select tailored recommendations. Distractors are plausible in general but insufficiently personalized, inconsistent with the user's memory, or unsupported by the history.

\paragraph{Answer Refusal.}
Answer Refusal tests whether models remain calibrated and avoid over-inference when memory evidence is absent or insufficient. The correct option is a refusal indicating that the history lacks sufficient evidence to answer.

\subsection{Construction Pipeline}

Task construction follows four steps. \textbf{(1) Target selection and QA generation.} We select memories instantiated in the history under each evidence setting. LLM agents generate one correct answer and three distractors, optionally using image or audio content, and shuffle the options. \textbf{(2) Clue annotation.} Agents identify turn- or item-level multimodal evidence in sessions. We remove invalid identifiers and verify support. \textbf{(3) Quality review.} Automatic and human checks examine answer support, distractor quality, ambiguity, text leakage, insufficient evidence, and label consistency. \textbf{(4) Question-only filtering.} The two evaluated backbones answer each question without its history in three runs; an instance is discarded only if all six responses are correct.

\begin{figure}[t]
\centering
\begin{minipage}[t]{0.60\columnwidth}
\vspace{0pt}
\centering
\includegraphics[width=\linewidth]{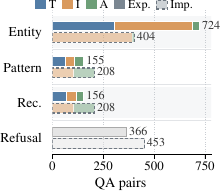}
\end{minipage}\hspace{0.05\columnwidth}%
\begin{minipage}[t]{0.349\columnwidth}
\vspace{0pt}
\raggedright
\small
\setlength{\tabcolsep}{1.5pt}
\renewcommand{\arraystretch}{0.98}
\begin{tabular*}{\linewidth}{@{\extracolsep{\fill}}lr@{}}
\toprule
\textbf{Statistic} & \textbf{Value} \\
\midrule
Users & 20 \\
Sessions & 648 \\
Turns & 5,798 \\
Images & 496 \\
Audio clips & 2,228 \\
QA pairs & 2,674 \\
Avg.\ sess./user & 32.4 \\
Avg.\ turns/sess. & 8.95 \\
Clues/Q (Exp.) & 3.14 \\
Clues/Q (Imp.) & 3.34 \\
\bottomrule
\end{tabular*}
\end{minipage}
\caption{Overview of \textsc{CUE-Mem}. Left: QA counts and clue-modality proportions by task and evidence setting (T: text, I: image, A: audio). Right: corpus-scale statistics and average clues per question. Answer Refusal has no supporting clues.}
\label{fig:cue_mem_stats}
\end{figure}

\subsection{Data Statistics}
At the corpus level, \textsc{CUE-Mem} contains 20 synthetic users, 648 sessions, and 5,798 dialogue turns, including 496 images and 2,228 audio clips. Each user has 32.4 sessions on average, with 8.95 turns per session. The benchmark contains 2,674 QA pairs: 1,401 explicit and 1,273 implicit. Entity Recall contributes 1,128 QAs, followed by Answer Refusal (819), Personalized Recommendation (364), and Long Pattern (363). Excluding Answer Refusal, explicit and implicit questions contain 3.14 and 3.34 supporting clues on average, respectively. Figure~\ref{fig:cue_mem_stats} summarizes these task-level counts and clue-modality distributions. The dataset, annotations, and evaluation code will be released upon publication.

\section{Experiments}

We conduct experiments on CUE-Mem to examine whether long-term memory systems can reliably form and use user memories from implicit multimodal cues. Our analysis is organized around three research questions:

\begin{itemize}
    \item \textbf{RQ1:} How well do textualized memory systems handle implicit multimodal cues?
    \item \textbf{RQ2:} How does textualization quality affect implicit memory performance?
    \item \textbf{RQ3:} Can native multimodal access alleviate the textualization bottleneck?
\end{itemize}

\subsection{Experimental Setup}

\paragraph{Evaluation protocol.}
We report accuracy under explicit and implicit evidence settings and define EI-Gap as $\mathrm{Acc}_{\mathrm{explicit}}-\mathrm{Acc}_{\mathrm{implicit}}$, where larger values indicate greater implicit degradation. No Context omits the history, whereas Oracle Evidence supplies the source text prompts used to synthesize multimodal clues. \textit{Memory Avg.} averages Entity Recall, Long Pattern, and Personalized Recommendation; Answer Refusal measures calibration under insufficient evidence and is reported separately. Statistical analyses appear in Appendix~D.1.

\paragraph{Memory paradigms and systems.}
We evaluate textualized memory, which captions images and audio before text-based processing, and native multimodal memory. Default textualization uses medium-grained image captions from GPT-5.4~\citep{singh2025openai} and hint-based audio captions from Gemini 3.1 Pro~\citep{googledeepmind2026gemini31pro}. Systems include Full Memory, NaiveRAG~\citep{lewis2020retrieval}, Generative Agents~\citep{park2023generative}, Reflexion~\citep{shinn2023reflexion}, MemGPT~\citep{packer2023memgpt}, A-Mem~\citep{xu2025mem}, and MemoryOS~\citep{kang2025memoryos}. Native multimodal RAG controls multimodal access at indexing and answer time; RQ2 compares textualization schemes.

\paragraph{Backbone models.}
We use Qwen3.6-35B-A3B~\citep{yang2025qwen3} and GPT-5.4-mini~\citep{singh2025openai} for textualized memory, and Qwen3-Omni-30B-A3B-Instruct~\citep{xu2025qwen3omini} and MiniCPM-o-4.5~\citep{cui2026minicpmo45} for native multimodal memory (Appendix~C).

\subsection{RQ1: How Well Do Textualized Memory Systems Handle Implicit Multimodal Cues?}

Table~\ref{tab:rq1_textualized} summarizes the main results of textualized memory systems on \textsc{CUE-Mem}. We highlight three findings.

\textbf{\textit{Textualized systems exhibit a substantial and persistent degradation on implicit memory.}}
Averaged across systems, Qwen3.6-35B-A3B and GPT-5.4-mini show EI-Gaps of 42.5 and 31.9 points on Entity Recall, 36.5 and 29.8 points on Long Pattern, and 15.2 and 19.3 points on Personalized Recommendation, respectively. The larger gaps on Entity Recall and Long Pattern indicate impaired preservation of peripheral entities and consolidation of recurring cues; Appendix~D.4 analyzes the effect of cue recurrence.

\textbf{\textit{Even the strongest memory systems remain far below Oracle Evidence.}}
The best implicit Memory Avg. reaches only 38.5 with Qwen3.6-35B-A3B and 41.9 with GPT-5.4-mini, compared with 80.8 and 78.4 for Oracle Evidence. Humans reading full textualized histories achieve 83.2 and 82.8 on explicit and implicit samples, suggesting that intrinsic ambiguity is not the primary source of this gap. This suggests that the main bottleneck lies in preserving and retrieving dispersed implicit evidence from long textualized histories, not in answering questions once evidence is available.

\textbf{\textit{Answer Refusal shows the opposite trend.}}
Both settings use misleading adversarial choices to induce unsupported memory use. Yet mean refusal accuracy rises from 63.2 to 97.2 for Qwen3.6-35B-A3B and from 38.1 to 80.2 for GPT-5.4-mini as evidence shifts from explicit to implicit. Systems retrieve and misapply explicit memories but often miss implicit cues and default to refusal; higher implicit accuracy thus reflects inaccessible memory, not better calibration.

\subsection{RQ2: How Does Cue-Aware Textualization Affect Implicit Memory?}

\begin{figure}[t]
\centering
\includegraphics[width=\linewidth]{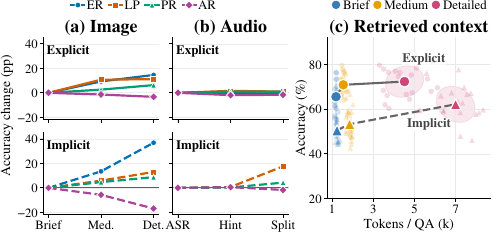}
\caption{
Effects and costs of cue-aware textualization.
(a--b) Accuracy changes from Brief captions and ASR, averaged across memory systems; solid/dashed lines denote explicit/implicit results.
(c) NaiveRAG accuracy versus retrieved tokens per QA; faint/large markers denote users/means, and ellipses show user variation.
ER, LP, PR, and AR denote the tasks.
}
\label{fig:rq2_caption_quality}
\label{fig:rq2_token_tradeoff}
\end{figure}

\begin{table*}[!t]
\centering
\small
\setlength{\tabcolsep}{3.0pt}
\begin{tabular}{llccccccccccc}
\toprule
\multirow{2.5}{*}{\textbf{Index}}
& \multirow{2.5}{*}{\textbf{Use}}
& \multicolumn{2}{c}{\textbf{Entity Recall}}
& \multicolumn{2}{c}{\textbf{Long Pattern}}
& \multicolumn{2}{c}{\textbf{Personalized Rec.}}
& \multicolumn{2}{c}{\textbf{Answer Refusal}}
& \multicolumn{3}{c}{\textbf{Memory Avg.}} \\
\cmidrule(lr){3-4}
\cmidrule(lr){5-6}
\cmidrule(lr){7-8}
\cmidrule(lr){9-10}
\cmidrule(lr){11-13}
& & Explicit & Implicit
& Explicit & Implicit
& Explicit & Implicit
& Explicit & Implicit
& Explicit & Implicit & EI-Gap \\
\midrule

\rowcolor{gray!6}
\multicolumn{2}{l}{Human (Full)}
& 82.4 & 45.1
& 88.1 & 60.3
& 70.2 & 57.1
& -- & --
& 80.2 & 54.2 & 26.0 \\

\rowcolor{gray!6}
\multicolumn{2}{l}{Human (Oracle)$^\star$}
& 94.2 & 86.1
& 97.1 & 95.3
& 91.4 & 82.2
& -- & --
& 94.2 & 87.9 & 6.3 \\

\midrule
\rowcolor{gray!10}
\multicolumn{13}{c}{\textit{Qwen3-Omni-30B-A3B-Instruct}} \\
\midrule
\rowcolor{gray!6}
Oracle$^\star$ & Text
& 90.7 & 94.8
& 79.4 & 81.9
& 76.2 & 67.9
& -- & --
& 82.1 & 81.5 & 0.6 \\

\rowcolor{gray!6}
Oracle$^\star$ & Multimodal
& 91.9 & 84.4
& 87.8 & 53.5
& 81.2 & 59.6
& -- & --
& 87.0 & 65.8 & 21.1 \\

\midrule
Text & Text
& 75.1 & 49.3
& 74.9 & 48.6
& 69.5 & 49.6
& 44.5 & 87.2
& 73.2 & 49.2 & 24.0 \\

Text & Multimodal
& 76.8 & 47.3
& 85.7 & \textbf{50.7}
& 75.3 & \textbf{56.3}
& 45.6 & 91.6
& 79.3 & \textbf{51.4} & 27.8 \\

Multimodal & Text
& 78.7 & 47.0
& 70.0 & 36.2
& 59.2 & 47.8
& 48.9 & \textbf{94.7}
& 69.3 & 43.7 & 25.6 \\

Multimodal & Multimodal
& 86.0 & \textbf{56.7}
& 79.8 & 42.2
& 67.7 & 54.5
& 48.6 & 93.8
& 77.8 & 51.1 & 26.7 \\
\midrule

\rowcolor{gray!10}
\multicolumn{13}{c}{\textit{MiniCPM-o-4.5}} \\
\midrule
\rowcolor{gray!6}
Oracle$^\star$ & Text
& 91.7 & 91.1
& 81.2 & 81.9
& 74.9 & 60.8
& -- & --
& 82.6 & 77.9 & 4.7 \\

\rowcolor{gray!6}
Oracle$^\star$ & Multimodal
& 59.3 & 43.1
& 59.1 & 39.4
& 55.7 & 49.3
& -- & --
& 58.0 & 43.9 & 14.1 \\

\midrule
Text & Text
& 75.0 & 43.1
& 74.0 & \textbf{48.6}
& 62.8 & \textbf{47.0}
& 49.2 & 89.8
& 70.6 & \textbf{46.2} & 24.4 \\

Text & Multimodal
& 55.7 & 32.4
& 55.2 & 38.7
& 52.9 & 42.2
& 44.5 & 91.4
& 54.6 & 37.8 & 16.8 \\

Multimodal & Text
& 78.5 & \textbf{43.6}
& 69.5 & 39.4
& 57.0 & 45.5
& 53.8 & \textbf{94.7}
& 68.3 & 42.8 & 25.5 \\

Multimodal & Multimodal
& 54.0 & 29.0
& 56.1 & 38.3
& 50.2 & 45.9
& 50.3 & 92.1
& 53.4 & 37.7 & 15.7 \\

\bottomrule

\end{tabular}
\caption{
Results of the RAG-based multimodal memory variant.
Human (Full) reads the complete native multimodal history; Human (Oracle) receives only oracle evidence.
Oracle$^\star$ uses annotated clues from the original multimodal history, bypassing retrieval.
Memory Avg. averages the first three tasks; best non-oracle model scores are bold.
}
\label{tab:rq3_multimodal}
\end{table*}

Motivated by RQ1, we isolate the role of textualization by evaluating samples whose oracle clues contain the target modality, using Qwen3.6-35B-A3B and averaging across memory systems. For images, GPT-5.4 generates Brief, Medium, and Detailed captions through prompts that increasingly emphasize peripheral details. For audio, ASR uses Qwen3-ASR-1.7B~\citep{shi2026qwen3asr}; Hint uses Gemini 3.1 Pro to caption mixed speech and background audio; Split separates them, applying Qwen3-ASR-1.7B to speech and Gemini 3.1 Pro to background audio. Appendix~C.3 details these configurations.

\textbf{\textit{Cue-aware textualization substantially improves implicit memory, but the EI-Gap remains.}}
Figure~\ref{fig:rq2_caption_quality}(a--b) shows that Detailed image captions raise implicit Memory Avg. from 26.9 to 46.4 and reduce the EI-Gap from 35.9 to 27.1 points. Audio splitting similarly raises implicit Memory Avg. from 40.1 to 51.0 and reduces the gap from 29.4 to 19.4 points. Richer preservation of background cues therefore alleviates, but does not resolve, the textualization bottleneck.

\textbf{\textit{The benefits of textualization are capability-dependent rather than universal.}}
Detailed image captions improve implicit Entity Recall and Long Pattern by 36.9 and 13.0 points, but Personalized Recommendation by only 8.8 points; audio splitting exhibits a similar capability-dependent pattern. Meanwhile, implicit Answer Refusal drops by 16.9 points for images. Better cue preservation aids evidence recovery and aggregation, but does not ensure effective evidence use and may also encourage over-inference from distractors.

\textbf{\textit{Richer cue preservation comes with a substantial efficiency cost.}}
As shown in Figure~\ref{fig:rq2_token_tradeoff}, Detailed captions yield the largest implicit accuracy gain, but increase retrieved input from 1.9k tokens per QA with Medium captions to 7.0k. Over the same change, implicit accuracy rises from 53.3 to 62.2, while explicit accuracy increases only from 70.9 to 72.4. Richer preservation thus improves implicit memory at a substantial and uneven inference cost; Appendix~D.2 provides accuracy and token statistics for caption settings.

\subsection{RQ3: Can Native Multimodal Access Alleviate the Textualization Bottleneck?}

\begin{figure}[t]
\centering
\includegraphics[width=\linewidth]{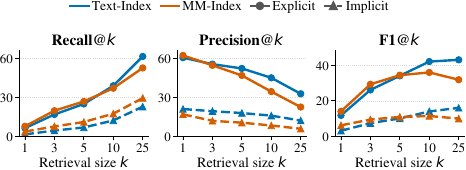}
\caption{
Supporting-evidence retrieval on explicit and implicit samples.
MM-Index yields modest recall gains but lower precision, particularly for implicit evidence; F1 summarizes the resulting trade-off.
}
\label{fig:rq3_retrieval_recall}
\end{figure}

RQ2 shows that richer textualization recovers more implicit cues but remains costly and incomplete. We test native multimodal access during indexing and evidence use in our RAG-based system (Table~\ref{tab:rq3_multimodal}). Text-Index and MM-Index encode textualized histories and original multimodal content, respectively. Text-Use and MM-Use supply retrieved textualized evidence and the corresponding original text, images, and audio, respectively. Appendix~C.4 provides implementation details for indexing, retrieval, and multimodal evidence use.

\textbf{\textit{The benefit of native multimodal evidence use is backbone-dependent.}}
Holding the index fixed, MM-Use improves Qwen's implicit Memory Avg. by 2.2 points with Text-Index and 7.4 points with MM-Index. For MiniCPM, it reduces performance by 8.4 and 5.1 points, respectively. Original modalities can recover information lost during textualization, but realizing this benefit depends on the backbone's ability to interpret and integrate evidence.

\textbf{\textit{Oracle evidence exposes a multimodal evidence-use bottleneck.}}
With oracle multimodal evidence, Qwen and MiniCPM trail their text-oracle scores by 15.7 and 34.0 points, while humans reach 87.9, versus 54.2 on full histories. Thus, localized native cues are interpretable, but models still struggle to extract and combine them.

\textbf{\textit{Multimodal indexing trades small recall gains for substantial noise.}}
Holding evidence use fixed, MM-Index lowers implicit Memory Avg. for both backbones. The drops are 5.5 (Text-Use) and 0.3 (MM-Use) points for Qwen, and 3.4 and 0.1 points for MiniCPM. Figure~\ref{fig:rq3_retrieval_recall} shows why: higher implicit recall comes with sharply lower precision, so context noise offsets the recovered evidence. Appendix~D.3 reports complete retrieval metrics. Appendix~E analyzes how failures in evidence preservation, retrieval, and use affect final answers across tasks and modalities.

\medskip
\noindent\colorbox{gray!10}{%
\parbox{\dimexpr\columnwidth-2\fboxsep\relax}{%
\textbf{Takeaway.} Across paradigms, implicit cues remain harder to recover than explicit evidence. Cue-aware textualization narrows this gap at growing token cost. Native multimodal access preserves missing information, but gains depend on the backbone; human oracle results expose weak model integration, while multimodal indexing adds noise.}}

\section{Conclusion}

We introduce \textsc{CUE-Mem}, a text-image-audio benchmark for long-term user memory from implicit multimodal cues. Paired explicit and implicit settings reveal a persistent gap across textualized and native multimodal systems. Cue-aware textualization narrows this gap at growing token cost, but mainly aids recovery and can weaken refusal calibration. Native multimodal access can preserve information lost through textualization, yet its benefits are backbone-dependent: multimodal indexing introduces noise, and native multimodal oracle evidence remains difficult for the evaluated models to interpret and integrate. Reliable implicit memory requires retaining subtle cues across long and noisy interaction histories while retrieving and using them selectively. \textsc{CUE-Mem} provides a testbed for these capabilities.

\bigskip

\bibliography{aaai2027}


\clearpage
\input{appendix}

\end{document}

%% file: appendix.tex
\setcounter{topnumber}{5}
\setcounter{bottomnumber}{3}
\setcounter{totalnumber}{8}
\setcounter{dbltopnumber}{5}
\renewcommand{\topfraction}{0.95}
\renewcommand{\bottomfraction}{0.95}
\renewcommand{\textfraction}{0.05}
\renewcommand{\floatpagefraction}{0.75}
\renewcommand{\dbltopfraction}{0.95}
\renewcommand{\dblfloatpagefraction}{0.75}
\makeatletter
\setlength{\@fptop}{0pt}
\setlength{\@fpsep}{12pt plus 2pt minus 2pt}
\setlength{\@fpbot}{0pt plus 1fil}
\setlength{\@dblfptop}{0pt}
\setlength{\@dblfpsep}{12pt plus 2pt minus 2pt}
\setlength{\@dblfpbot}{0pt plus 1fil}
\makeatother
\setcounter{secnumdepth}{2}
\appendix

\section{Construction and Quality-Control Details}

\subsection{Evidence Taxonomy and Assignment Prompts}
\label{app:evidence_assignment}

\paragraph{Operational taxonomy.}
We define evidence type at the level of a target user memory, rather than at
the level of an entire session:
\begin{itemize}
  \small
  \setlength{\itemsep}{0pt}
  \setlength{\parsep}{0pt}
  \setlength{\topsep}{2pt}
  \item \textbf{Explicit:} the target attribute is directly asserted in the
  dialogue or is the foreground subject of a shared image.
  \item \textbf{Implicit:} the target is never stated and must instead be
  inferred from a peripheral visual object or an ambient, non-speech sound.
\end{itemize}
Thus, ``implicit'' does not mean that the answer
requires ordinary semantic inference from an explicit sentence.  It denotes
contextual, unstated evidence that is separable from the current conversational
topic.  Table~\ref{tab:evidence_taxonomy} gives the modality-specific
realization rules.  A target may have more than one source modality, but the
same explicit/implicit criterion applies to all of its supporting occurrences.
In particular, any target with a dialogue source is explicit; implicit targets
cannot have a dialogue source.

\begin{table*}[!htbp]
\centering
\small
\setlength{\tabcolsep}{5pt}
\begin{tabular}{p{0.09\textwidth}p{0.28\textwidth}p{0.28\textwidth}p{0.25\textwidth}}
\toprule
\textbf{Modality} & \textbf{Explicit evidence} & \textbf{Implicit evidence} & \textbf{Construction-time validity test} \\
\midrule
Text &
The user directly states the target fact, preference, relationship, habit, or
routine. &
Not used as a target-bearing source.  Text may establish the foreground
conversation, but may neither name nor paraphrase the implicit target. &
The target and close paraphrases must occur in an explicit script and must be
absent from every implicit dialogue turn and assistant reply. \\
\addlinespace
Image &
The target person, pet, place, or object is the centered or otherwise dominant
subject and may be discussed in the dialogue. &
A recurring target object is visible only at an edge, corner, or background
region, possibly partially occluded; it is not discussed. &
The target is recognizable at inspection, yet peripheral to the foreground
subject and semantically disjoint from the dialogue. \\
\addlinespace
Audio &
The target information is conveyed by the user's speech; the waveform is
treated as another carrier of directly stated content. &
A recognizable environmental sound is mixed behind the user's speech and is
absent from the speech transcript. &
The ambient event is audible without masking speech, is compatible with the
scene, and cannot be recovered from ASR alone. \\
\bottomrule
\end{tabular}
\caption{Operational evidence taxonomy.  Foreground/background status and
whether the target is stated jointly determine explicitness; modality alone
does not.}
\label{tab:evidence_taxonomy}
\end{table*}

\paragraph{From profiles to sessions and tasks.}
We first expand each persona seed into a structured profile containing basic
attributes, relationships, pets, and six preference domains:
\begin{itemize}
  \small
  \setlength{\itemsep}{0pt}
  \setlength{\parsep}{0pt}
  \setlength{\topsep}{2pt}
  \item \emph{Food and Drink}
  \item \emph{Home and Space}
  \item \emph{Body and Health}
  \item \emph{Hobbies and Entertainment}
  \item \emph{Work and Learning}
  \item \emph{Mobility and Travel}
\end{itemize}
For each user, the proposal agent generates
6--8 implicit and 6--8 explicit preferences.  At least four implicit
preferences must admit visual realization, and at least four explicit
preferences must admit both dialogue and foreground-image realization.
Relationship and pet existence is always introduced explicitly; an unseen
relative or pet is never introduced solely through a photograph or animal
sound.

Each retained target stores a category identifier, a concrete memory value,
an evidence type, one or more source modalities, and a realization rationale.
For audio targets, the rationale begins with a canonical sound-event keyword;
for visual targets it specifies the object and its peripheral placement.  Event
pairing and assignment follow these rules:
\begin{itemize}
  \small
  \setlength{\itemsep}{0pt}
  \setlength{\parsep}{0pt}
  \setlength{\topsep}{2pt}
  \item Each coherent event plan pairs exactly one explicit preference with
  one implicit preference.
  \item The local validator copies targets from the source profile rather than
  trusting model-generated fields and rejects invalid category identifiers.
  \item Every explicit preference recurs in at least two events, and every
  implicit preference recurs in at least three.
  \item A visual implicit cue may only be paired with a photographable explicit
  foreground subject.
  \item Across a user's 30--45 event groups, partners and scenes are varied to
  avoid a fixed one-to-one association.
  \item Events are ordered chronologically, converted to multi-turn sessions,
  and assigned stable identifiers that later connect each QA clue to a turn,
  image, or audio clip.
\end{itemize}

Preference targets support Long Pattern and Personalized Recommendation;
people, pets, and recurring items support Entity Recall.  Answer Refusal is
constructed from attributes that are absent or insufficiently supported.
Consequently, a task label is assigned only after its evidence occurrences
have been placed and verified.  Explicit and implicit examples are balanced
at the aggregate level; they are not one-to-one counterfactual pairs.

\noindent\textbf{Prompt template: evidence proposal.}\quad
The following abridged template preserves the constraints used by the
generation script (the complete construction prompt contains the full category
descriptions and JSON schema):
\begin{examplebox}
\textbf{Evidence Proposal Prompt.}

\medskip
\textbf{Given the Basic/Relationship/Pets profile, propose 6--8 concrete
[implicit|explicit] preferences across the six domains.  For implicit
preferences, use only visual and/or audio evidence; never dialogue.  Place
visual evidence at an edge/corner/background and prefix every audio rationale
with its assigned canonical sound keyword.  For explicit preferences, require
dialogue evidence and optionally a foreground image; never audio ambience.
Do not invent people or pets, and keep explicit targets unrelated to the
already generated implicit targets.  Return JSON only.}
\end{examplebox}
Implicit targets are proposed first.  The resulting list is supplied to the
explicit proposal prompt as an exclusion set, preventing the two conditions
from describing the same underlying habit in different words.  The proposal
model is \textbf{deepseek-v4-pro} with high reasoning effort; malformed,
out-of-range, or keyword-incomplete JSON is retried up to three times.

\noindent\textbf{Prompt template: event assignment.}\par
\begin{examplebox}
\textbf{Event Assignment Prompt.}

\medskip
\textbf{Pair the supplied candidates into 30--45 groups.  Each group contains
exactly one explicit and one implicit target copied verbatim from the
manifest.  Cover every explicit target at least twice and every implicit
target at least three times.  Use varied partners and a coherent shared
physical scene.  If the implicit source is visual, choose a photographable
explicit foreground subject and keep the implicit anchor only at the
edge/corner.  The recommended main scene may describe only the explicit
target and must contain no implicit object, action, environment, sound, or
hint.  Return JSON only.}
\end{examplebox}
Assignment uses \textbf{deepseek-v4-pro} at temperature 0.2.  A deterministic
validator recomputes coverage, checks visual compatibility and anchor
presence, detects leakage in the recommended scene, and sends the full error
list back for at most two repair rounds.

\paragraph{Ambient-event vocabulary.}
We pooled event labels from ESC-50~\citep{piczak2015esc} and
FSD50K~\citep{fonseca2021fsd50k}, then removed speech-dominant, unsafe,
near-duplicate, culturally incompatible, or difficult-to-place events.
Manual deduplication merged 17 groups of synonymous labels, reducing 113 raw
entries to 96 usable event types.
Table~\ref{tab:audio_pool} groups representative retained labels; the
construction metadata retain the full canonical-label and alias mapping.

\begin{table}[!htbp]
\centering
\small
\setlength{\tabcolsep}{4pt}
\begin{tabular}{p{0.26\columnwidth}p{0.64\columnwidth}}
\toprule
\textbf{Family} & \textbf{Representative retained event types} \\
\midrule
Animals/nature & dog, cat, birds, rooster, waves, rain, wind \\
Domestic & vacuum, doorbell, clock, alarm, microwave, boiling \\
Food/drink & chopping, frying, cutlery, pouring, chewing, coffee grinder \\
Work/learning & typing, mouse click, printer, writing, chalk, telephone \\
Transport/travel & bicycle bell, subway, engine, horn, aircraft, suitcase \\
Sports/leisure & running, basketball, tennis, swimming, camera shutter \\
Music/social & piano, guitar, drums, applause, cheering, laughter \\
\midrule
\multicolumn{2}{l}{\textbf{Total after human deduplication: 96 event types.}} \\
\bottomrule
\end{tabular}
\caption{Semantic families in the retained ambient-event vocabulary.}
\label{tab:audio_pool}
\end{table}

\subsection{Multimodal Synthesis Details}
\label{app:synthesis}

\paragraph{Event and dialogue scripts.}
For each validated group, \textbf{deepseek-v4-pro} first generates an event
record with date, setting, foreground topic, modality plan, and separate fields
for the image description, human speech, and background audio (temperature
0.8).  Later events are conditioned on earlier events from the same user.
Dialogue is then generated at temperature 0.8 with at least eight user and
eight assistant turns and at most 100 characters per message.  The prompt
contains the user's complete implicit-target list as a prohibition set.  A
second pass (temperature 0.4) assigns each required ambient keyword to 3--8
user turns; each tag is at most 18 Chinese characters, describes a single
sound, and may recur over 2--3 adjacent user turns.  Local checks reject
non-user placements, multiple sounds in one tag, missing canonical keywords,
or fewer than three valid placements.

\paragraph{Images.}
We first create reference assets for recurring people, pets, and profile
objects.  An event image is a 1:1, 1024$\times$1024 first-person smartphone
photograph.  Its prompt requests a realistic, sharp image under natural
lighting with no text, logo, watermark, or visible photographer.  Reference
assets are supplied inline, and identity instructions require the same face,
hairstyle, age, body type, skin tone, pet markings, and object material/color
across sessions.  Explicit visual evidence occupies the foreground.  For an
implicit target, the event-description agent must use an edge/corner/background
phrase and expose only a small or partly occluded region.  Ordinary foreground
objects and scene-appropriate clutter serve as distractors; the prompt is not
allowed to make the target the sole, centered, or highlighted object.

We use \textbf{gemini-3-pro-image-preview} both to generate reference images
for recurring people specified under \emph{Relationships}, pets, and personal
items, and to synthesize the photographs associated with subsequent events.
Whenever a recurring entity appears in an event photograph, its corresponding
reference image is provided as visual conditioning.  This reference-guided
generation preserves the characteristic appearance of people, pets, and
objects across sessions and provides reliable visual grounding for the
identify-type questions in Entity Recall.  Human reviewers subsequently
compare each recurring entity against its reference and across event images;
only instances that maintain consistent identity and appearance are retained.

\paragraph{Speech and ambient audio.}
Each user receives a voice design generated by
\textbf{Qwen3-TTS-12Hz-1.7B-VoiceDesign} from the profile's gender, age, and
voice-timbre description.  The fixed anchor utterance introduces the user's
name.  All subsequent Chinese user utterances are synthesized with
\textbf{Qwen3-TTS-12Hz-1.7B-Base} in x-vector-only voice-cloning mode using
that anchor, bfloat16 inference, and eager attention.  This single-anchor
design fixes speaker identity across sessions.

Ambient tracks are generated from the canonical event prompt using Kling's
\textbf{/v1/audio/text-to-audio} endpoint at a requested duration of 5\,s.
Tracks are cached by normalized prompt, allowing the same event type to be
reused without an additional stochastic generation.  The track is repeated or
trimmed to the speech duration and mixed only into the turns selected above.
Speech remains the dominant channel and the background must remain audible;
clips with masked speech, inaudible cues, clipping, or a mismatch between tag
and sound are rejected during review.

\paragraph{Representative realizations.}
For ``keeps small succulents,'' the explicit event may center a pet playing on
the living-room floor, while a consistent blue-green succulent and wooden
plant tray occupy only a sliver of the windowsill at the image edge.  For
``regularly cooks at home,'' the conversation discusses an unrelated explicit
home preference while canonical chopping and frying events recur behind
several user utterances.  In both cases the target is repeated across sessions,
but neither user nor assistant names it.

\subsection{Automatic and Human Quality Control}
\label{app:quality_control}

\paragraph{Automatic checks.}
Quality control is applied to the event-level source records before QA
construction and again to question-level clues after QA generation.
Table~\ref{tab:qc_criteria} summarizes the acceptance criteria.  Structural
checks require parseable schemas, valid user/session/turn identifiers,
existing media paths, non-empty media, alternating dialogue roles, and an
exact match between planned and realized target identifiers.  Text checks
search the full user--assistant dialogue for the target and close paraphrases;
any match in an implicit event is leakage.  Image checks require the intended
foreground subject, target recognizability, and small peripheral placement of
an implicit target.  Audio checks require an audible tagged event that is
absent from the speech transcript, while the transcript must remain
intelligible.

\begin{table*}[!htbp]
\centering
\small
\setlength{\tabcolsep}{4.5pt}
\begin{tabular}{p{0.18\textwidth}p{0.47\textwidth}p{0.25\textwidth}}
\toprule
\textbf{Check} & \textbf{Acceptance criterion} & \textbf{Typical rejection} \\
\midrule
Schema and synthesis &
All required fields and referenced assets exist; dialogue and media are
natural, readable/audible, and mutually coherent. &
Malformed record, corrupt media, unnatural script, modality mismatch. \\
Cross-session visual identity &
SigLIP~2 similarity between each recurring entity and its profile reference
passes the fixed identity threshold; incompatible references are rejected. &
Changed face, pet markings, object color/material, or swapped identity. \\
Cross-session speaker identity &
ERes2NetV2 embeddings are used to compare every synthesized utterance with the
user's fixed voice anchor and enforce cross-session speaker consistency. &
Speaker drift, wrong user voice, unstable timbre. \\
Implicit visual validity &
Dialogue--target semantic overlap remains below the leakage threshold, and
the implicit target's occupied fraction of the full image remains below the
salience threshold while remaining recognizable. &
Target named/paraphrased, centered, dominant, fully exposed, or absent. \\
Implicit audio validity &
The canonical ambient event is audible, scene-compatible, and absent from the
speech transcript; speech is not masked. &
ASR leakage, inaudible/wrong ambience, clipping, masked speech. \\
QA and clue validity &
One unambiguous answer is supported by all retained clue IDs; distractors are
plausible but unsupported, and evidence type and modality labels agree with
the source profile. &
Wrong label, invalid clue ID, ambiguous option, insufficient support, or
explicitness mismatch. \\
\bottomrule
\end{tabular}
\caption{Automatic and human-auditable quality criteria.}
\label{tab:qc_criteria}
\end{table*}

Long-term visual consistency is measured with SigLIP~2
(\textbf{google/siglip2-so400m-patch14-384}) against the corresponding profile
reference.  We use ERes2NetV2 as the speaker encoder for cross-session
speaker-consistency verification.  Each synthesized utterance is compared
with the user's fixed reference voice in the ERes2NetV2 embedding space, and
recordings exhibiting speaker drift or an inconsistent voice identity are
rejected.

\paragraph{Clue-level verification.}
After a QA agent proposes clue identifiers, identifiers are resolved against
the formatted history.  Invalid identifiers are removed.  A separate
rechecking pass sends only text clues to the verifier; image and audio clues
are preserved for human inspection rather than discarded through text-only
judgment.  In implicit QAs, any text clue bearing the target is treated as
leakage and excluded from modality statistics.  When a user turn has both a
text identifier and a waveform identifier, the waveform is retained as the
atomic clue.  Entity-item explicitness is taken from the profile's
\textbf{entity\_explicitness} field rather than inferred from the item modality.
Answer Refusal has no supporting clues by construction.

\paragraph{Human review.}
Twelve annotators completed two review rounds totaling approximately 550
annotator-hours.  In the first round, three annotators independently rated
each surviving sample on (i) dialogue naturalness, (ii) multimodal coherence,
and (iii) evidence validity.  Evidence validity includes target presence,
explicitness/implicitness, recurrence, answerability, and the absence of
leakage.  A sample is directly accepted only when all three annotators approve
all three dimensions.  Any disagreement is sent to expert adjudication in the
second round; the adjudicator either accepts the original sample, requests
repair and re-review, or rejects it.  The average Fleiss' $\kappa$ over the
three dimensions is 0.73.

\begin{table}[!htbp]
\centering
\small
\setlength{\tabcolsep}{4pt}
\begin{tabular}{lrr}
\toprule
\textbf{Stage} & \textbf{Rejected at stage} & \textbf{Remaining} \\
\midrule
Generated candidates & --- & 100.00\% \\
Automatic checks & 42.80\% of input & 57.20\% \\
Human review & 22.10\% of input & 44.56\% \\
\bottomrule
\end{tabular}
\caption{Filtering flow.  ``Input'' in the human-review row denotes samples
remaining after automatic checks; hence 44.56\% of the initial candidates
survive both stages.  Per-reason counts were not preserved in the transferred
snapshot and will be added from the original annotation export.}
\label{tab:filtering_flow}
\end{table}

Automatic rejection reasons are grouped as script/schema failure, image
quality or identity failure, corrupted or inconsistent audio, cross-modal
mismatch, and implicit-cue leakage/salience.  Human rejection reasons are
unnatural dialogue, incoherent media, unrecognizable or overly salient
evidence, inconsistent recurring entities/speakers, ambiguous answers,
unsupported labels, and invalid modality assignments.  Repair is allowed only
before final acceptance and the repaired record must pass the complete check
set again.

\section{Task Construction and Complete Examples}

\subsection{QA Generation and Annotation}

\paragraph{Shared construction pipeline.}
All four tasks are stored as four-way, single-answer multiple-choice
questions.  Construction starts from a target already instantiated in the
profile and its tagged events.  For Entity Recall, Long Pattern, and
Personalized Recommendation, a generator first produces a question, one
correct option, and three parallel, same-domain distractors; the correct
position is randomized over A--D.  A second, answer-conditioned pass receives
the target, the known correct option, and every matched event, and returns all
directly supporting turn, image, and audio identifiers.  The implementation
then intersects these identifiers with the identifiers that actually occur in
the matched events and removes duplicates.  We further recheck the resulting
memory clues to ensure that they provide strong support for the correct answer
without introducing irrelevant noise.

Image-option questions use the same semantics but add a three-stage path:
generate four parallel visual descriptions, synthesize the four option images,
and extract answer-supporting history clues.  The option captions included in
the benchmark JSON are evaluation-time textualizations, not substitutes for
the generated option images in the native-multimodal setting.

\begin{table*}[!htbp]
\centering
\small
\setlength{\tabcolsep}{4.5pt}
\begin{tabular}{p{0.14\textwidth}p{0.22\textwidth}p{0.25\textwidth}p{0.29\textwidth}}
\toprule
\textbf{Task} & \textbf{Target and answerability} &
\textbf{Correct option} & \textbf{Distractor/refusal constraint} \\
\midrule
Entity Recall &
A named relationship or pet attribute, or a recurrent user-associated item
attribute, grounded in at least one retained clue. &
The profile-consistent identity, relationship, appearance, occupation, age,
personality, or object attribute. &
Three alternatives match the queried attribute and granularity.  They may
change one or more entity attributes but cannot also be true of the target. \\
\addlinespace
Long Pattern &
A profile preference linked to its tagged events; exactly one question is
generated for each preference. &
The activity, taste, routine, trigger, or owner-level tendency expressed by
that preference. &
Three plausible patterns in the same life domain.  They cannot be eliminated
solely by option length or type, and the stem cannot quote the target
preference. \\
\addlinespace
Personalized Recommendation &
A preference-grounded recommendation scenario; the recommended item itself
must not already be mentioned in the history. &
A new item or activity that matches the core style, need, setting, or value of
the user's established preference. &
At least two hard distractors remain in the same product/activity class but
differ on a preference-critical dimension; no ``all/none/uncertain'' options
are allowed. \\
\addlinespace
Answer Refusal &
A natural-looking stem formed by changing exactly one premise slot of a real
memory.  The changed premise must be unsupported. &
Exactly one option is \emph{``The history does not specify''} (the fixed
Chinese refusal string used throughout the benchmark JSON). &
A clue-supported but question-incompatible answer is retained as the lure,
with two additional same-domain distractors.  The lure's source clues are
annotations of the trap, not evidence that the corrupted premise is true. \\
\bottomrule
\end{tabular}
\caption{Task-specific construction and answerability criteria.}
\label{tab:task_construction_criteria}
\end{table*}

\paragraph{Explicit and implicit adversaries.}
The refusal generator draws from eight single-slot transformations:
\begin{itemize}
    \item \textbf{Person/owner swap.} Reassigns a true user preference or
    attribute to a different named Relationship entity.  The replacement is
    never another benchmark user.
    \item \textbf{Attribute-slot swap.} Retains the general subject but
    changes one defining object or attribute slot, such as replacing the
    referenced pet species.
    \item \textbf{Category/type swap.} Moves a real preference or behavior
    into an unsupported semantic category while preserving a superficially
    plausible description.
    \item \textbf{Lexical-concept swap.} Replaces a supported concept with a
    different, potentially biased concept that is not licensed by the memory.
    \item \textbf{Polarity flip.} Reverses the direction of a supported
    attitude, for example changing a preference into a dislike.
    \item \textbf{False presupposition.} Introduces an event or factual
    premise that never occurs in the memory and asks the question as though it
    were established.
    \item \textbf{Event-context swap.} Places a real answer in a different,
    incompatible event context, making the otherwise supported information
    inapplicable to the question.
    \item \textbf{Implicit overclaim.} Converts an inference supported only
    by peripheral imagery or ambient sound into an unsupported assertion that
    the user explicitly stated, explained, or invariably performed it.  This
    transformation is used only for implicit targets.
\end{itemize}
For every generated record, code verifies the requested transformation label,
four nonempty options, exactly one refusal option, and a valid answer letter.
It also preserves the true target's clue set and a short trap rationale so
that reviewers can check why the lure is tempting yet invalid.

\paragraph{Difficulty and final verification.}
Question-only filtering uses Qwen3.6-35B-A3B and GPT-5.4-mini.  Each backbone
has three stored runs (\textbf{run1--run3}); a non-refusal question is removed
only when all six predictions equal the gold label.  This removes 525
non-refusal questions.  Refusal questions are deliberately exempt because
the refusal phrase itself can be a question-only cue.  Pruning is then applied
identically to the base, caption-granularity, audio-caption, oracle, and RQ3
data/result trees.  The resulting base files contain 1,128 Entity Recall, 363
Long Pattern, 364 Personalized Recommendation, and 819 Answer Refusal
questions, exactly matching the main paper.

The final verifier checks option and answer fields, clue-ID existence,
explicitness metadata for item questions, and cross-variant QA identity.

\begin{figure*}[!htbp]
\centering
\setlength{\abovecaptionskip}{2pt}
\setlength{\belowcaptionskip}{0pt}
\begin{mybox}
\textbf{Entity Recall---Explicit Image Evidence}
\hfill \textbf{p3 / 3-Relationship-2-img\_profession}

\smallskip
\textbf{Memory clues.}
\emph{D08:00, user:} ``Did you see the picture? Yueyue is auscultating a
little boy in the clinic.''  The same turn foregrounds Zhang Yue in a white
coat using a stethoscope; \emph{D08:01}, \emph{D08:04}, and \emph{D08:05}
further describe her pediatric care and medical training.

\smallskip
  \begin{minipage}[t]{0.31\linewidth}\centering
  \IfFileExists{supplementary_images/event/images/pid_0003_task_3-4-0.jpg}{%
    \includegraphics[width=\linewidth,height=3.1cm,keepaspectratio]{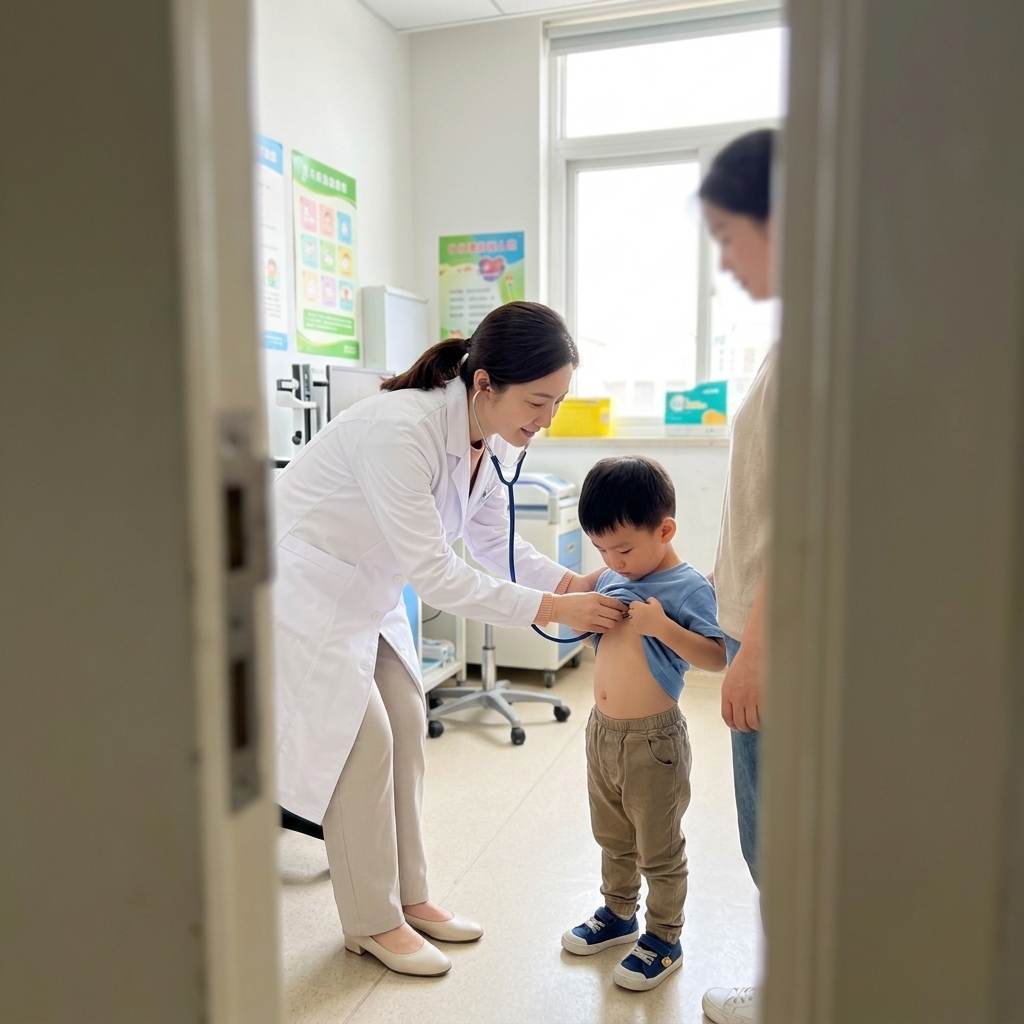}%
  }{%
    \fbox{\parbox[c][2.9cm][c]{0.88\linewidth}{\centering\scriptsize
    Original memory image\\not in transferred snapshot}}%
  }\\[0.45ex]
  \scriptsize\textbf{D08-001.png}\enspace foreground pediatric examination
  \end{minipage}

\smallskip
\textbf{Question.} Which image best matches Zhang Yue's professional work
setting?

\smallskip
  \begin{minipage}[t]{0.235\linewidth}\centering
  \IfFileExists{supplementary_images/qa/entity_images/3-Relationship-2-img_profession_A.jpg}{%
    \includegraphics[width=\linewidth,height=2.65cm,keepaspectratio]{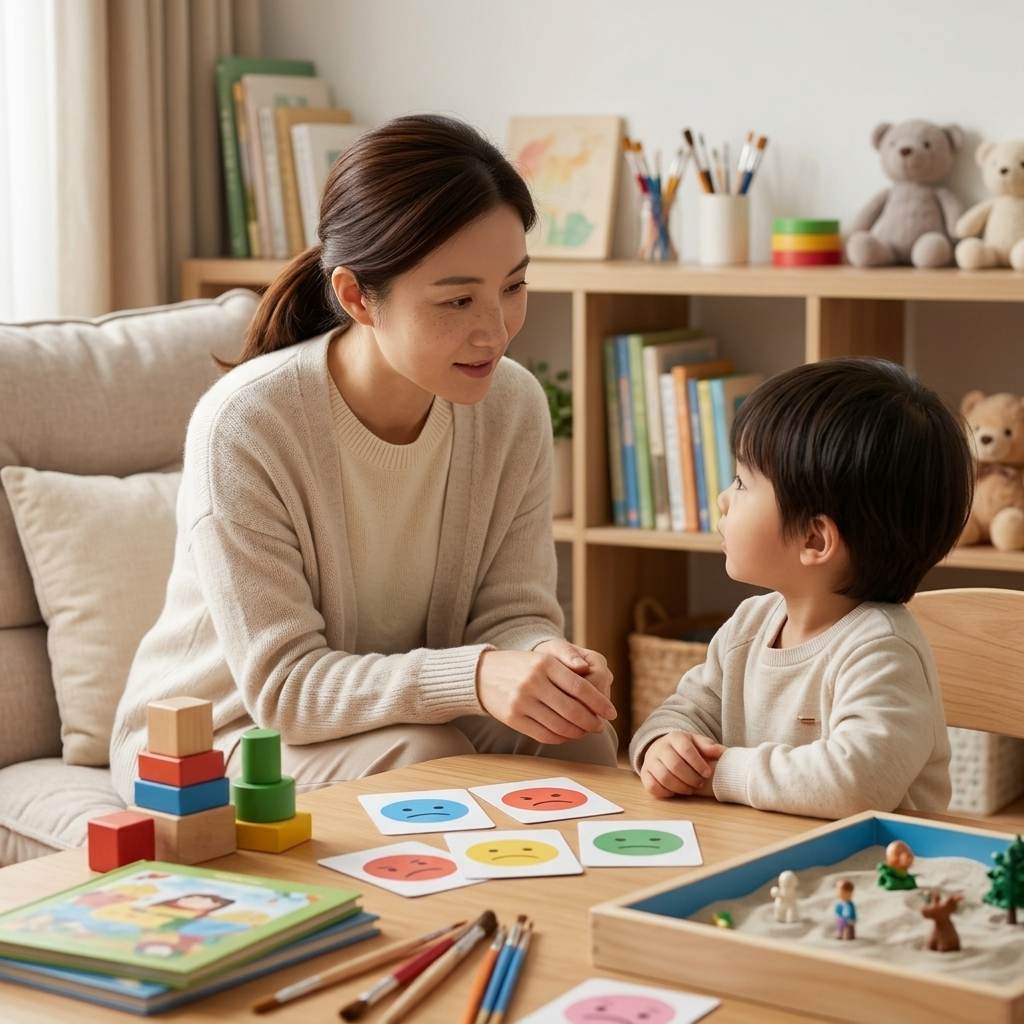}%
  }{%
    \fbox{\parbox[c][2.45cm][c]{0.88\linewidth}{\centering\scriptsize
    Option image unavailable}}%
  }\\[0.45ex]
  \scriptsize\textbf{A}\enspace child counseling
  \end{minipage}
  \begin{minipage}[t]{0.235\linewidth}\centering
  \IfFileExists{supplementary_images/qa/entity_images/3-Relationship-2-img_profession_B.jpg}{%
    \includegraphics[width=\linewidth,height=2.65cm,keepaspectratio]{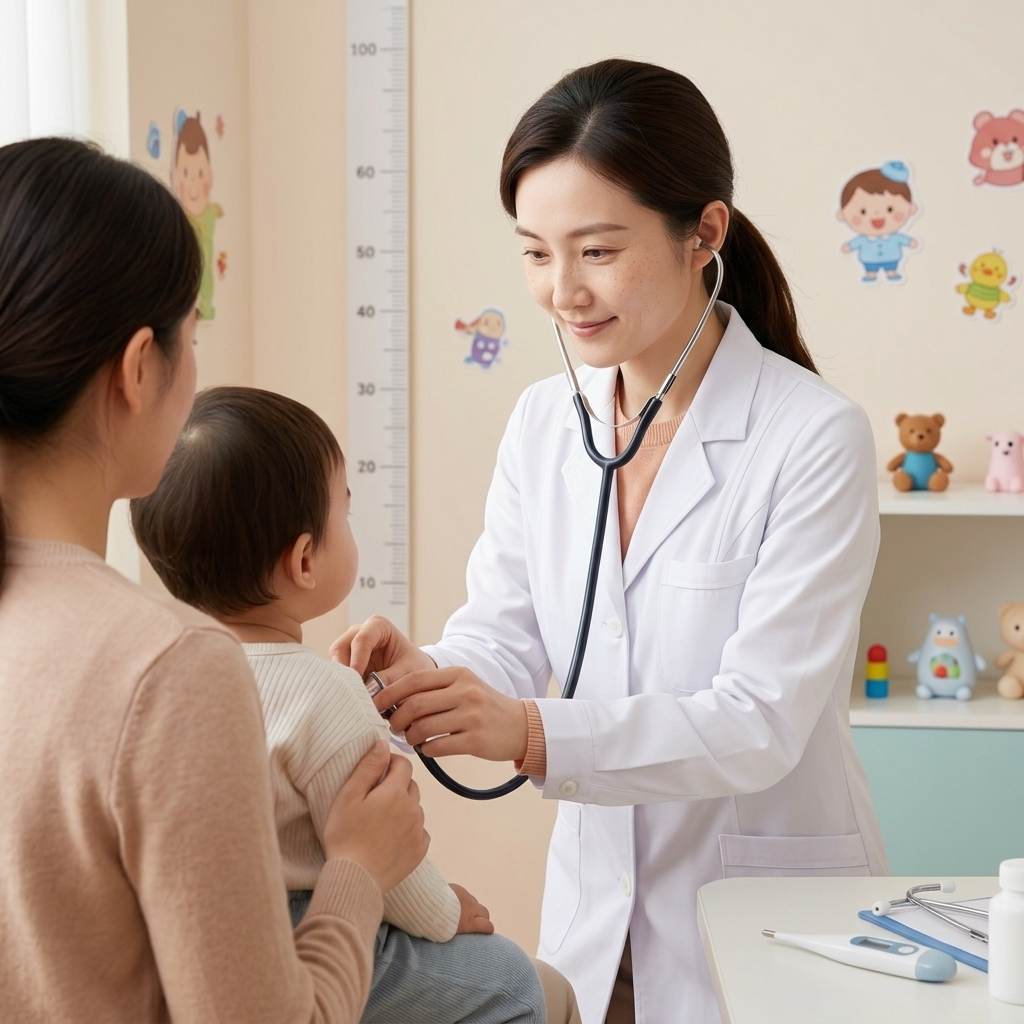}%
  }{%
    \fbox{\parbox[c][2.45cm][c]{0.88\linewidth}{\centering\scriptsize
    Option image unavailable}}%
  }\\[0.45ex]
  \scriptsize\textbf{B}\enspace pediatric clinic
  \end{minipage}
  \begin{minipage}[t]{0.235\linewidth}\centering
  \IfFileExists{supplementary_images/qa/entity_images/3-Relationship-2-img_profession_C.jpg}{%
    \includegraphics[width=\linewidth,height=2.65cm,keepaspectratio]{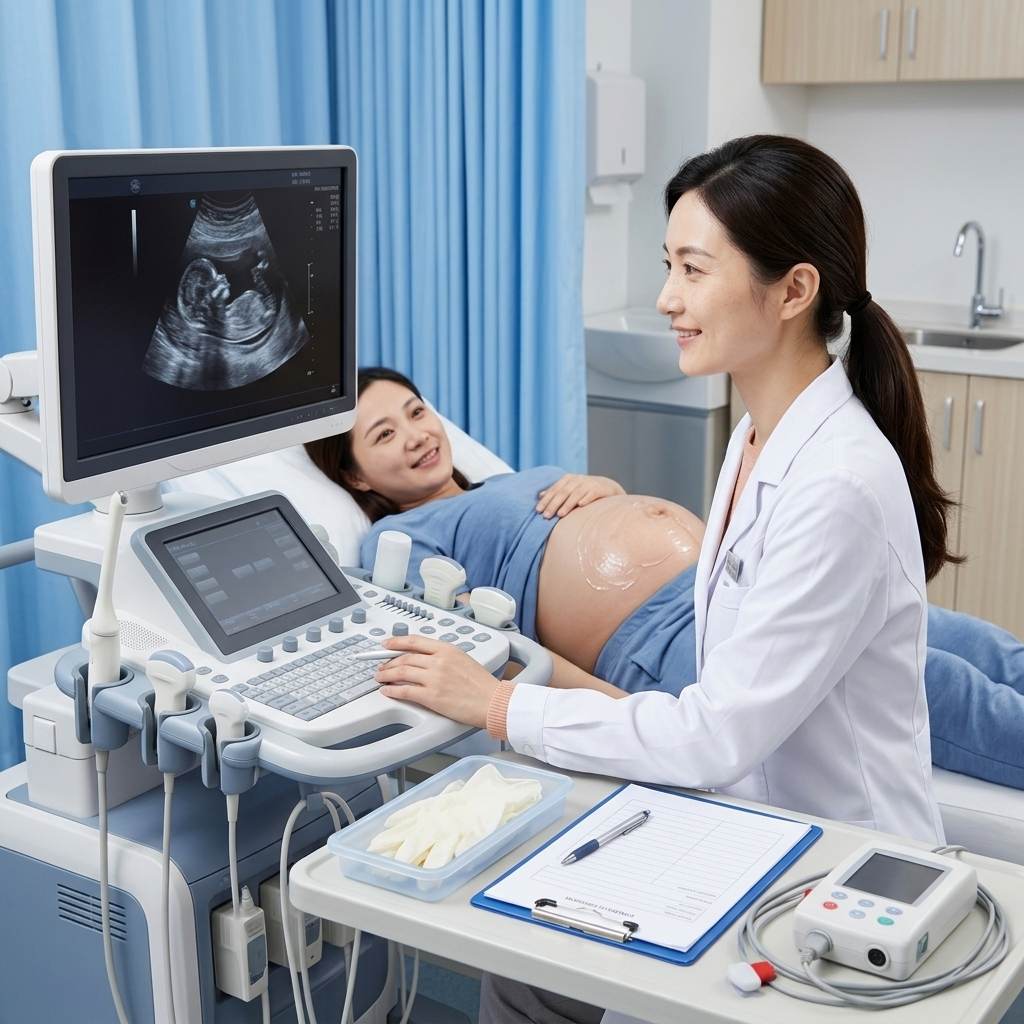}%
  }{%
    \fbox{\parbox[c][2.45cm][c]{0.88\linewidth}{\centering\scriptsize
    Option image unavailable}}%
  }\\[0.45ex]
  \scriptsize\textbf{C}\enspace obstetric ultrasound
  \end{minipage}
  \begin{minipage}[t]{0.235\linewidth}\centering
  \IfFileExists{supplementary_images/qa/entity_images/3-Relationship-2-img_profession_D.jpg}{%
    \includegraphics[width=\linewidth,height=2.65cm,keepaspectratio]{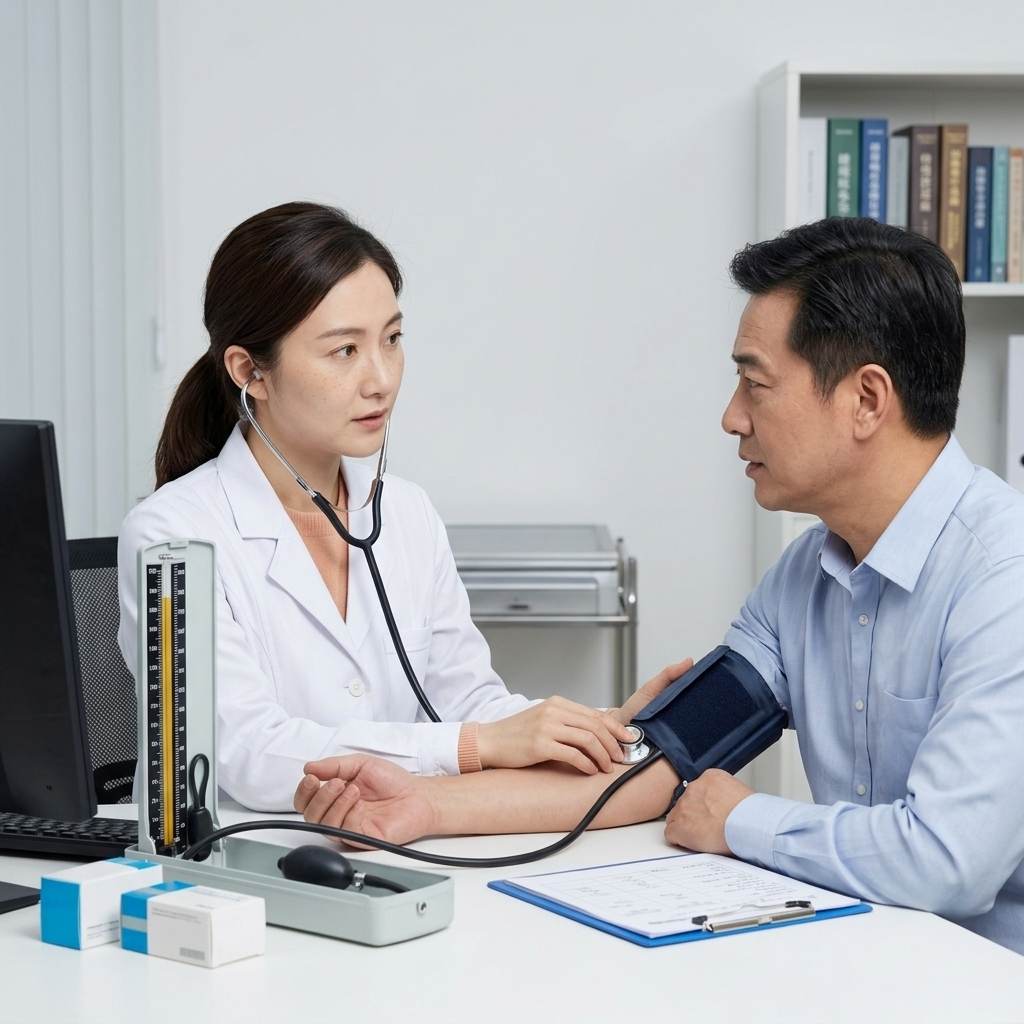}%
  }{%
    \fbox{\parbox[c][2.45cm][c]{0.88\linewidth}{\centering\scriptsize
    Option image unavailable}}%
  }\\[0.45ex]
  \scriptsize\textbf{D}\enspace adult clinic
  \end{minipage}

\smallskip
\textbf{Gold answer: B.}\quad The occupation is directly stated and visually
foregrounded, so the evidence is explicit.
\end{mybox}
\caption{Complete explicit Entity Recall example.}
\label{fig:example-entity-explicit}
\end{figure*}

\subsection{Complete Task Examples}
\label{app:task_examples}

\begin{figure*}[!htbp]
\centering
\begin{mybox}
\textbf{Entity Recall---Implicit Image Evidence}
\hfill \textbf{p0 / 0-Items-6-img\_identify}

\medskip
\textbf{Memory clues.}
The same dog-eared issue of \emph{Wallpaper*} recurs incidentally across five
sessions.  The foreground conversations concern mood-board review
(\emph{D06}), a colleague's analysis (\emph{D12}), cats (\emph{D17} and
\emph{D21}), and route planning (\emph{D28}).  The dialogue never states the
magazine's title or visual identity.

\medskip
  \begin{minipage}[t]{0.19\linewidth}\centering
  \IfFileExists{supplementary_images/event/images/pid_0000_task_0-13-0.jpg}{%
    \includegraphics[width=\linewidth,height=2.45cm,keepaspectratio]{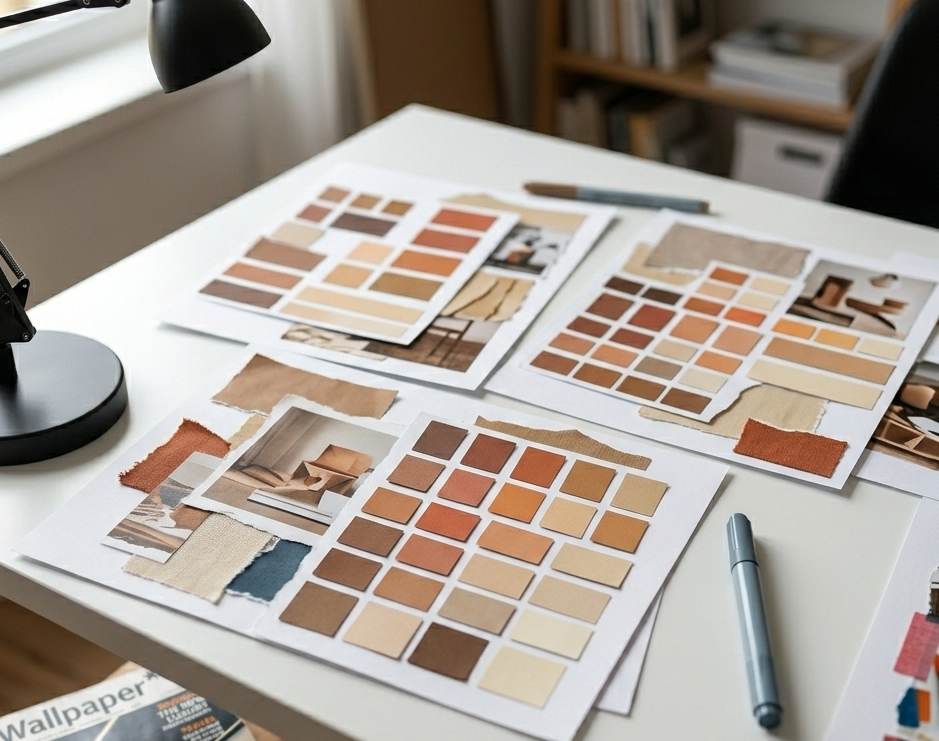}%
  }{%
    \fbox{\parbox[c][2.25cm][c]{0.86\linewidth}{\centering\scriptsize
    Memory image unavailable}}%
  }\\[0.45ex]
  \scriptsize\textbf{D06-001.png}\\[-0.2ex]mood-board desk
  \end{minipage}
  \begin{minipage}[t]{0.19\linewidth}\centering
  \IfFileExists{supplementary_images/event/images/pid_0000_task_0-20-0.jpg}{%
    \includegraphics[width=\linewidth,height=2.45cm,keepaspectratio]{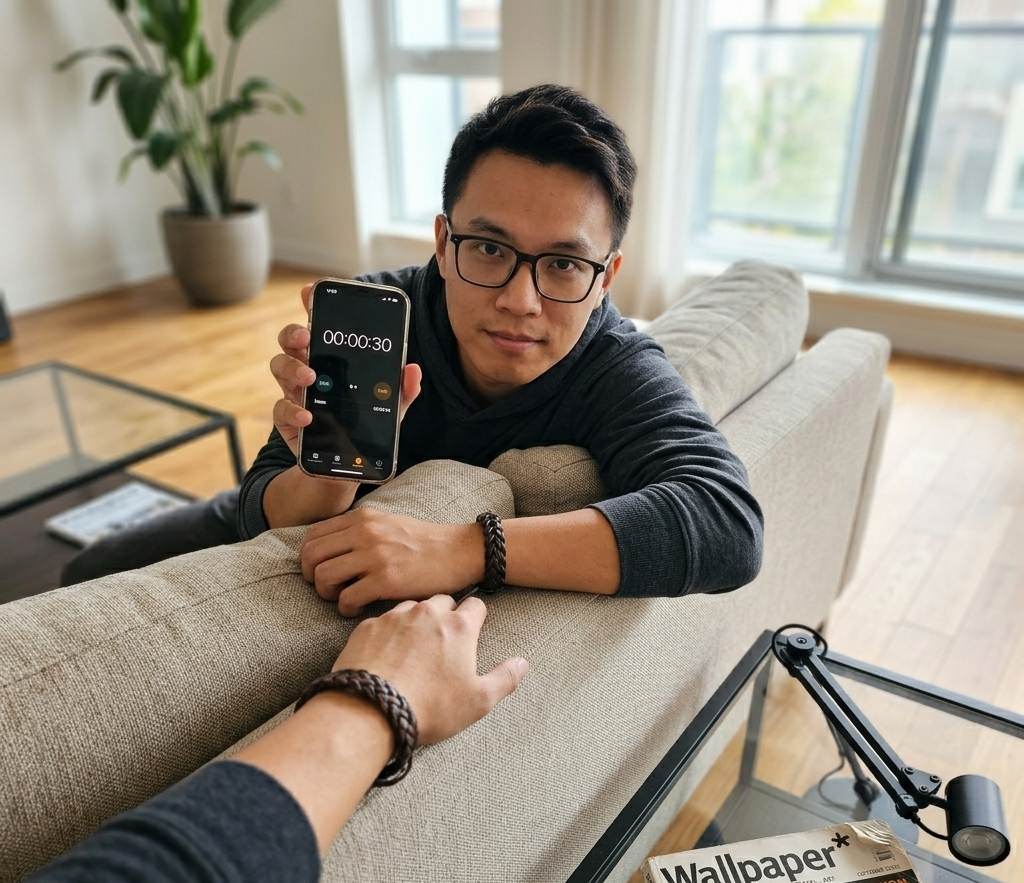}%
  }{%
    \fbox{\parbox[c][2.25cm][c]{0.86\linewidth}{\centering\scriptsize
    Memory image unavailable}}%
  }\\[0.45ex]
  \scriptsize\textbf{D12-001.png}\\[-0.2ex]living-room table
  \end{minipage}
  \begin{minipage}[t]{0.19\linewidth}\centering
  \IfFileExists{supplementary_images/event/images/pid_0000_task_0-6-0.jpg}{%
    \includegraphics[width=\linewidth,height=2.45cm,keepaspectratio]{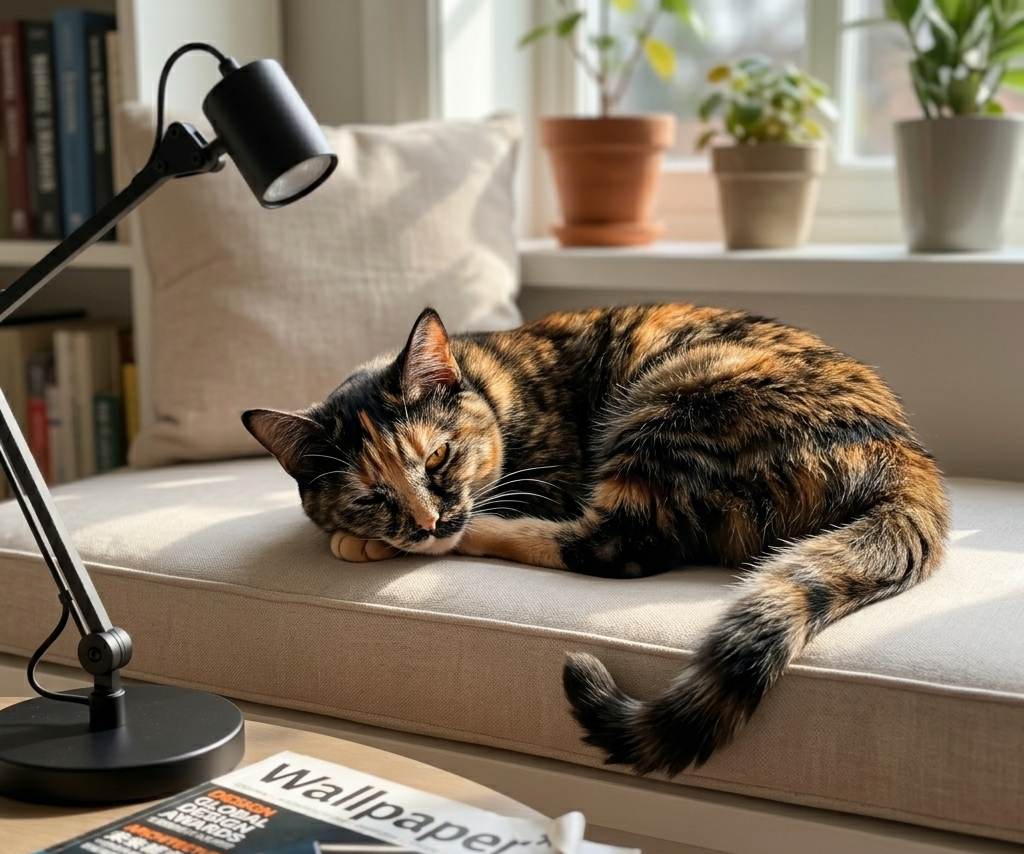}%
  }{%
    \fbox{\parbox[c][2.25cm][c]{0.86\linewidth}{\centering\scriptsize
    Memory image unavailable}}%
  }\\[0.45ex]
  \scriptsize\textbf{D17-001.png}\\[-0.2ex]beside resting cat
  \end{minipage}
  \begin{minipage}[t]{0.19\linewidth}\centering
  \IfFileExists{supplementary_images/event/images/pid_0000_task_0-2-0.jpg}{%
    \includegraphics[width=\linewidth,height=2.45cm,keepaspectratio]{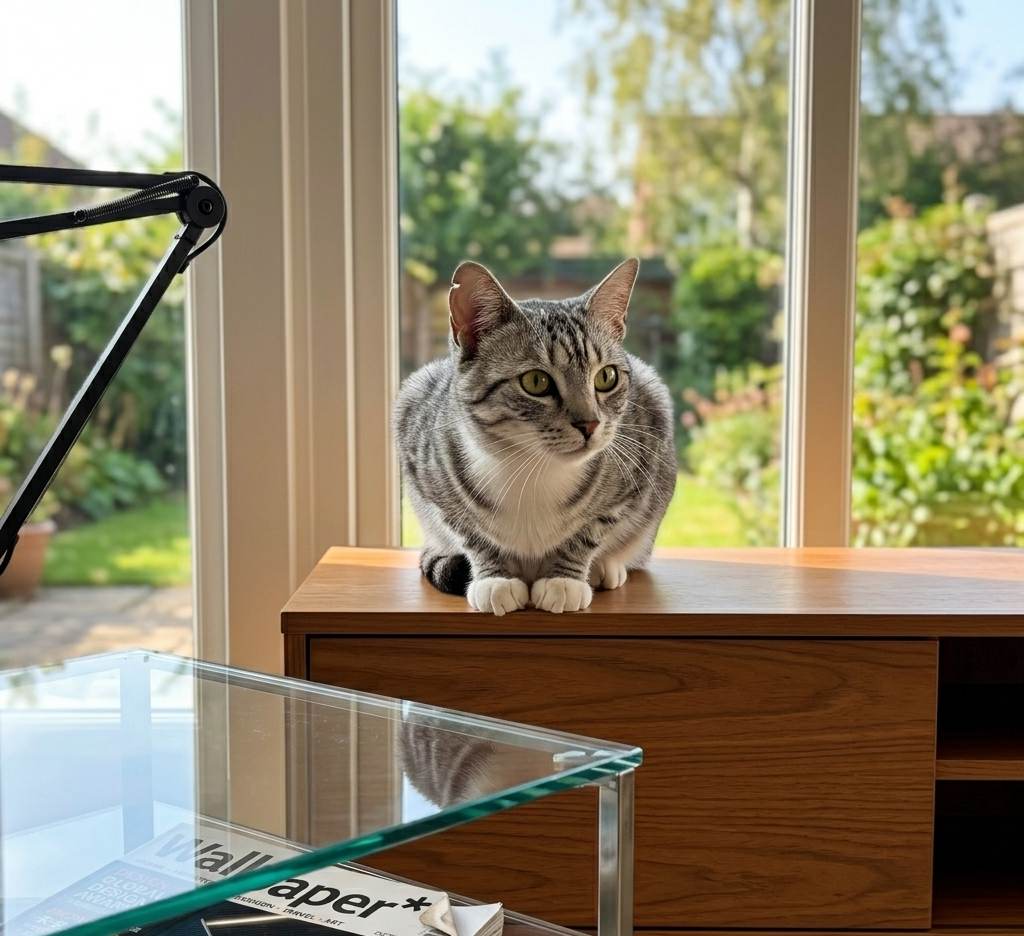}%
  }{%
    \fbox{\parbox[c][2.25cm][c]{0.86\linewidth}{\centering\scriptsize
    Memory image unavailable}}%
  }\\[0.45ex]
  \scriptsize\textbf{D21-001.png}\\[-0.2ex]foreground table
  \end{minipage}
  \begin{minipage}[t]{0.19\linewidth}\centering
  \IfFileExists{supplementary_images/event/images/pid_0000_task_0-18-0.jpg}{%
    \includegraphics[width=\linewidth,height=2.45cm,keepaspectratio]{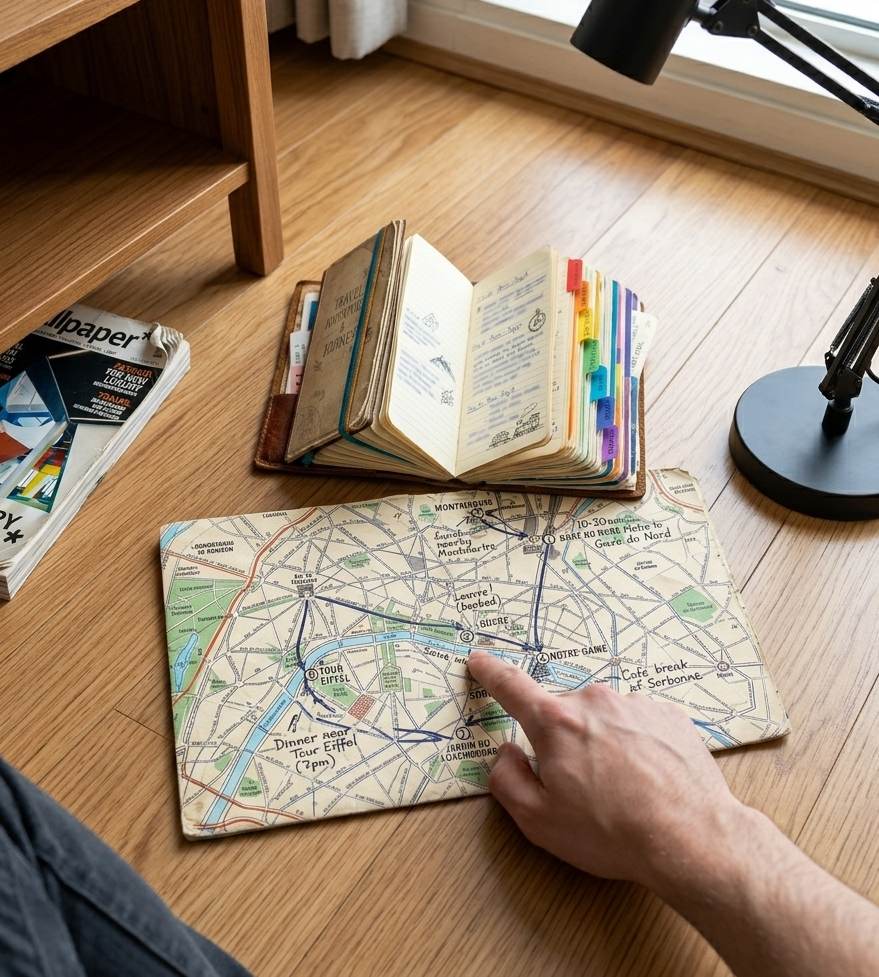}%
  }{%
    \fbox{\parbox[c][2.25cm][c]{0.86\linewidth}{\centering\scriptsize
    Memory image unavailable}}%
  }\\[0.45ex]
  \scriptsize\textbf{D28-001.png}\\[-0.2ex]route-planning floor
  \end{minipage}

\medskip
\textbf{Question.} Which image is the user's magazine?

\medskip
  \begin{minipage}[t]{0.235\linewidth}\centering
  \IfFileExists{supplementary_images/qa/entity_images/0-Items-6-img_identify_A.jpg}{%
    \includegraphics[width=\linewidth,height=2.65cm,keepaspectratio]{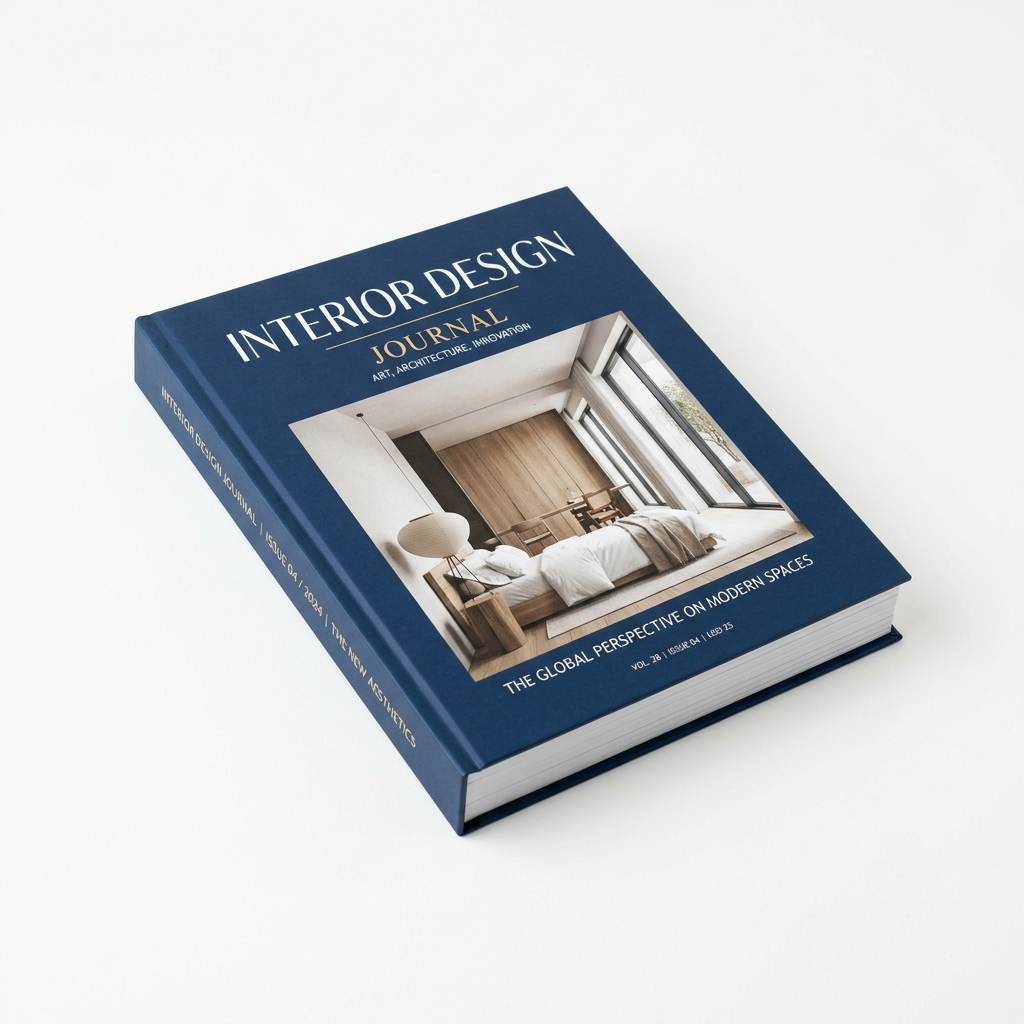}%
  }{%
    \fbox{\parbox[c][2.45cm][c]{0.88\linewidth}{\centering\scriptsize
    Option image unavailable}}%
  }\\[0.45ex]
  \scriptsize\textbf{A}\enspace blue design journal
  \end{minipage}
  \begin{minipage}[t]{0.235\linewidth}\centering
  \IfFileExists{supplementary_images/profile/generated_portraits/profile_0_items_wallpaper_magazine_12.jpg}{%
    \includegraphics[width=\linewidth,height=2.65cm,keepaspectratio]{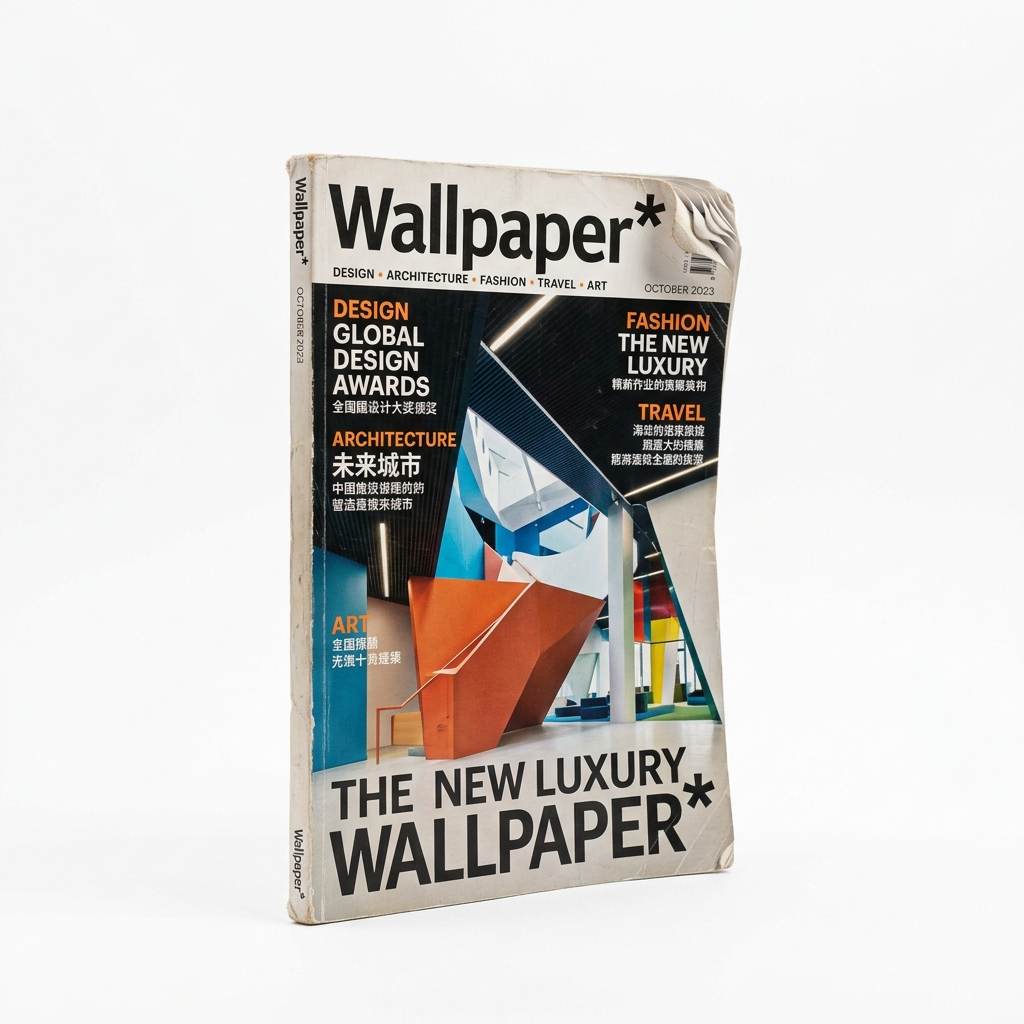}%
  }{%
    \fbox{\parbox[c][2.45cm][c]{0.88\linewidth}{\centering\scriptsize
    Option image unavailable}}%
  }\\[0.45ex]
  \scriptsize\textbf{B}\enspace dog-eared \emph{Wallpaper*}
  \end{minipage}
  \begin{minipage}[t]{0.235\linewidth}\centering
  \IfFileExists{supplementary_images/qa/entity_images/0-Items-6-img_identify_C.jpg}{%
    \includegraphics[width=\linewidth,height=2.65cm,keepaspectratio]{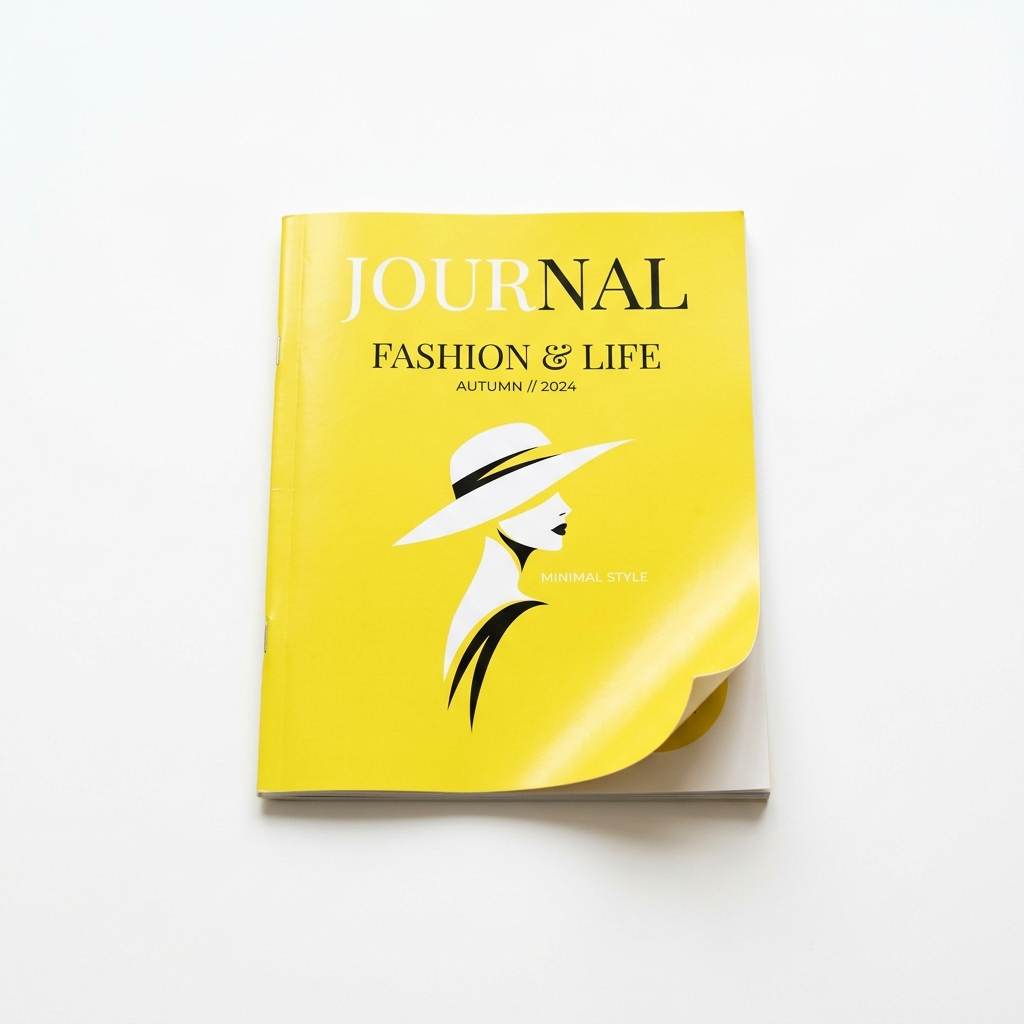}%
  }{%
    \fbox{\parbox[c][2.45cm][c]{0.88\linewidth}{\centering\scriptsize
    Option image unavailable}}%
  }\\[0.45ex]
  \scriptsize\textbf{C}\enspace yellow fashion magazine
  \end{minipage}
  \begin{minipage}[t]{0.235\linewidth}\centering
  \IfFileExists{supplementary_images/qa/entity_images/0-Items-6-img_identify_D.jpg}{%
    \includegraphics[width=\linewidth,height=2.65cm,keepaspectratio]{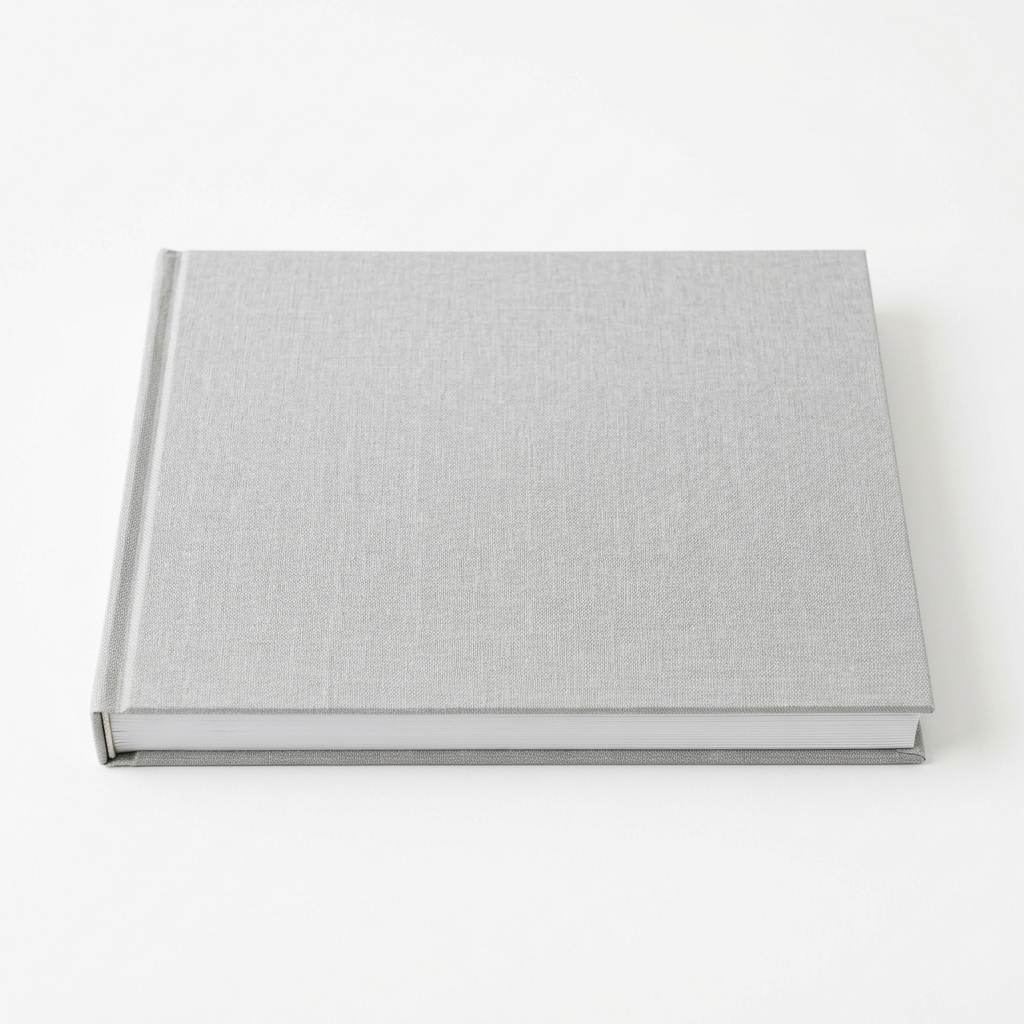}%
  }{%
    \fbox{\parbox[c][2.45cm][c]{0.88\linewidth}{\centering\scriptsize
    Option image unavailable}}%
  }\\[0.45ex]
  \scriptsize\textbf{D}\enspace gray landscape art book
  \end{minipage}

\medskip
\textbf{Gold answer: B.}\quad The exact item must be recognized from a
recurring peripheral object rather than from the foreground dialogue, so the
evidence is implicit.
\end{mybox}
\caption{Complete implicit Entity Recall example.}
\label{fig:example-entity-implicit}
\end{figure*}

\begin{figure*}[!htbp]
\centering
\begin{mybox}
\textbf{Long Pattern---Explicit Evidence}
\hfill \textbf{p1 / 1-WorkAndLearning-0}

\medskip
\textbf{Memory clues.}
Across \emph{D04} and \emph{D25}, the user repeatedly and directly describes
the same library corner: a fixed window seat used for nearly four years,
tree-filtered daylight, and an outdoor canopy that helps her regain focus
during deep reading and writing.

\medskip
\fbox{\parbox[c][1.45cm][c]{0.96\linewidth}{\centering
\textbf{Dialogue and speech evidence}\\[0.25ex]
\emph{D04:00}: ``Why did I insist on this seat?''\quad
\emph{D04-001--006.wav}: fixed corner, framed tree, soft light, higher focus\\
\emph{D25:00}: wind through plane-tree leaves\quad
\emph{D25-001--007.wav}: tree-facing seat, four-year habit, writing anchor}}

\medskip
\textbf{Question.} Which photograph best matches the user's habitual learning
environment?

\medskip
  \begin{minipage}[t]{0.235\linewidth}\centering
  \IfFileExists{supplementary_images/qa/pref_images/1-WorkAndLearning-0_A.jpg}{%
    \includegraphics[width=\linewidth,height=2.65cm,keepaspectratio]{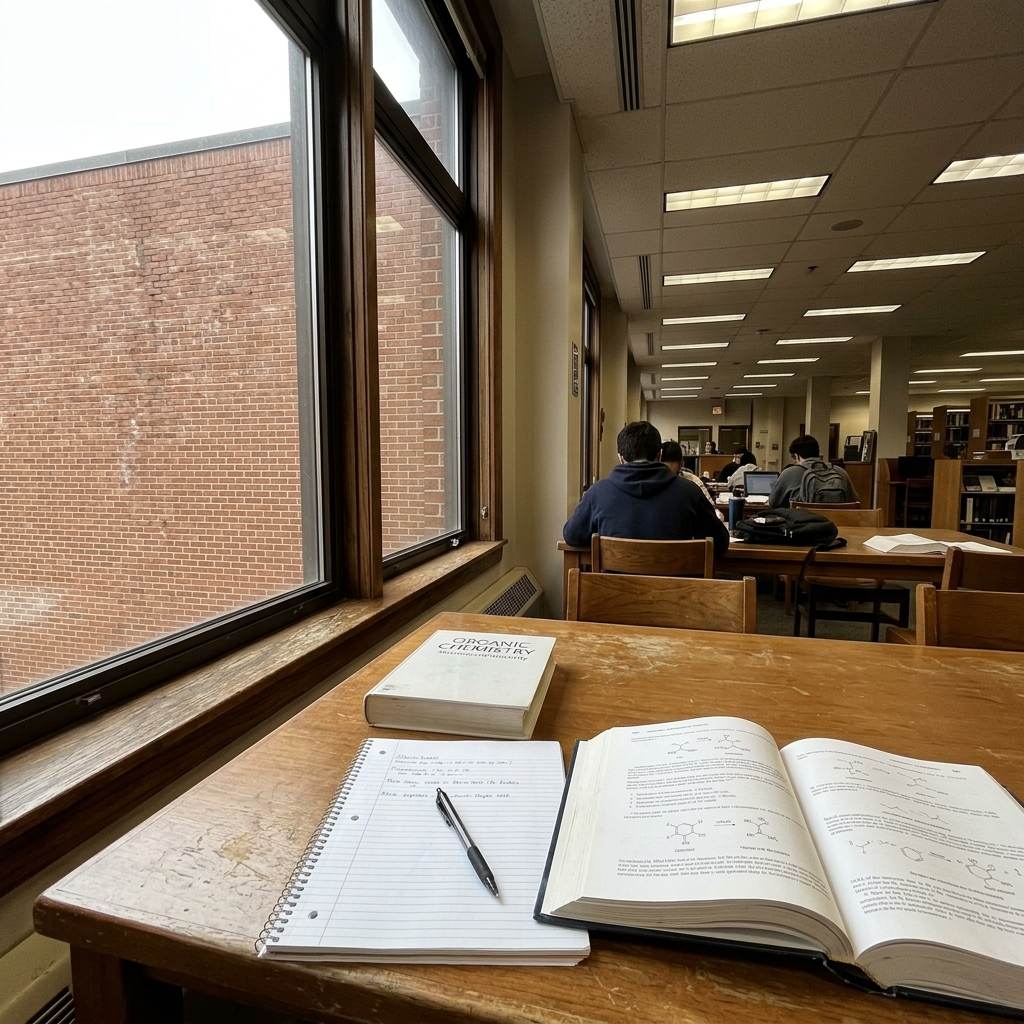}%
  }{%
    \fbox{\parbox[c][2.45cm][c]{0.88\linewidth}{\centering\scriptsize
    Option image unavailable}}%
  }\\[0.45ex]
  \scriptsize\textbf{A}\enspace window facing a brick wall
  \end{minipage}
  \begin{minipage}[t]{0.235\linewidth}\centering
  \IfFileExists{supplementary_images/qa/pref_images/1-WorkAndLearning-0_B.jpg}{%
    \includegraphics[width=\linewidth,height=2.65cm,keepaspectratio]{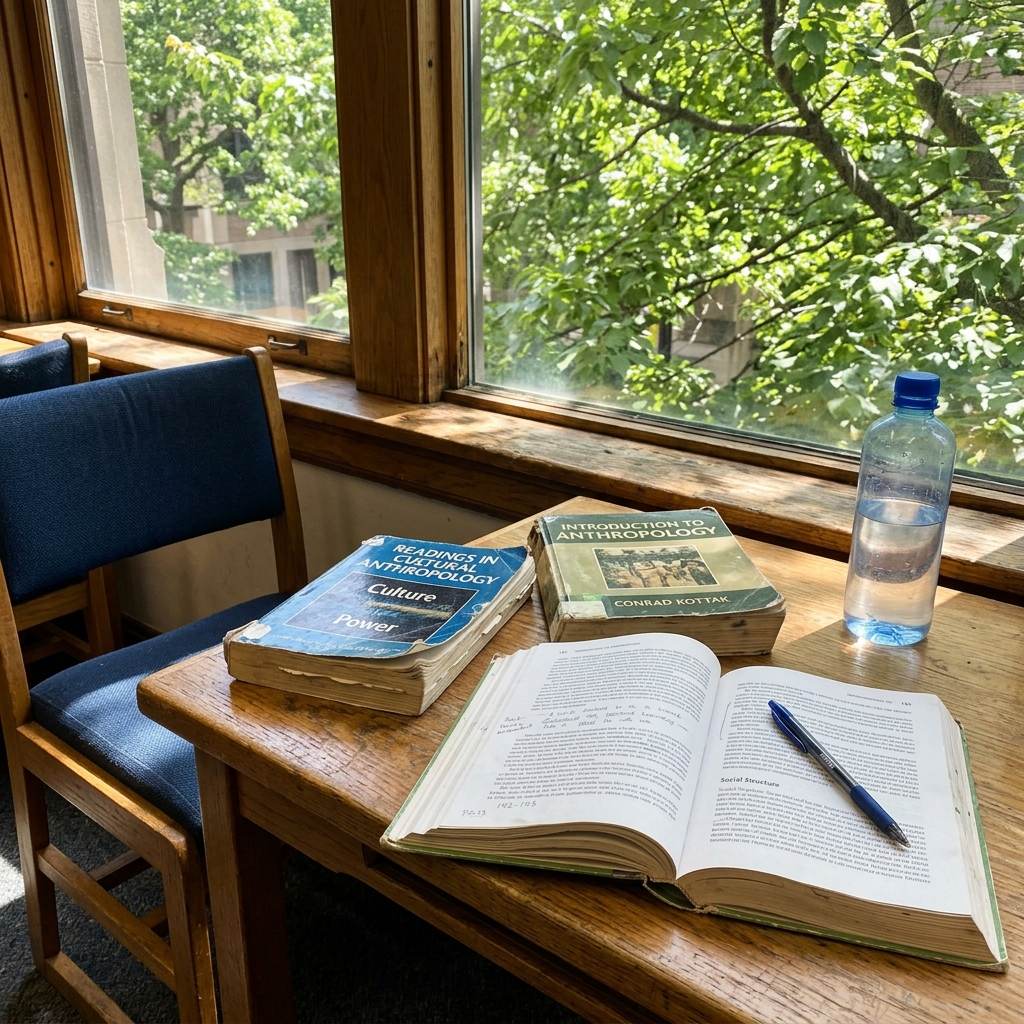}%
  }{%
    \fbox{\parbox[c][2.45cm][c]{0.88\linewidth}{\centering\scriptsize
    Option image unavailable}}%
  }\\[0.45ex]
  \scriptsize\textbf{B}\enspace tree-view library corner
  \end{minipage}
  \begin{minipage}[t]{0.235\linewidth}\centering
  \IfFileExists{supplementary_images/qa/pref_images/1-WorkAndLearning-0_C.jpg}{%
    \includegraphics[width=\linewidth,height=2.65cm,keepaspectratio]{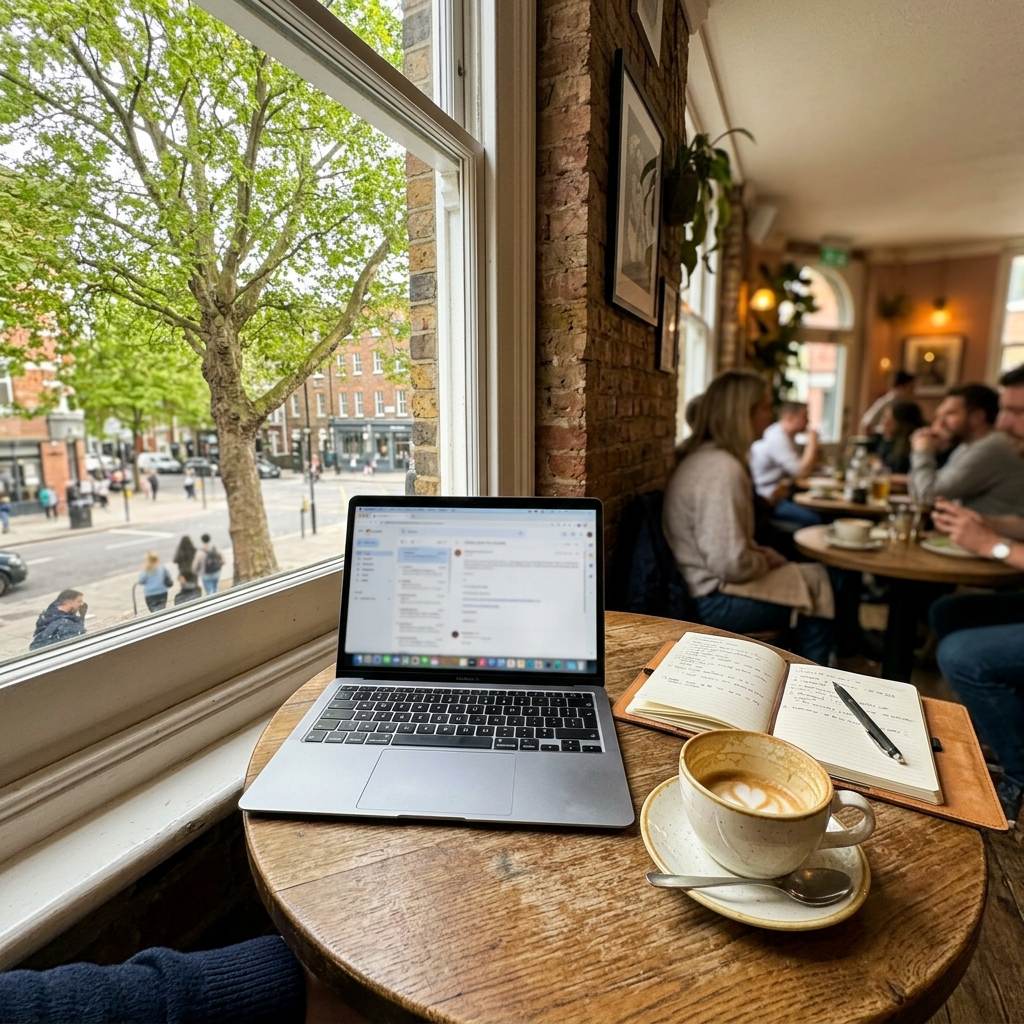}%
  }{%
    \fbox{\parbox[c][2.45cm][c]{0.88\linewidth}{\centering\scriptsize
    Option image unavailable}}%
  }\\[0.45ex]
  \scriptsize\textbf{C}\enspace window seat in a café
  \end{minipage}
  \begin{minipage}[t]{0.235\linewidth}\centering
  \IfFileExists{supplementary_images/qa/pref_images/1-WorkAndLearning-0_D.jpg}{%
    \includegraphics[width=\linewidth,height=2.65cm,keepaspectratio]{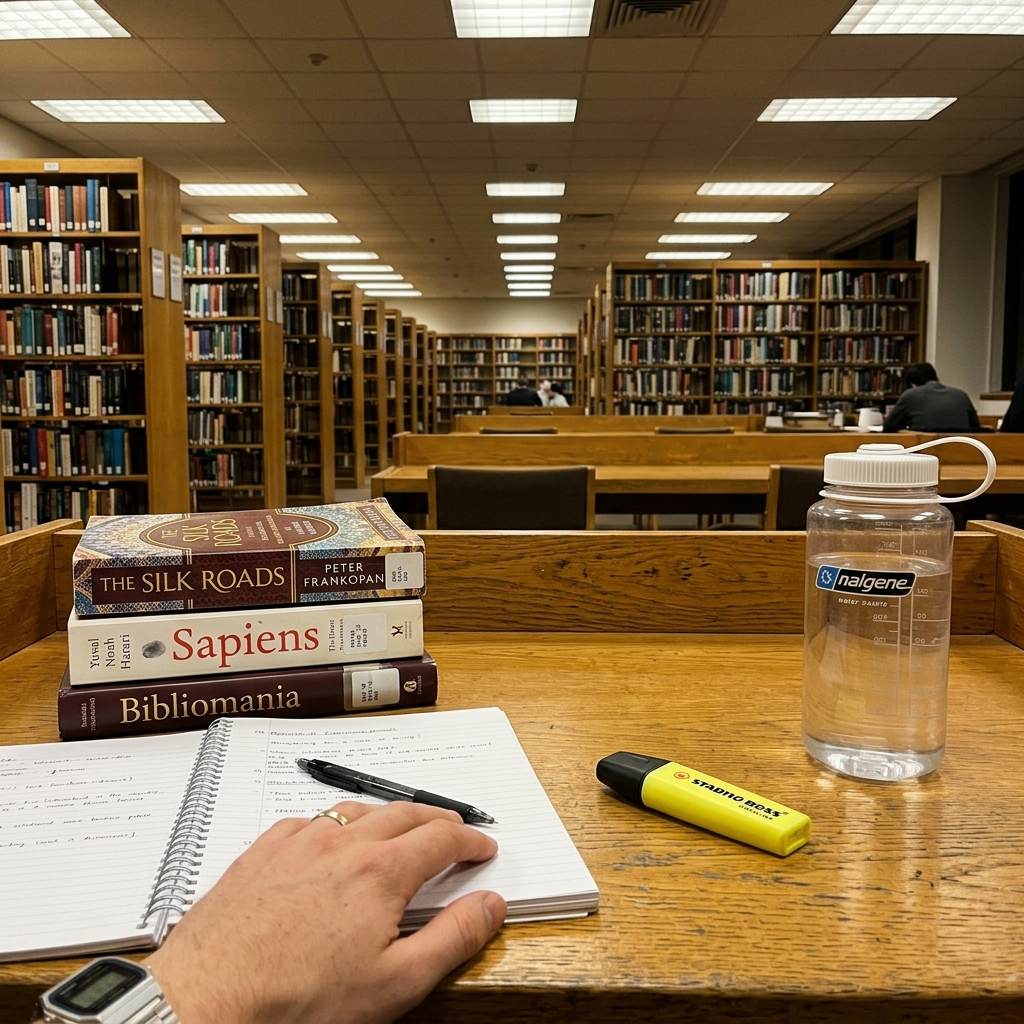}%
  }{%
    \fbox{\parbox[c][2.45cm][c]{0.88\linewidth}{\centering\scriptsize
    Option image unavailable}}%
  }\\[0.45ex]
  \scriptsize\textbf{D}\enspace interior library desk
  \end{minipage}

\medskip
\textbf{Gold answer: B.}\quad The fixed library location, tree view, and
lighting preference are repeatedly verbalized, making the pattern explicit.
\end{mybox}
\caption{Complete explicit Long Pattern example.}
\label{fig:example-pattern-explicit}
\end{figure*}

\begin{figure*}[!htbp]
\centering
\begin{mybox}
\textbf{Long Pattern---Implicit Audio Evidence}
\hfill \textbf{p2 / 2-FoodAndDrink-1}

\medskip
\textbf{Memory clues.}
Across \emph{D10}, \emph{D18}, and \emph{D27}, the foreground speech concerns
cycling, a mint plant, and the user's cat.  The target pattern---preparing
ingredients and stir-frying at home as a relaxing routine---is carried only
by recurring non-speech kitchen sounds in the mixed waveforms.

\medskip
\fbox{\parbox[c][1.45cm][c]{0.96\linewidth}{\centering
\textbf{Audio evidence}\\[0.25ex]
\emph{D10-001--006.wav}\quad$\bullet$\quad
\emph{D18-001--004.wav}\quad$\bullet$\quad
\emph{D27-001--008.wav}\\
\scriptsize target-bearing channel: ambient food preparation and cooking
sounds; the foreground speech does not state the cooking preference}}

\medskip
\textbf{Question.} Which image best reflects the user's everyday cooking
habit?

\medskip
  \begin{minipage}[t]{0.235\linewidth}\centering
  \IfFileExists{supplementary_images/qa/pref_images/2-FoodAndDrink-1_A.jpg}{%
    \includegraphics[width=\linewidth,height=2.65cm,keepaspectratio]{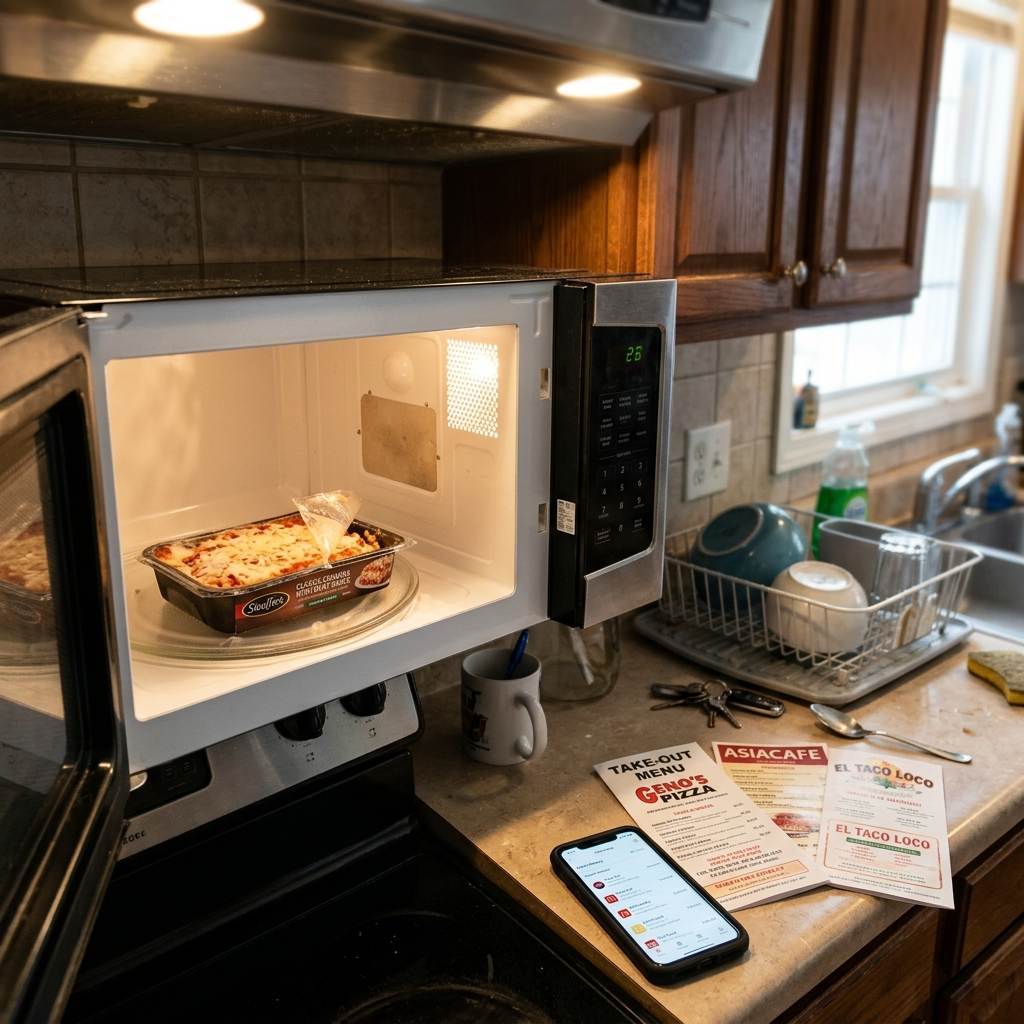}%
  }{%
    \fbox{\parbox[c][2.45cm][c]{0.88\linewidth}{\centering\scriptsize
    Option image unavailable}}%
  }\\[0.45ex]
  \scriptsize\textbf{A}\enspace microwaved frozen meal
  \end{minipage}
  \begin{minipage}[t]{0.235\linewidth}\centering
  \IfFileExists{supplementary_images/qa/pref_images/2-FoodAndDrink-1_B.jpg}{%
    \includegraphics[width=\linewidth,height=2.65cm,keepaspectratio]{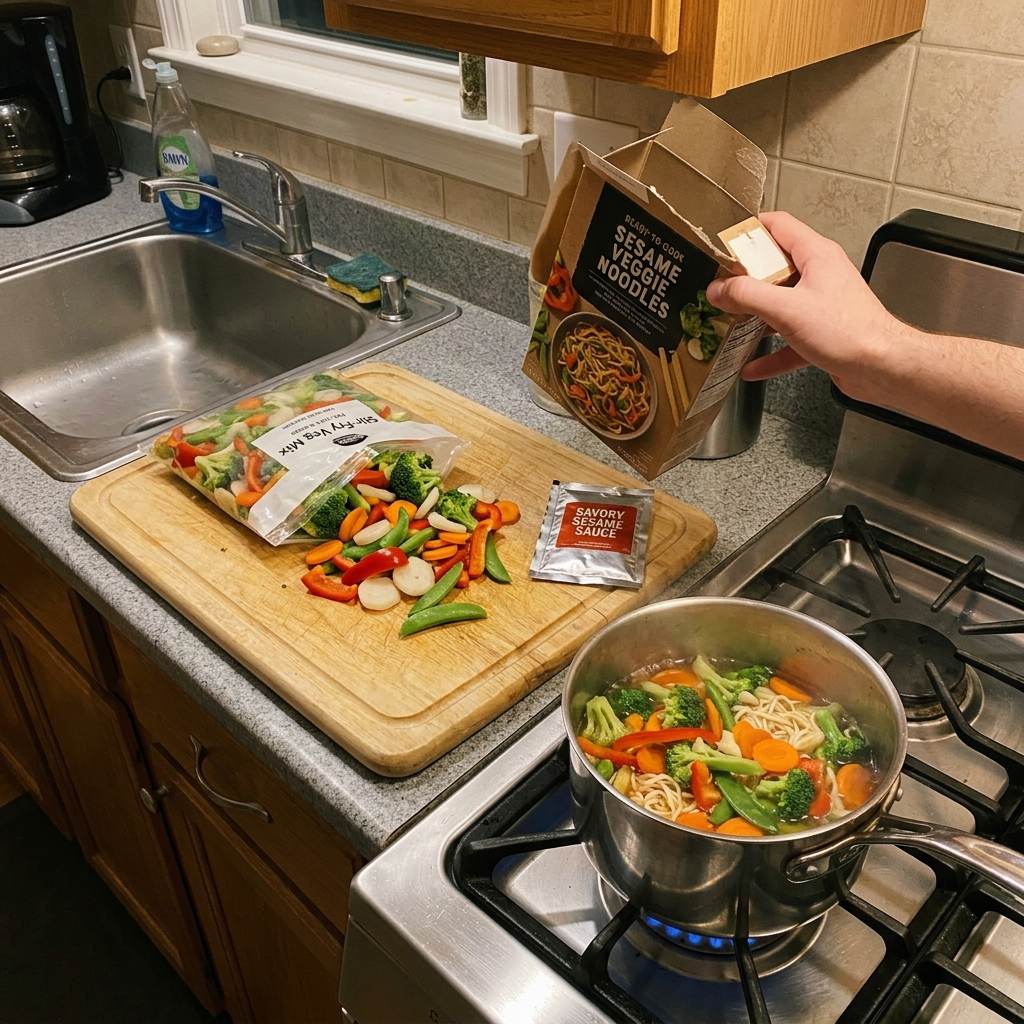}%
  }{%
    \fbox{\parbox[c][2.45cm][c]{0.88\linewidth}{\centering\scriptsize
    Option image unavailable}}%
  }\\[0.45ex]
  \scriptsize\textbf{B}\enspace pre-portioned meal kit
  \end{minipage}
  \begin{minipage}[t]{0.235\linewidth}\centering
  \IfFileExists{supplementary_images/qa/pref_images/2-FoodAndDrink-1_C.jpg}{%
    \includegraphics[width=\linewidth,height=2.65cm,keepaspectratio]{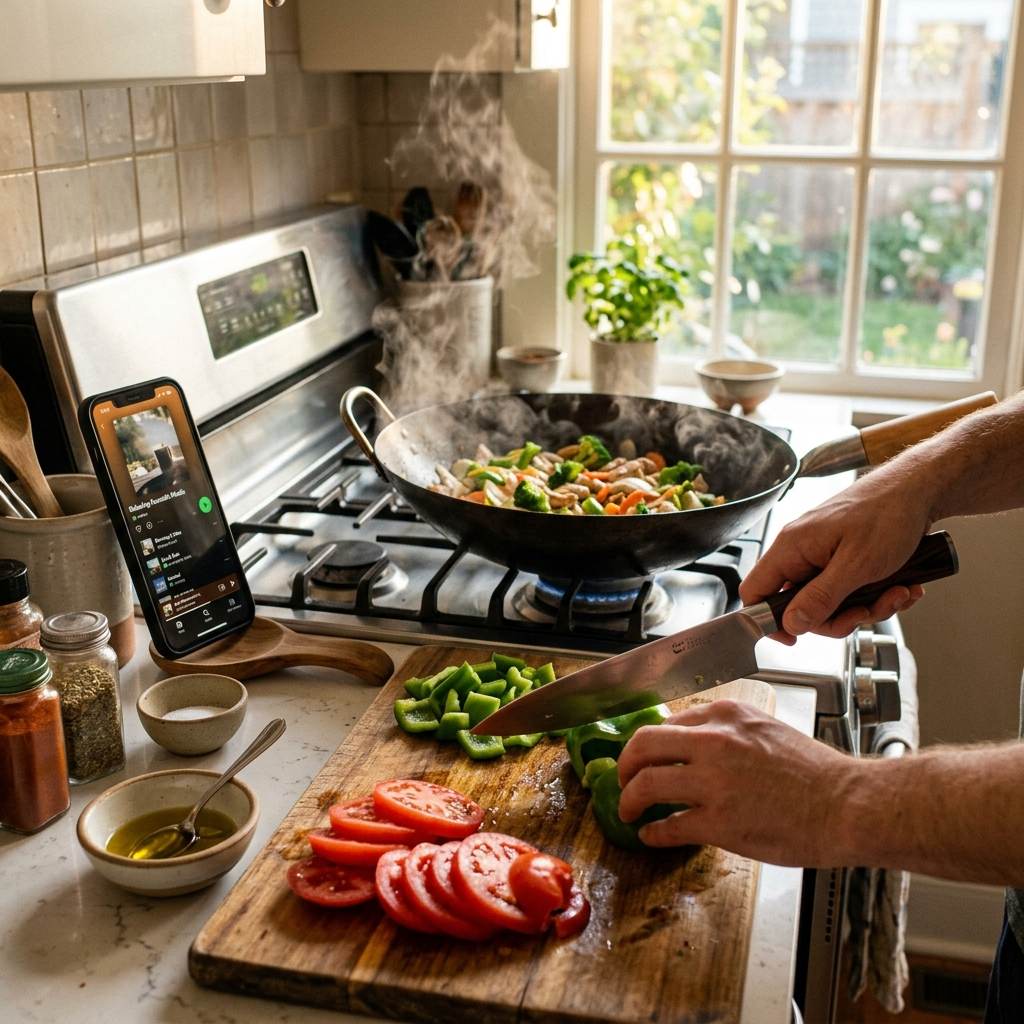}%
  }{%
    \fbox{\parbox[c][2.45cm][c]{0.88\linewidth}{\centering\scriptsize
    Option image unavailable}}%
  }\\[0.45ex]
  \scriptsize\textbf{C}\enspace fresh prep and stir-frying
  \end{minipage}
  \begin{minipage}[t]{0.235\linewidth}\centering
  \IfFileExists{supplementary_images/qa/pref_images/2-FoodAndDrink-1_D.jpg}{%
    \includegraphics[width=\linewidth,height=2.65cm,keepaspectratio]{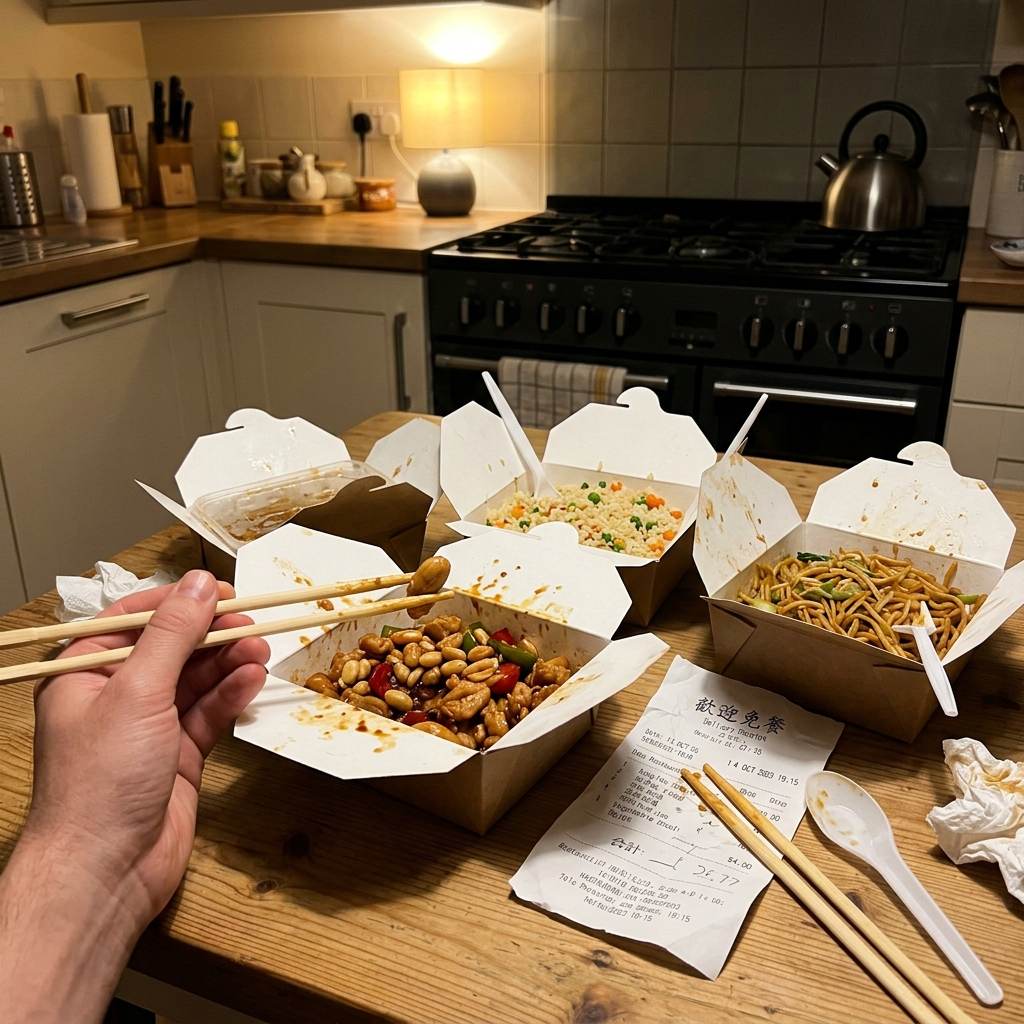}%
  }{%
    \fbox{\parbox[c][2.45cm][c]{0.88\linewidth}{\centering\scriptsize
    Option image unavailable}}%
  }\\[0.45ex]
  \scriptsize\textbf{D}\enspace delivered takeout
  \end{minipage}

\medskip
\textbf{Gold answer: C.}\quad The recurring cooking routine is recoverable
from ambient sound rather than from spoken content, so the evidence is
implicit.
\end{mybox}
\caption{Complete audio-supported implicit Long Pattern example.}
\label{fig:example-pattern-implicit}
\end{figure*}

\begin{figure*}[!htbp]
\centering
\begin{mybox}
\textbf{Personalized Recommendation---Explicit Evidence}
\hfill \textbf{p3 / 3-HobbiesAndEntertainment-0}

\medskip
\textbf{Memory clues.}
In \emph{D01}, \emph{D14}, and \emph{D20}, the user directly discusses
weekend calligraphy practice, copying model scripts, brush handling, ink, and
seal use.  The accompanying images place the calligraphy tools and written
sheets in the foreground.

\medskip
  \begin{minipage}[t]{0.31\linewidth}\centering
  \IfFileExists{supplementary_images/event/images/pid_0003_task_3-28-0.jpg}{%
    \includegraphics[width=\linewidth,height=3.1cm,keepaspectratio]{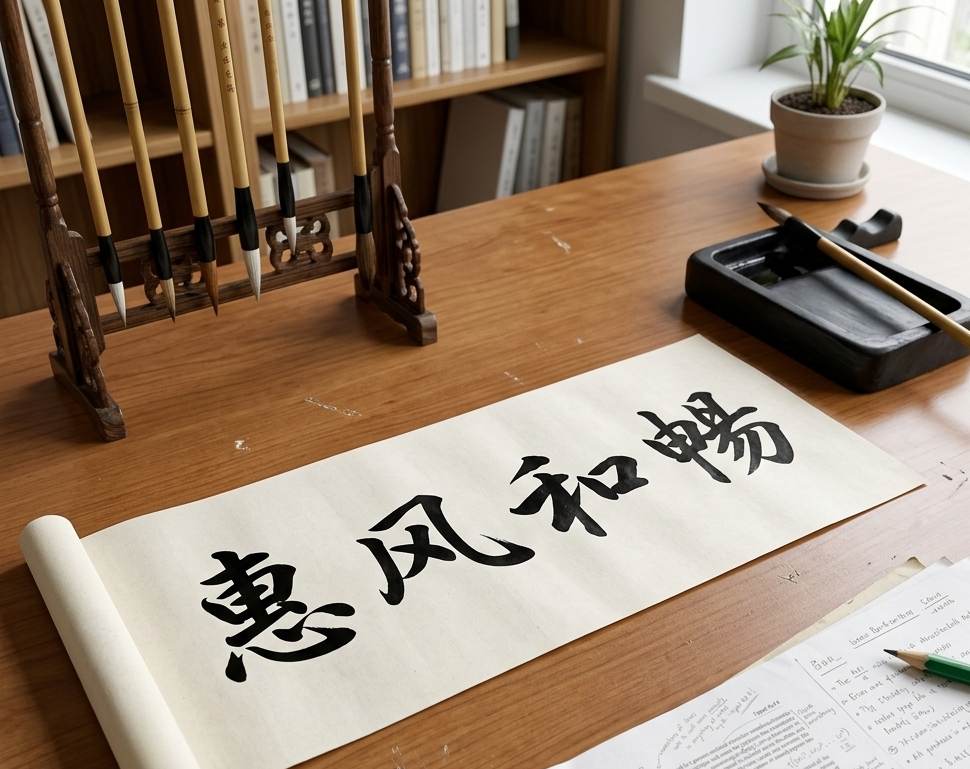}%
  }{%
    \fbox{\parbox[c][2.9cm][c]{0.88\linewidth}{\centering\scriptsize
    Original memory image\\not in transferred snapshot}}%
  }\\[0.45ex]
  \scriptsize\textbf{D01-001.png}\enspace calligraphy sheet and inkstone
  \end{minipage}
  \begin{minipage}[t]{0.31\linewidth}\centering
  \IfFileExists{supplementary_images/event/images/pid_0003_task_3-29-0.jpg}{%
    \includegraphics[width=\linewidth,height=3.1cm,keepaspectratio]{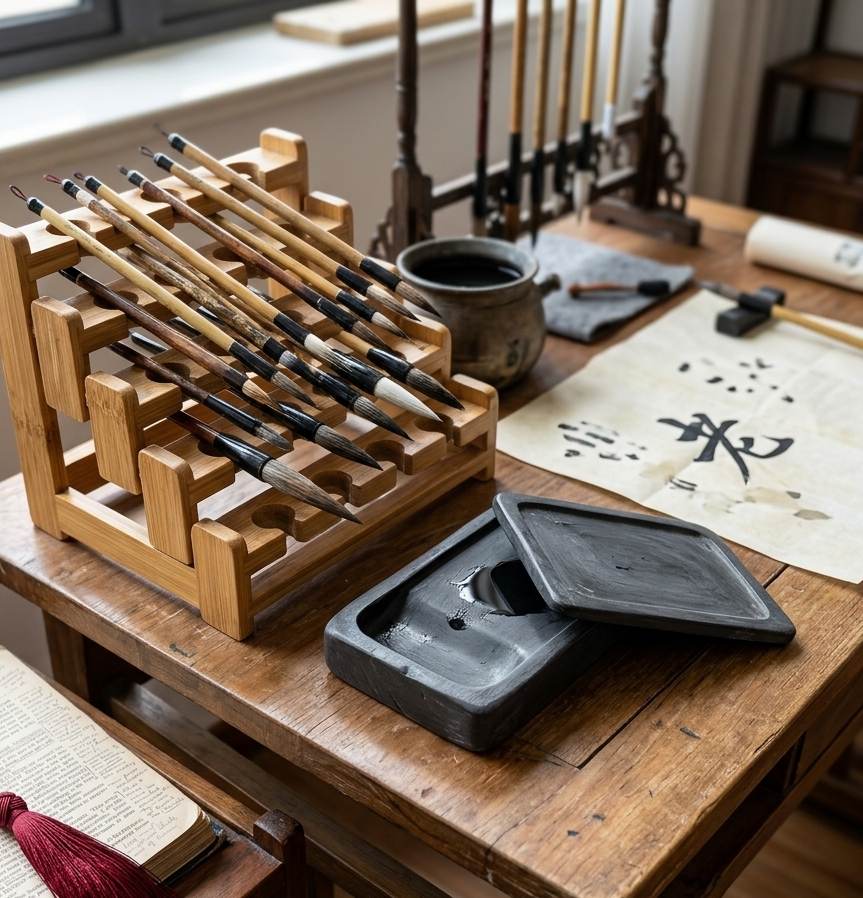}%
  }{%
    \fbox{\parbox[c][2.9cm][c]{0.88\linewidth}{\centering\scriptsize
    Original memory image\\not in transferred snapshot}}%
  }\\[0.45ex]
  \scriptsize\textbf{D14-001.png}\enspace brush rack and model script
  \end{minipage}
  \begin{minipage}[t]{0.31\linewidth}\centering
  \IfFileExists{supplementary_images/event/images/pid_0003_task_3-27-0.jpg}{%
    \includegraphics[width=\linewidth,height=3.1cm,keepaspectratio]{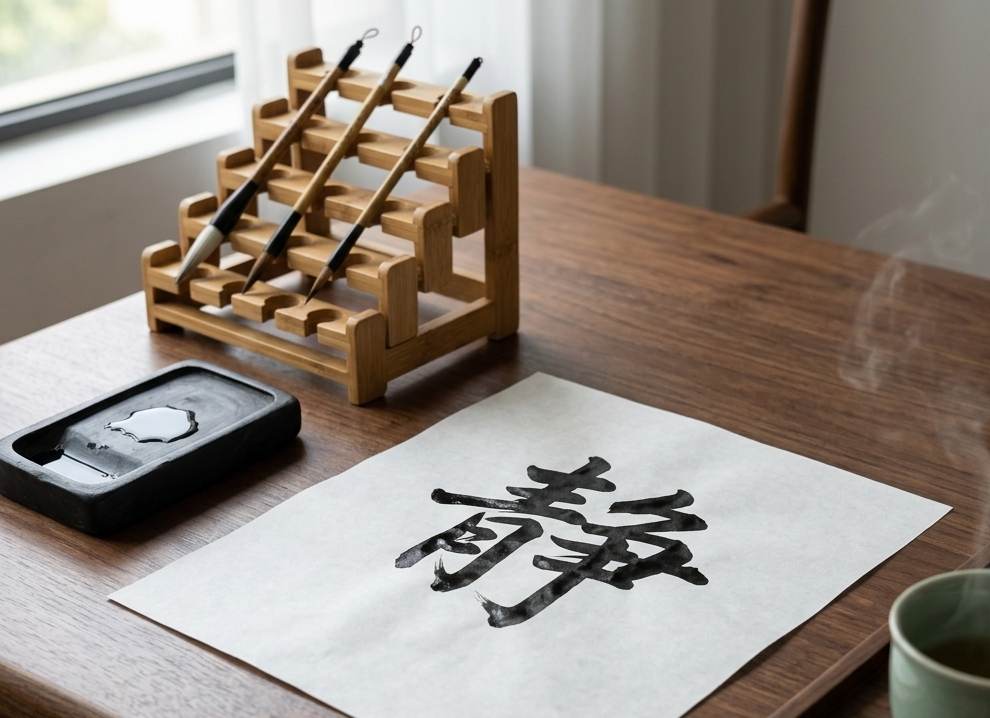}%
  }{%
    \fbox{\parbox[c][2.9cm][c]{0.88\linewidth}{\centering\scriptsize
    Original memory image\\not in transferred snapshot}}%
  }\\[0.45ex]
  \scriptsize\textbf{D20-001.png}\enspace weekend writing setup
  \end{minipage}

\medskip
\textbf{Question.} Which new weekend creative experience would the user most
likely appreciate?

\medskip
  \begin{minipage}[t]{0.235\linewidth}\centering
  \IfFileExists{supplementary_images/qa/rec_images/3-HobbiesAndEntertainment-0_A.jpg}{%
    \includegraphics[width=\linewidth,height=2.65cm,keepaspectratio]{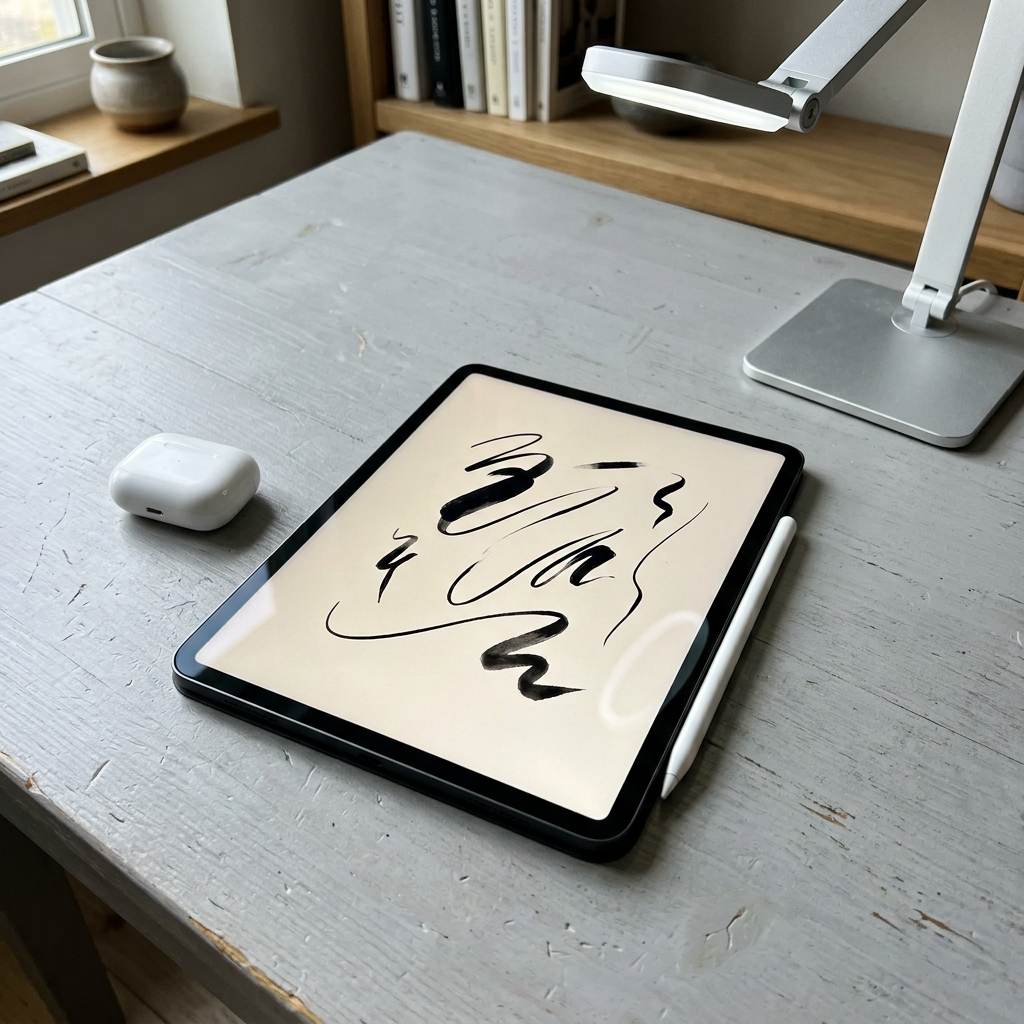}%
  }{%
    \fbox{\parbox[c][2.45cm][c]{0.88\linewidth}{\centering\scriptsize
    Option image unavailable}}%
  }\\[0.45ex]
  \scriptsize\textbf{A}\enspace digital brushwork
  \end{minipage}
  \begin{minipage}[t]{0.235\linewidth}\centering
  \IfFileExists{supplementary_images/qa/rec_images/3-HobbiesAndEntertainment-0_B.jpg}{%
    \includegraphics[width=\linewidth,height=2.65cm,keepaspectratio]{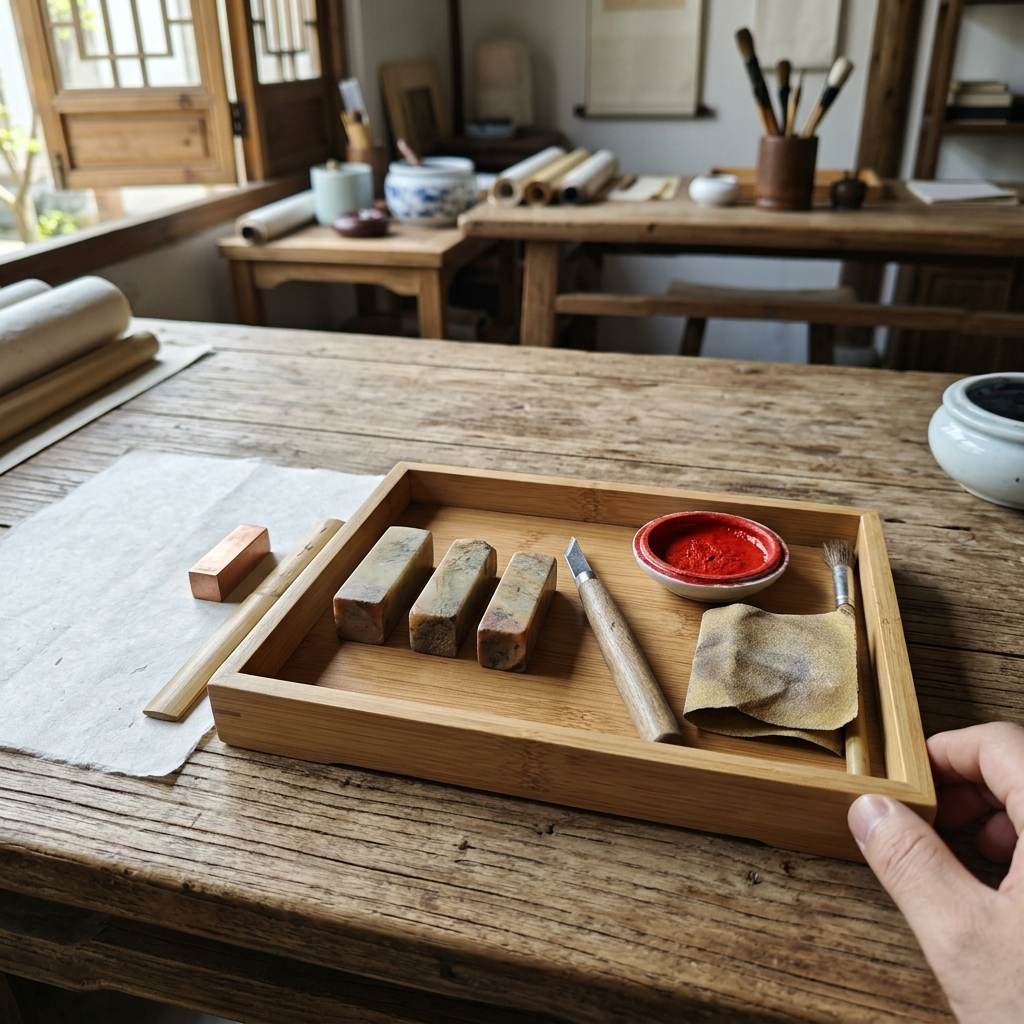}%
  }{%
    \fbox{\parbox[c][2.45cm][c]{0.88\linewidth}{\centering\scriptsize
    Option image unavailable}}%
  }\\[0.45ex]
  \scriptsize\textbf{B}\enspace seal carving
  \end{minipage}
  \begin{minipage}[t]{0.235\linewidth}\centering
  \IfFileExists{supplementary_images/qa/rec_images/3-HobbiesAndEntertainment-0_C.jpg}{%
    \includegraphics[width=\linewidth,height=2.65cm,keepaspectratio]{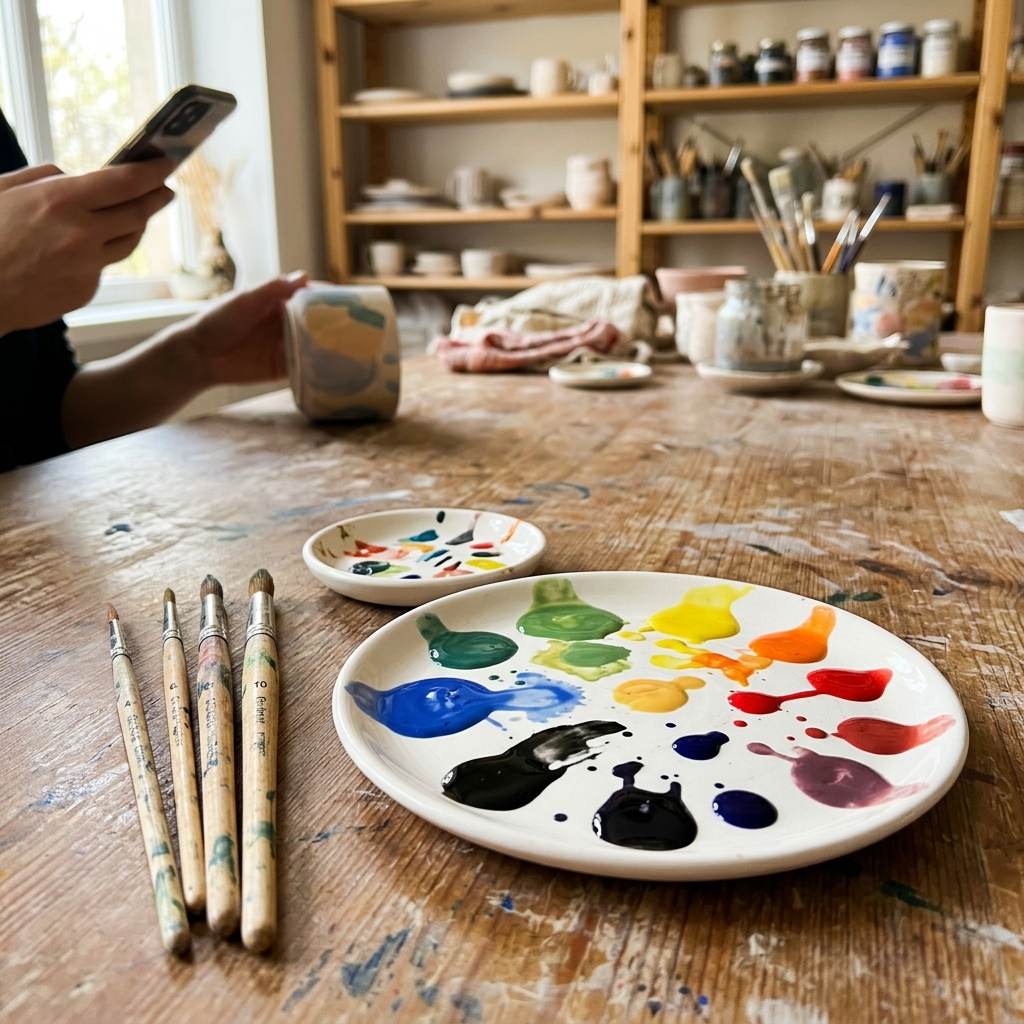}%
  }{%
    \fbox{\parbox[c][2.45cm][c]{0.88\linewidth}{\centering\scriptsize
    Option image unavailable}}%
  }\\[0.45ex]
  \scriptsize\textbf{C}\enspace pottery painting
  \end{minipage}
  \begin{minipage}[t]{0.235\linewidth}\centering
  \IfFileExists{supplementary_images/qa/rec_images/3-HobbiesAndEntertainment-0_D.jpg}{%
    \includegraphics[width=\linewidth,height=2.65cm,keepaspectratio]{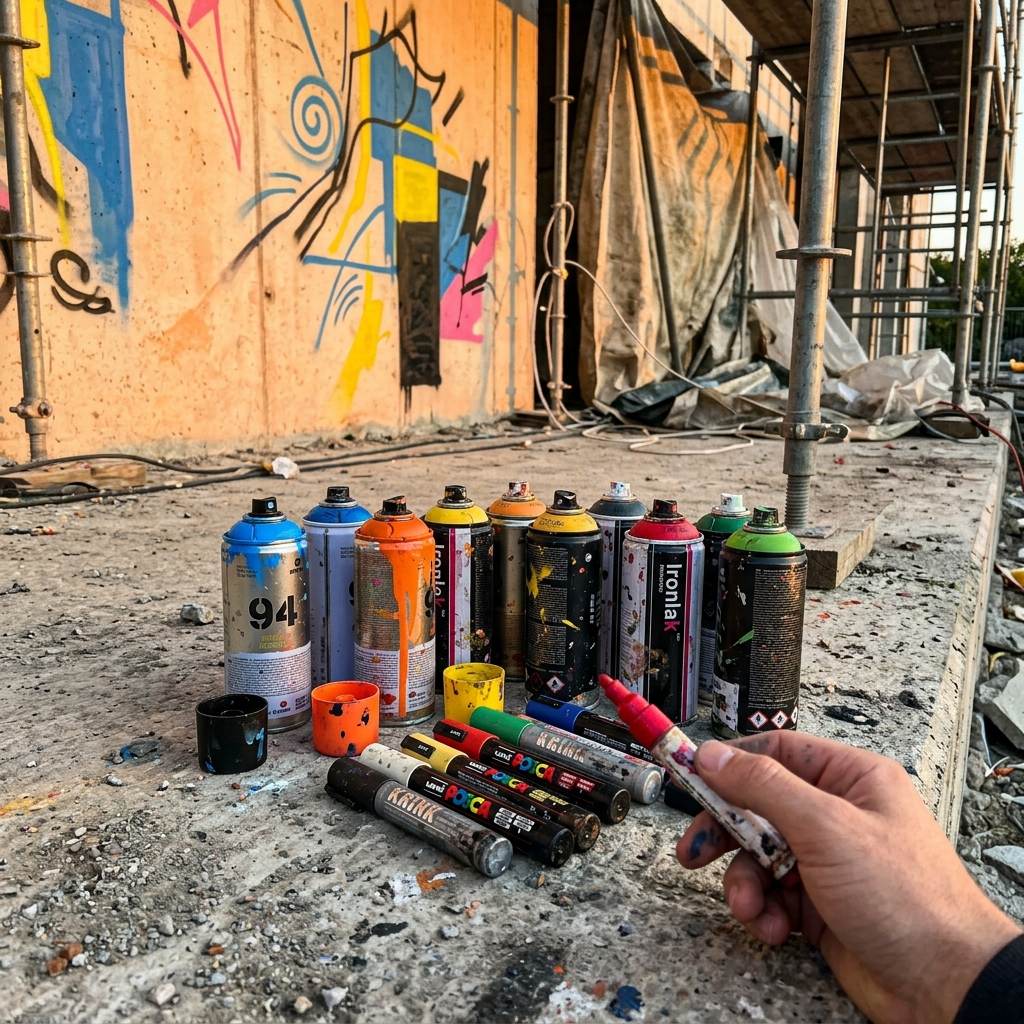}%
  }{%
    \fbox{\parbox[c][2.45cm][c]{0.88\linewidth}{\centering\scriptsize
    Option image unavailable}}%
  }\\[0.45ex]
  \scriptsize\textbf{D}\enspace spray painting
  \end{minipage}

\medskip
\textbf{Gold answer: B.}\quad Seal carving is new, but it transfers directly
from the explicitly established calligraphy-and-seal interest.
\end{mybox}
\caption{Complete explicit Personalized Recommendation example.}
\label{fig:example-rec-explicit}
\end{figure*}

\begin{figure*}[!htbp]
\centering
\begin{mybox}
\textbf{Personalized Recommendation---Implicit Image Evidence}
\hfill \textbf{p0 / 0-BodyAndHealth-1}

\medskip
\textbf{Memory clues.}
The retained visual clue set spans \emph{D01}, \emph{D10}, \emph{D18}, and
\emph{D29}; yoga or stretching equipment appears peripherally and is clearest
in \emph{D01}, \emph{D10}, and \emph{D29}.  The foreground conversations
concern hiking recovery, cats, and travel planning rather than stating a
preference for gentle at-home flexibility and relaxation practice.

\medskip
  \begin{minipage}[t]{0.19\linewidth}\centering
  \IfFileExists{supplementary_images/event/images/pid_0000_task_0-9-0.jpg}{%
    \includegraphics[width=\linewidth,height=2.45cm,keepaspectratio]{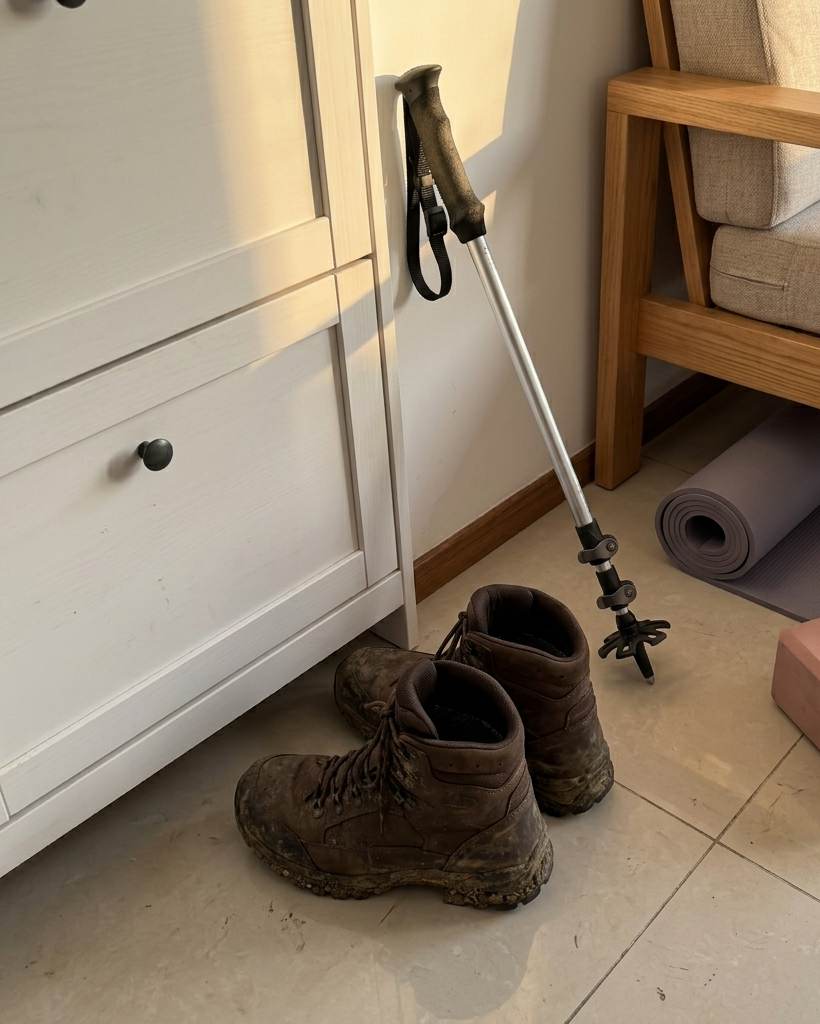}%
  }{%
    \fbox{\parbox[c][2.25cm][c]{0.86\linewidth}{\centering\scriptsize
    Memory image unavailable}}%
  }\\[0.45ex]
  \scriptsize\textbf{D01-001.png}\\[-0.2ex]hiking gear and yoga props
  \end{minipage}
  \begin{minipage}[t]{0.19\linewidth}\centering
  \IfFileExists{supplementary_images/event/images/pid_0000_task_0-1-0.jpg}{%
    \includegraphics[width=\linewidth,height=2.45cm,keepaspectratio]{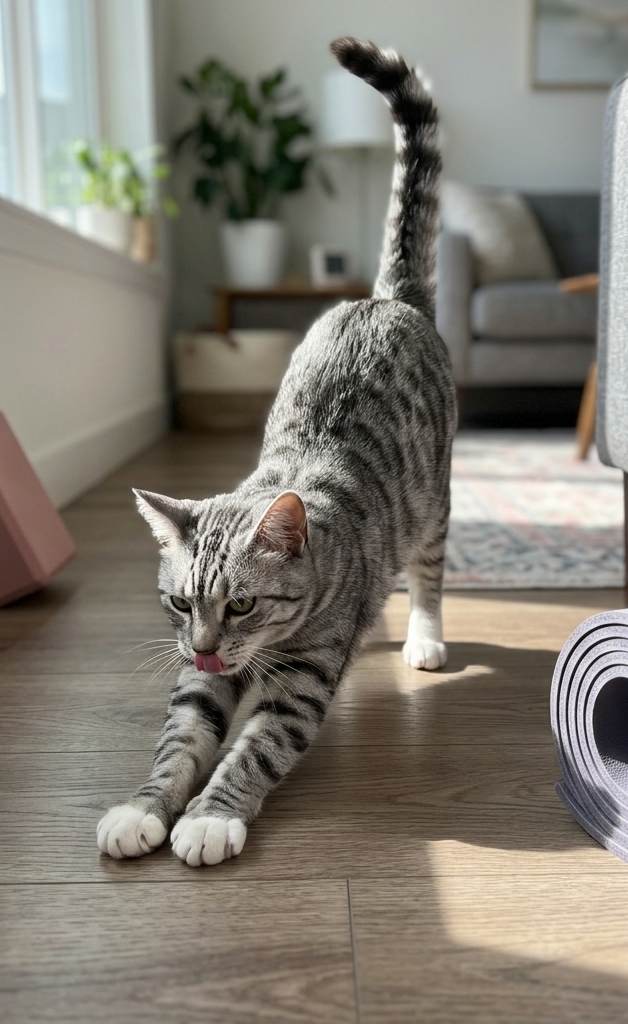}%
  }{%
    \fbox{\parbox[c][2.25cm][c]{0.86\linewidth}{\centering\scriptsize
    Memory image unavailable}}%
  }\\[0.45ex]
  \scriptsize\textbf{D10-001.png}\\[-0.2ex]mat beside stretching cat
  \end{minipage}
  \begin{minipage}[t]{0.19\linewidth}\centering
  \IfFileExists{supplementary_images/event/images/pid_0000_task_0-17-0.jpg}{%
    \includegraphics[width=\linewidth,height=2.45cm,keepaspectratio]{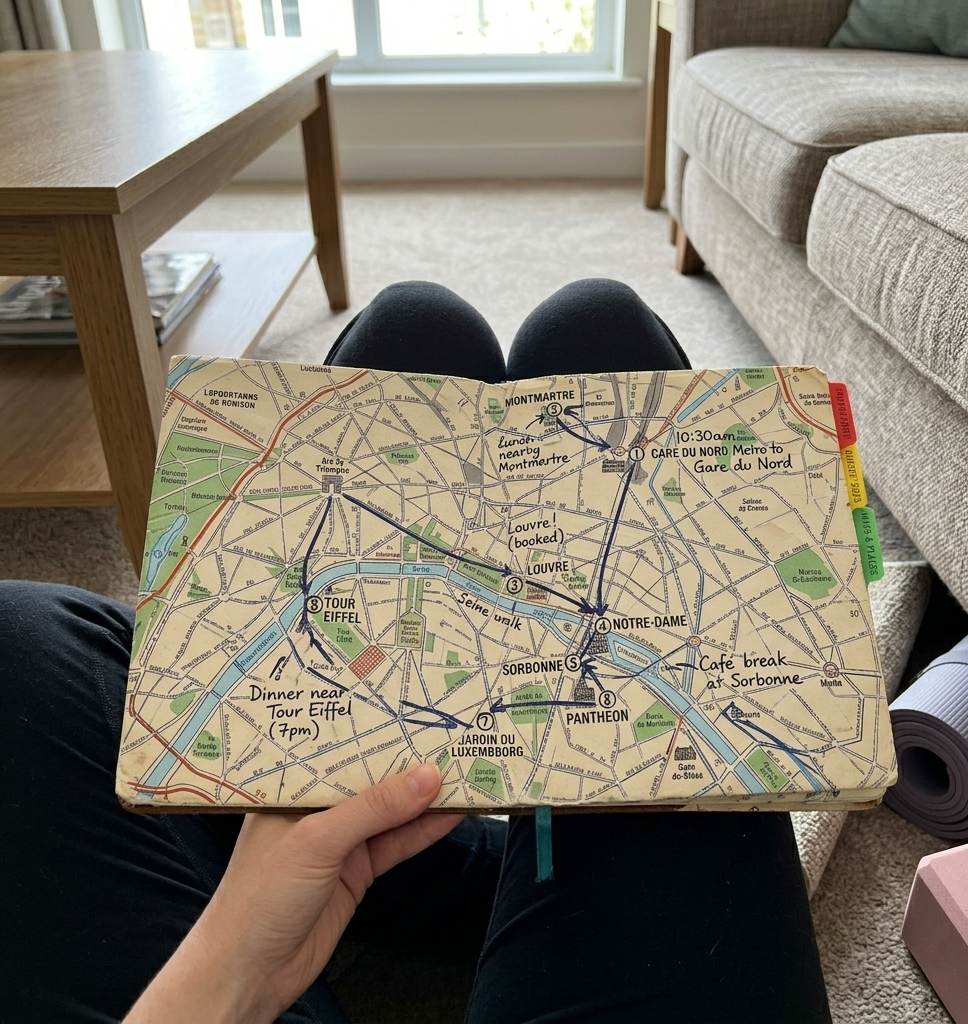}%
  }{%
    \fbox{\parbox[c][2.25cm][c]{0.86\linewidth}{\centering\scriptsize
    Memory image unavailable}}%
  }\\[0.45ex]
  \scriptsize\textbf{D18-001.png}\\[-0.2ex]retained travel-planning clue
  \end{minipage}
  \begin{minipage}[t]{0.19\linewidth}\centering
  \IfFileExists{supplementary_images/event/images/pid_0000_task_0-5-0.jpg}{%
    \includegraphics[width=\linewidth,height=2.45cm,keepaspectratio]{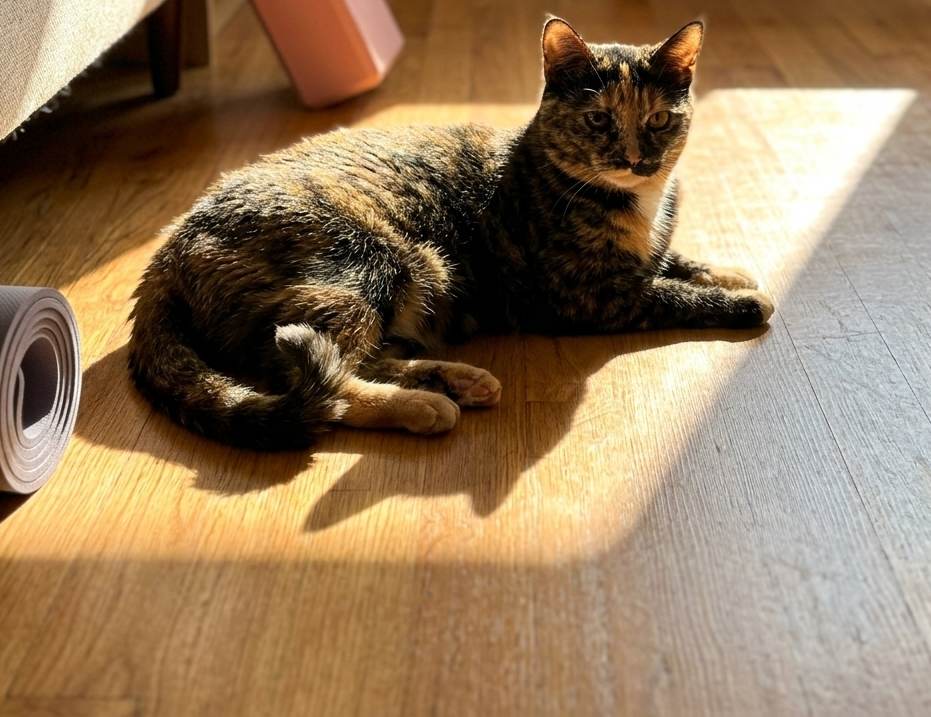}%
  }{%
    \fbox{\parbox[c][2.25cm][c]{0.86\linewidth}{\centering\scriptsize
    Memory image unavailable}}%
  }\\[0.45ex]
  \scriptsize\textbf{D29-001.png}\\[-0.2ex]mat and yoga block
  \end{minipage}

\medskip
\textbf{Question.} Which new everyday mind--body activity would the user most
likely appreciate?

\medskip
  \begin{minipage}[t]{0.235\linewidth}\centering
  \IfFileExists{supplementary_images/qa/rec_images/0-BodyAndHealth-1_A.jpg}{%
    \includegraphics[width=\linewidth,height=2.65cm,keepaspectratio]{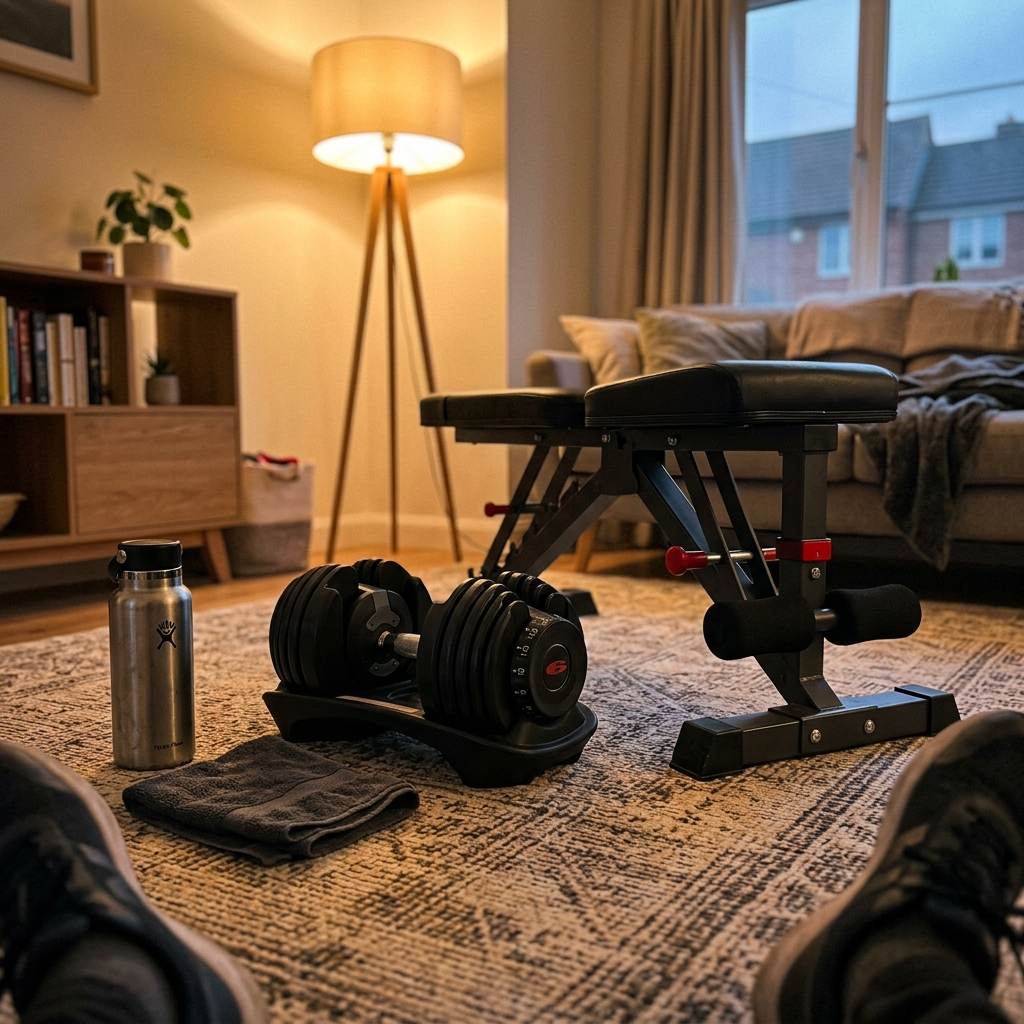}%
  }{%
    \fbox{\parbox[c][2.45cm][c]{0.88\linewidth}{\centering\scriptsize
    Option image unavailable}}%
  }\\[0.45ex]
  \scriptsize\textbf{A}\enspace adjustable weights
  \end{minipage}
  \begin{minipage}[t]{0.235\linewidth}\centering
  \IfFileExists{supplementary_images/qa/rec_images/0-BodyAndHealth-1_B.jpg}{%
    \includegraphics[width=\linewidth,height=2.65cm,keepaspectratio]{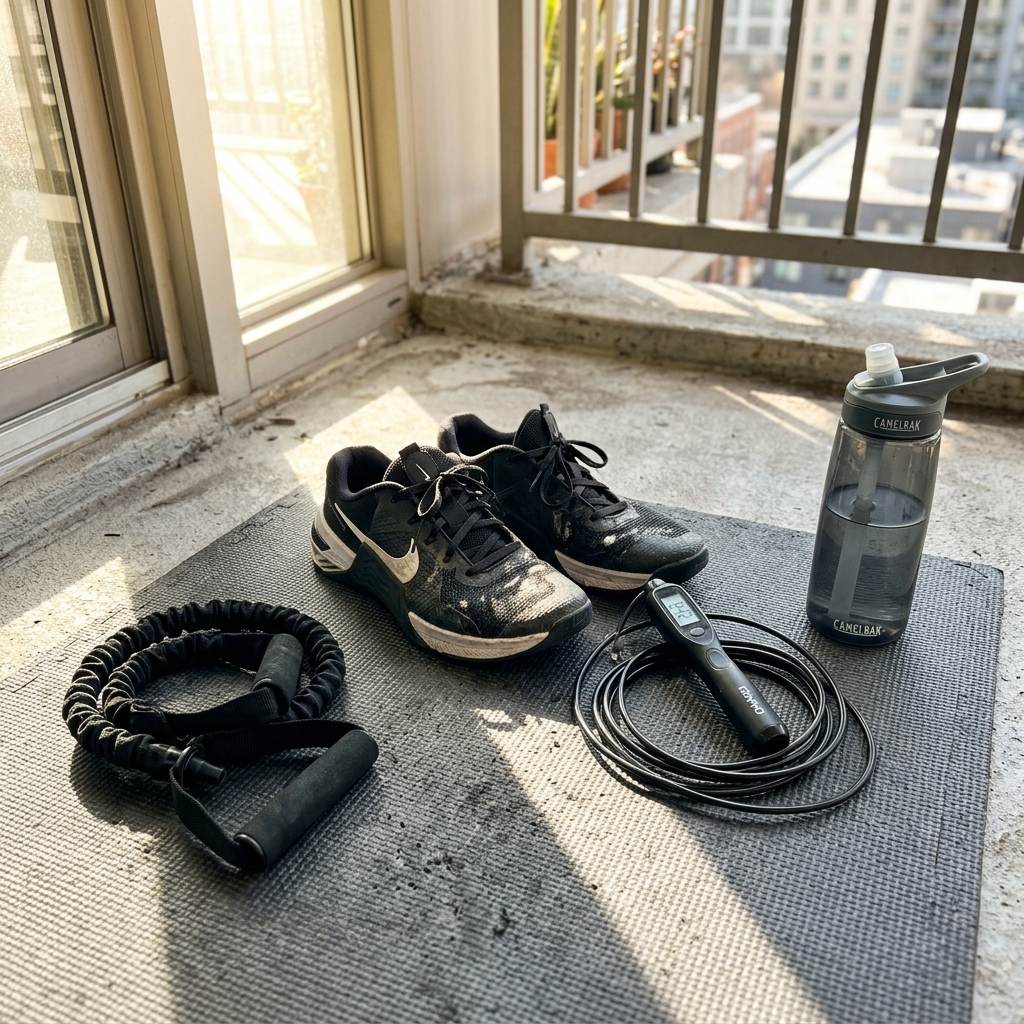}%
  }{%
    \fbox{\parbox[c][2.45cm][c]{0.88\linewidth}{\centering\scriptsize
    Option image unavailable}}%
  }\\[0.45ex]
  \scriptsize\textbf{B}\enspace high-intensity skipping
  \end{minipage}
  \begin{minipage}[t]{0.235\linewidth}\centering
  \IfFileExists{supplementary_images/qa/rec_images/0-BodyAndHealth-1_C.jpg}{%
    \includegraphics[width=\linewidth,height=2.65cm,keepaspectratio]{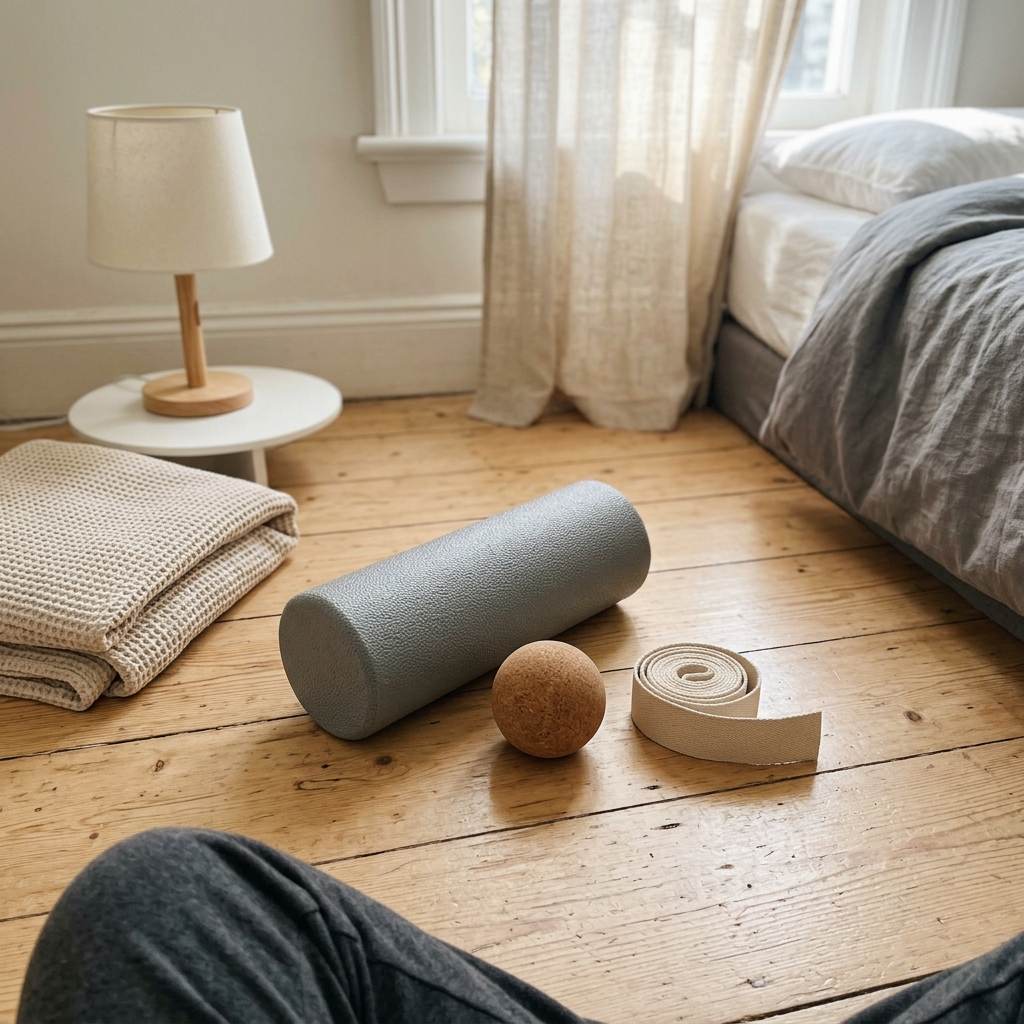}%
  }{%
    \fbox{\parbox[c][2.45cm][c]{0.88\linewidth}{\centering\scriptsize
    Option image unavailable}}%
  }\\[0.45ex]
  \scriptsize\textbf{C}\enspace gentle myofascial release
  \end{minipage}
  \begin{minipage}[t]{0.235\linewidth}\centering
  \IfFileExists{supplementary_images/qa/rec_images/0-BodyAndHealth-1_D.jpg}{%
    \includegraphics[width=\linewidth,height=2.65cm,keepaspectratio]{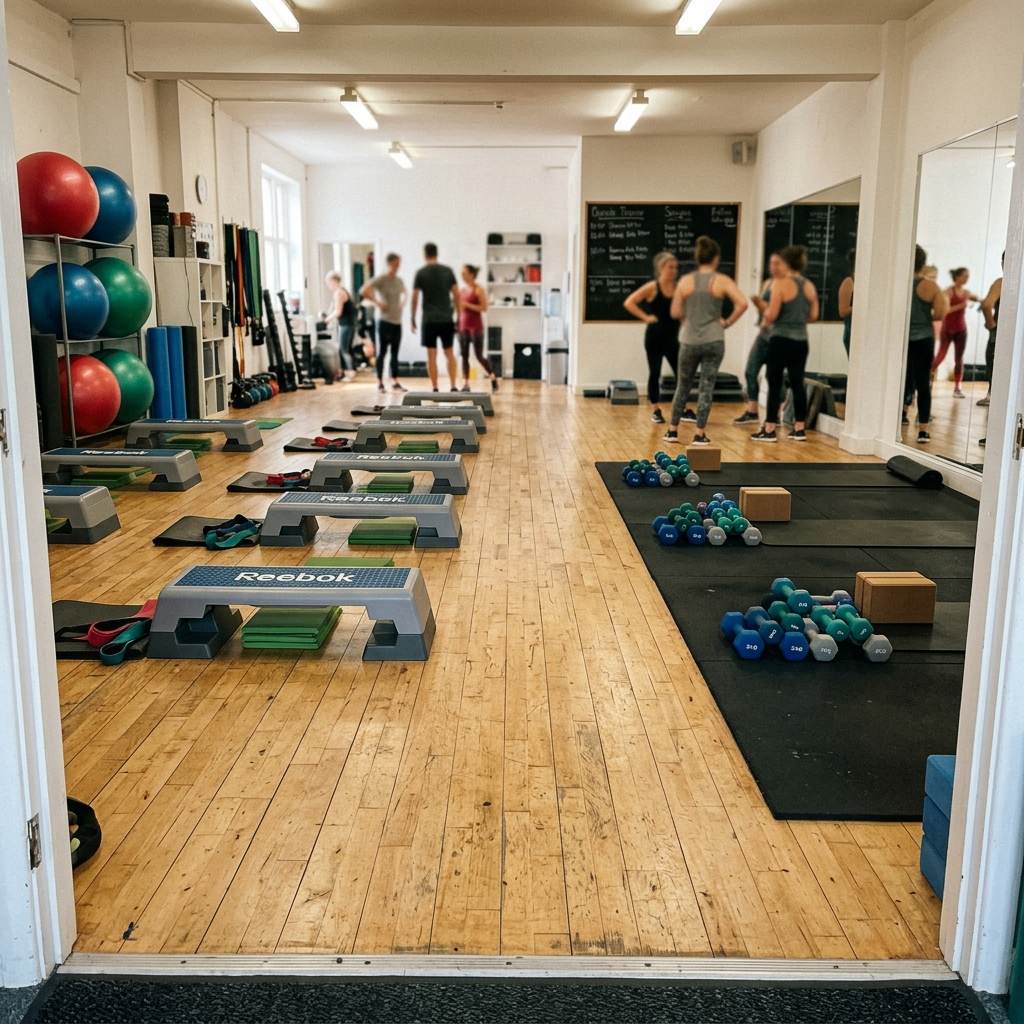}%
  }{%
    \fbox{\parbox[c][2.45cm][c]{0.88\linewidth}{\centering\scriptsize
    Option image unavailable}}%
  }\\[0.45ex]
  \scriptsize\textbf{D}\enspace group fitness class
  \end{minipage}

\medskip
\textbf{Gold answer: C.}\quad The recommendation transfers the inferred
preference for quiet, low-load, at-home flexibility work to a new but
compatible recovery activity.
\end{mybox}
\caption{Complete image-supported implicit Personalized Recommendation example.}
\label{fig:example-rec-implicit}
\end{figure*}

\begin{figure*}[!htbp]
\centering
\begin{mybox}
\textbf{Answer Refusal---Explicit Lure}
\hfill \textbf{p13 / 13-HomeAndSpace-0-adv0}

\medskip
\textbf{Memory clues.}
\emph{D03:04} states that the user waters balcony foliage every morning;
\emph{D18:03} describes checking soil and leaf tips; and \emph{D28:08}
describes misting pennywort daily.  None of these memories occurs while the
user is interacting with Yumo.

\medskip
\textbf{Question.} While interacting with her pet Yumo, what household habit
did Lin Huixian mention?

\medskip
\begin{tabular}{@{}p{0.48\linewidth}p{0.48\linewidth}@{}}
A. Daily cleaning and watering balcony foliage. &
B. The history does not specify.\\
C. Grooming Yumo and cleaning its litter box. &
D. Walking Yumo on the lawn each morning.
\end{tabular}

\medskip
\textbf{Gold answer: B.}\quad Option A is a true explicit memory, but the stem
moves it into an unsupported pet-interaction context.  This is an
\emph{event-context swap}.
\end{mybox}
\caption{Complete explicit-lure Answer Refusal example.}
\label{fig:example-refusal-explicit}
\end{figure*}

\begin{figure*}[!htbp]
\centering
\begin{mybox}
\textbf{Answer Refusal---Implicit Lure}
\hfill \textbf{p0 / 0-FoodAndDrink-1-adv2}

\medskip
\textbf{Lure-source memory clues.}
Recurring ambient kitchen sounds indicate an everyday pattern of preparing
ingredients and stir-frying at home.  In \emph{D00}, \emph{D19}, and
\emph{D25}, the foreground speech instead concerns a citrus diffuser,
bookshelf organization, and blackout curtains.  In \emph{D09:02--04}, the
user discusses a hot home-cooked meal, garlic and chili aromas, and beef
stir-fried for thirty seconds; \emph{D34:01} mentions coming out of the
kitchen after cooking.  None of these memories states that the user cooked
for a gathering of friends.

\medskip
\fbox{\parbox[c][2.15cm][c]{0.96\linewidth}{\centering
\textbf{Implicit audio evidence behind the lure}\\[0.25ex]
\emph{D00-001--006.wav}: chopping and stir-frying; speech about a diffuser\\
\emph{D09-001--004.wav}: rapid chopping, food entering hot oil, and wok
impacts; speech about finishing work and eating a hot meal\\
\emph{D19-001--006.wav}: chopping and stir-frying; speech about a bookshelf
\quad $\bullet$ \quad
\emph{D25-001--006.wav}: chopping and stir-frying; speech about curtains\\
\emph{D34-001--003.wav}: stir-frying; speech about citrus fragrance and the
kitchen}}

\medskip
\textbf{Question.} When Lin Yue personally cooks for a gathering of friends,
which cooking style does she tend toward?

\medskip
\begin{tabular}{@{}p{0.48\linewidth}p{0.48\linewidth}@{}}
A. Cutting ingredients fresh and stir-frying them for wok hei. &
B. Low-temperature slow cooking that emphasizes natural flavors.\\
C. Primarily steaming and stewing with careful heat control. &
D. The history does not specify.
\end{tabular}

\medskip
\textbf{Gold answer: D.}\quad Option A is a tempting lure because the memory
supports an everyday preference for fresh stir-frying.  The question,
however, inserts an unobserved event in which the user cooks for friends, so
the cooking style in that specific situation is unsupported.  This is a
\emph{false presupposition}.
\end{mybox}
\caption{Complete implicit-lure Answer Refusal example.}
\label{fig:example-refusal-implicit}
\end{figure*}

\section{Experimental Setup Details}
\label{app:experimental_details}

\subsection{Models and Inference}

\paragraph{Backbones and roles.}
Table~\ref{tab:model_roles} summarizes the model names reported in the main
paper and their roles in the experiments.

\begin{table*}[!htbp]
\centering
\small
\setlength{\tabcolsep}{5pt}
\begin{tabular}{p{0.42\textwidth}p{0.48\textwidth}}
\toprule
\textbf{Role} & \textbf{Reported model} \\
\midrule
RQ1/RQ2 answer backbone &
Qwen3.6-35B-A3B \\
RQ1 answer backbone &
GPT-5.4-mini \\
RQ2 image captioner &
GPT-5.4 \\
RQ2 speech/background models &
Qwen3-ASR-1.7B; Gemini 3.1 Pro \\
RQ3 native answer backbones &
Qwen3-Omni-30B-A3B-Instruct; MiniCPM-o-4.5 \\
RQ3 embedding &
Gemini Embedding 2 \\
\bottomrule
\end{tabular}
\caption{Model roles and reported model names.}
\label{tab:model_roles}
\end{table*}

\paragraph{Common RQ1/RQ2 inference.}
The benchmark stores one observation per dialogue round.  For textual memory,
user/assistant text is concatenated with image and voice captions; image-option
questions append the four option captions to the stem.  The category-specific
prompt asks the model to select one of A--D.  Runs used temperature 0,
\textbf{max\_tokens}=1,024, and global seed 42; the seed is also sent to the
API only for Qwen-compatible model names, and Qwen thinking mode is disabled.
The runner uses up to 16 concurrent answer requests in the documented main
commands.  A response is parsed by taking an initial A--D letter, or otherwise
the first standalone A--D token.  Empty responses remain incomplete and are
retried on resume.  API failures are retried five times with exponential
backoff from one second; on an input-length error, the output budget is halved
down to 128 tokens.

\paragraph{Question-only filtering and repeated runs.}
The question-only prompt contains no history and explicitly forbids claiming
access to past text or media.  Its three stored runs per backbone use the same
answer parser, seed 42, temperature 0, five API attempts, and up to 16 workers.
These are the six predictions used only for the all-six-correct pruning rule.
In contrast, the transferred RQ1--RQ3 result trees contain one result artifact
per configuration rather than three independently identified runs.  Thus, the
repository does not support the stronger statement requested in the drafting
comment that every reported configuration is a three-run mean; resolving this
requires either provenance for two missing runs or new repeated experiments.

\paragraph{Human baseline.}
The browser interface implements full-multimodal, full-text, oracle-multimodal,
and oracle-text views, records A--D responses, and keeps the history and QA
panes independently scrollable.  It hides scene descriptions, generated
captions, background-audio metadata, and image-option captions from
participants; oracle clues are collapsed by default.  Three human annotators
completed the human-baseline evaluation across all 20 profiles and all four
variants, requiring more than 320 annotator-hours in total.

\begin{figure*}[!htbp]
\centering
\begin{minipage}[t]{0.68\textwidth}\vspace{0pt}
  \centering
  \includegraphics[width=\linewidth]{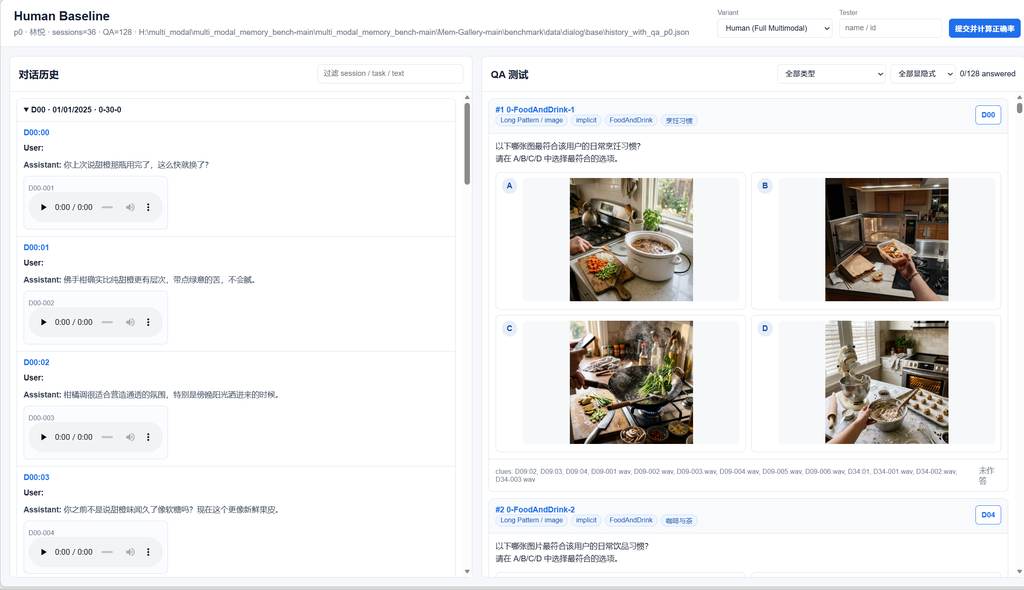}\\[-0.2ex]
  \small\textbf{(a)} Human-baseline annotation interface.
\end{minipage}\hfill
\begin{minipage}[t]{0.28\textwidth}\vspace{0pt}
  \centering
  \includegraphics[width=\linewidth]{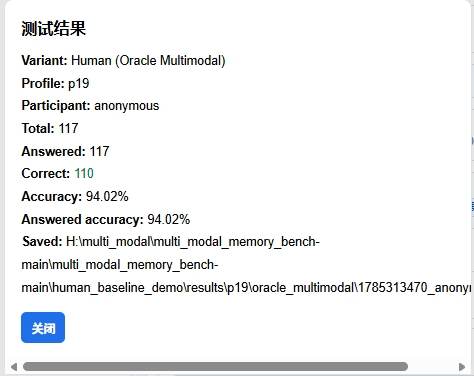}\\[-0.2ex]
  \small\textbf{(b)} Example completion and accuracy summary.
\end{minipage}
\caption{Human-baseline evaluation interface and an example submitted result.
The interface presents the multimodal dialogue history and four-option QA side
by side, while the completion dialog records coverage and accuracy for a
profile--variant assignment.}
\label{fig:human_baseline_interface}
\end{figure*}

\subsection{Memory-System Configurations}

\paragraph{Shared representation and retrieval defaults.}
All seven systems receive the same ordered dialogue-round observations.
Images and audio are converted to the selected captions before storage in
RQ1/RQ2.  The shared text retriever is
\textbf{paraphrase-multilingual-MiniLM-L12-v2} (384 dimensions), loaded from
the repository-local \textbf{benchmark/run/model} directory.  Embeddings are
normalized and ranked by cosine similarity.  Unless noted otherwise,
retrieval uses top--10 and recalled text is truncated to 4,096 words.  The
snapshot exposes no hyperparameter sweep logs; the values below are the final
checked-in configuration, not selections reconstructed after the fact.

\begin{table*}[!htbp]
\centering
\scriptsize
\setlength{\tabcolsep}{3.5pt}
\begin{tabular}{p{0.12\textwidth}p{0.18\textwidth}p{0.17\textwidth}p{0.43\textwidth}}
\toprule
\textbf{Paper name} & \textbf{Implementation} & \textbf{Recall} &
\textbf{Storage/update policy and system-specific values} \\
\midrule
Full Memory & \textbf{FUMemory} & Ordered history, 4,096-word budget &
Linear, round-level storage; no consolidation or learned update. \\
NaiveRAG & \textbf{LTMemory} & MiniLM cosine, top--10 &
Linear, round-level observations; embeddings added at storage time. \\
Generative Agents & \textbf{GAMemory} &
Semantic, recency, and importance fusion; top--10 &
Exponential recency decay 0.995; importance scored 1--10 by the answer
backbone; reflection threshold 0.3, three salient questions and three
insights, reflection top--10. \\
Reflexion & \textbf{RFMemory} & Full-history recall &
Full-memory store plus a trial reflector that updates a concise high-level
insight from the previous insight and the new trial. \\
MemGPT & \textbf{MGMemory} & Semantic recall for active/archive stores,
top--10 &
Recursive LLM summary on a 4,096-word flush boundary; warning threshold 0.7;
the checked-in benchmark sets \textbf{skip\_trigger=True}. \\
A-Mem & \textbf{AMemMemory} & MiniLM cosine, top--10 &
Each round/caption is an A-Mem note; robust agentic-memory evolution uses
threshold 100 and interval 15; memory-side LLM is patched to the current
answer backbone; cached notes are keyed by data/config hashes. \\
MemoryOS & \textbf{MemoryOSMemory} &
top--10 sessions and top--10 knowledge items &
Short/mid/long capacities 10/2,000/100; retrieval queue 7; similarity
threshold 0.6; fast mode with batches of 15 and LLM summaries; heat threshold
999,999 suppresses frequent profile updates; final answering remains in the
benchmark runner. \\
\bottomrule
\end{tabular}
\caption{Checked-in RQ1 memory-system configurations.  \textbf{LTMemory} is
the implementation of the NaiveRAG row in the paper.}
\label{tab:memory_configs}
\end{table*}

\paragraph{Adaptations.}
The runner attaches \textbf{image\_id}/\textbf{voice\_id} to each stored round
and appends captions to systems that accept text only.  A-Mem and MemoryOS
export/import caches so the expensive history-ingestion phase can be reused.
MemoryOS also prepends \textbf{dialogue\_id} to stored content so retrieved
items can be mapped back to benchmark clue IDs.  These adaptations affect
storage and provenance only; the common answer backbone always makes the final
A--D decision.

\subsection{Textualization Configurations}
\label{app:textualization_config}

\paragraph{Image captions.}
The base captioning utility calls GPT-5.4 with low reasoning effort (falling
back to temperature 0), up to six attempts, and the instruction:
\begin{examplebox}
\textbf{Image Captioning Prompt.}

\medskip
\textbf{Generate a concise caption for this image. Describe only visible
details, output plain text only, and answer in Chinese.}
\end{examplebox}
The repository contains the realized Brief, Medium, and Detailed histories and
option captions.  It does not contain the three distinct prompt templates or
generation logs that produced those files, so the exact increasing
peripheral-detail instructions, output budgets, and any tried alternatives
cannot be reproduced from this snapshot.  We consequently do not fabricate
prompt wording here; the missing templates are listed in the unresolved-items
document.

\paragraph{Image-caption examples.}
The following cases present faithful English translations of the realized
Chinese captions.  Brief captions identify the dominant subject, Medium
captions add object relations and scene context, and Detailed captions
enumerate foreground, background, and peripheral content.

\begin{table*}[!htbp]
\centering
\setlength{\fboxsep}{8pt}
\colorbox{gray!8}{%
\begin{minipage}{0.94\textwidth}\small
\textbf{Image Case I1---Design-material workspace (profile 1, event 13;
stored task 0-13-0)}

\begin{minipage}[t]{0.28\linewidth}\vspace{0pt}\centering
\includegraphics[width=\linewidth,height=5.0cm,keepaspectratio]{supplementary_images/event/images/pid_0000_task_0-13-0.jpg}\\[0.4ex]
\scriptsize Original event image
\end{minipage}\hfill
\begin{minipage}[t]{0.68\linewidth}\vspace{0pt}\footnotesize
\textbf{Brief.} Several color-swatch cards, fabric samples, and a pen are
arranged on a white tabletop.

\smallskip
\textbf{Medium.} The image shows an interior designer's workstation covered
with material boards containing brown, beige, and orange color swatches and
fabric samples.  A pen and a black desk lamp sit nearby, while a bookcase and
window in the background suggest a professional studio.  The caption captures
the primary design activity but does not identify the peripheral magazine.

\smallskip
\textbf{Detailed.} The image is a close-up of a professional design table.
Several overlapping sample sheets display low-saturation earth tones,
including beige, tan, ochre, brown, and orange, and some sheets incorporate
physical textile and paper samples.  A silver-gray marker lies to the right.
At the upper left, a \textbf{black adjustable-arm desk lamp} provides local
illumination.  At the lower-left edge, the cover of a
\textbf{slightly dog-eared \emph{Wallpaper*} design magazine} is visible.
Bookshelves and an office chair remain softly blurred in the background.
Unlike the lower-granularity captions, the Detailed caption retains the
complete pair of peripheral entity anchors, thereby preserving evidence for
the implicit preference for consulting design magazines for inspiration.
\end{minipage}
\end{minipage}}
\end{table*}

\begin{table*}[!htbp]
\centering
\setlength{\fboxsep}{8pt}
\colorbox{gray!8}{%
\begin{minipage}{0.94\textwidth}\small
\textbf{Image Case I2---Ferret in a gym bag inside a car (profile 4, task 4-33-0)}

\begin{minipage}[t]{0.28\linewidth}\vspace{0pt}\centering
\includegraphics[width=\linewidth,height=5.0cm,keepaspectratio]{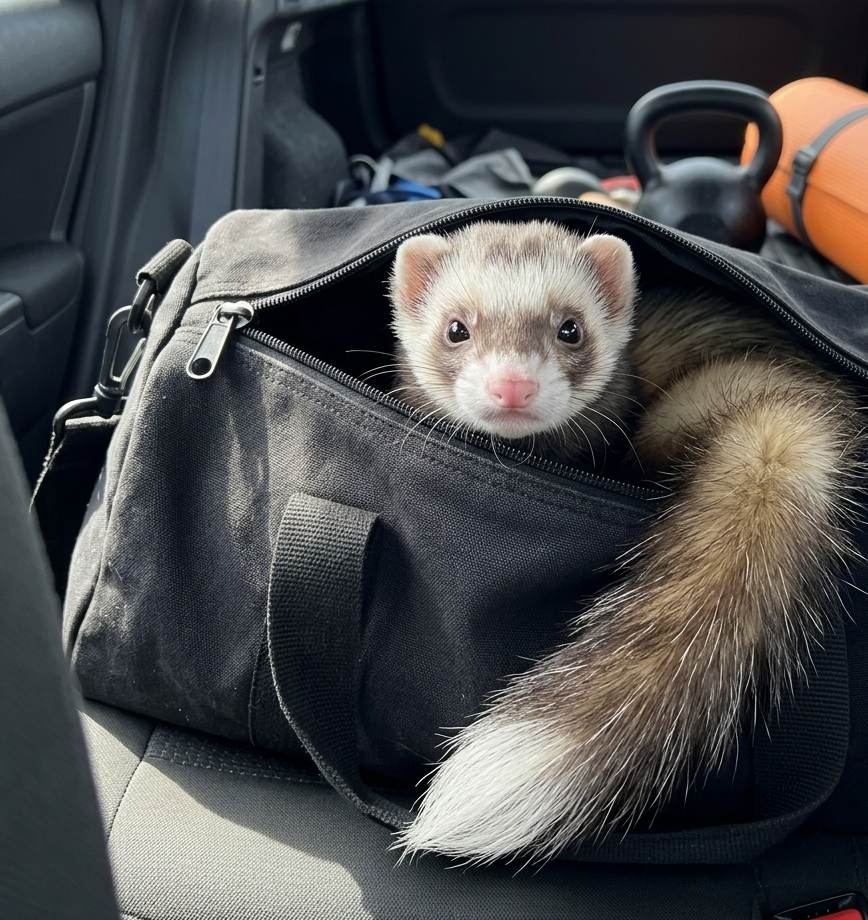}\\[0.4ex]
\scriptsize Original event image
\end{minipage}\hfill
\begin{minipage}[t]{0.68\linewidth}\vspace{0pt}\footnotesize
\textbf{Brief.} A ferret peers out of a black backpack, with its tail hanging
outside the bag.

\smallskip
\textbf{Medium.} A curious ferret with a pink nose and dark eyes peers through
the zipper opening of a black canvas bag in the rear of a car.  The caption
mentions generic fitness objects in the background but misidentifies them as
a dumbbell and an orange water bottle, so it does not preserve the intended
entity anchors.

\smallskip
\textbf{Detailed.} The ferret occupies the foreground, with most of its body
inside a worn black canvas gym bag and its bushy, white-tipped tail extending
outside.  The surrounding door panel and seat establish that the scene is
inside a car.  Behind the bag, the caption resolves two peripheral fitness
objects: an \textbf{iron matte-black training kettlebell} and a
\textbf{rolled orange yoga mat secured by fixed black straps}.  These details
constitute the entity anchors for the implicit preference that the user
regularly drives between gyms and training venues while keeping exercise
equipment in the car; neither Brief nor Medium recovers them correctly.
\end{minipage}
\end{minipage}}
\end{table*}

\paragraph{Audio captions.}
The \emph{ASR} variant replaces each mixed clip by the Qwen3-ASR-1.7B speech
transcript and discards background sound.  The \emph{Hint} variant sends the
mixed waveform to Gemini 3.1 Pro at temperature 0 and requests exactly two
Chinese lines: a near-verbatim ``Human speech:'' transcript and a
``Background sound:'' description restricted to audible content, with
``none'' values when absent.  Up to six exponentially backed-off attempts are
made.  The \emph{Split} variant does not run source separation on the mixed
waveform: it combines the Qwen speech-only transcript with a separately
synthesized background-audio track resolved through the manifest.  Gemini
3.1 Pro captions that background track at temperature 0 using:
\begin{examplebox}
\textbf{Background-Audio Captioning Prompt.}

\medskip
\textbf{Describe this background audio in one Chinese sentence. Describe only
what can be heard; do not guess images, people, or dialogue. Output plain text
without a heading.}
\end{examplebox}
The two fields are finally concatenated as ``Human speech'' and ``Background
sound.''  The Qwen3-ASR decoding command and parameters are not present in the
repository.

\paragraph{Audio-caption examples.}
Each box shows the three textualizations of the same mixed waveform.  The
known background track is included only to make the information retained by
each configuration explicit.

\begin{table*}[!htbp]
\centering
\setlength{\fboxsep}{8pt}
\colorbox{gray!8}{%
\begin{minipage}{0.94\textwidth}\small
\textbf{Audio Case A1---Background event omitted from the mixed-waveform hint
(profile 0, D00-001)}

\footnotesize
\textbf{Known background track.} Chopping vegetables.

\smallskip
\textbf{ASR.} Human speech: I just smelled the new diffuser again; the
bitterness of the bergamot is quite striking.

\smallskip
\textbf{Hint.} Human speech: I just smelled the new diffuser again; the
bitterness of the bergamot is quite striking.\\
Background sound: No obvious background sound.

\smallskip
\textbf{ASR+BG Split.} Human speech: I just smelled the new diffuser again;
the bitterness of the bergamot is quite striking.\\
Background sound: Rapid, rhythmic chopping; a knife repeatedly lands on a
cutting board with crisp impacts.
\end{minipage}}
\end{table*}

\begin{table*}[!htbp]
\centering
\setlength{\fboxsep}{8pt}
\colorbox{gray!8}{%
\begin{minipage}{0.94\textwidth}\small
\textbf{Audio Case A2---Background event confused in the mixed waveform
(profile 0, D15-002)}

\footnotesize
\textbf{Known background track.} A coffee machine operating.

\smallskip
\textbf{ASR.} Human speech: I decisively replaced the coral pink with a
grayish terracotta; the whole design immediately became grounded and no
longer felt unsettled.

\smallskip
\textbf{Hint.} Human speech: I decisively replaced the coral pink with a
grayish terracotta; the whole design immediately became grounded and no
longer felt unsettled.\\
Background sound: Water being poured.

\smallskip
\textbf{ASR+BG Split.} Human speech: I decisively replaced the coral pink
with a grayish terracotta; the whole design immediately became grounded and
no longer felt unsettled.\\
Background sound: A coffee machine is operating, producing a continuous
mechanical hum and slight vibration.
\end{minipage}}
\end{table*}

\paragraph{Token accounting.}
The analysis uses \textbf{tiktoken} \textbf{cl100k\_base} as a common
approximation.  It counts the exact strings written to memory (dialogue,
image-caption, and voice-caption components), the top--10 NaiveRAG recalled
strings per QA, and the final question plus option captions.  Across all 20
users, the stored-history totals are 606,440, 670,142, and 1,196,534 tokens
for Brief, Medium, and Detailed, respectively---means of 30,322, 33,507.1,
and 59,826.7 tokens per user.

\raggedbottom

\subsection{Native Multimodal RAG Implementation}
\label{app:multimodal_config}

\paragraph{Controlled index/use matrix.}
RQ3 crosses how memory is indexed with what the answer model receives:
\begin{center}
\small
\begin{tabular}{lll}
\toprule
 & \textbf{Text-Use} & \textbf{MM-Use} \\
\midrule
\textbf{Text-Index} & TT & TM \\
\textbf{MM-Index} & MT & MM \\
\bottomrule
\end{tabular}
\end{center}
Text-Index embeds one textualized dialogue round, including available image
and audio captions.  MM-Index emits one text item for the original
user/assistant content and a separate item for every original image and audio
attachment, placing all items in a unified embedding matrix.  Text-Use renders
the retrieved turns and image-option captions as text.  MM-Use instead restores
the original images and audio associated with the retrieved turn, while
retaining user/assistant text; if a media path is missing, it falls back to the
stored caption.

\paragraph{Retrieval.}
The final checked-in path uses
\textbf{google/gemini-embedding-2} with 3,072-dimensional L2-normalized
vectors.  Similarity is the dot product of normalized vectors (cosine).
Questions with image options form option-conditioned question--image query
items.  For each query item, the retriever ranks candidates and combines
turn-level ranks with reciprocal-rank fusion, $1/(60+r)$; the candidate pool
is $\max(20,4k)$ and the final $k$ is 5.  Text-Index and unified MM-Index use
the same composed-query mechanism, isolating the index modality rather than
changing the query.  Retrieved item metadata records item ID, turn/session ID,
modality, path, score, rank, and query source, allowing evaluation against
turn/image/audio clue IDs.

\flushbottom

\paragraph{Answer formatting and limits.}
Both native backbones use the category prompt and the same A--D output
contract.  The client uses temperature 0, 64 output tokens without a rationale
or 512 with the stored rationale format, and disables Qwen thinking mode.

\paragraph{Oracle controls.}
Oracle Text selects all sessions containing a gold clue and renders their
textual annotations.  Oracle Multimodal selects the same evidence sessions and
passes original user/assistant text plus the corresponding raw image/audio
attachments; a clue-turn-only mode also exists, whereas the default oracle
scope is the full evidence session.  The oracle runner uses temperature 0,
64/512 output tokens, four retries with exponential backoff from two seconds,
and caption fallback for missing media.  No retrieval takes place in either
oracle.

\subsection{Code and Reproducibility Package}
\label{app:code_release}
The reproducibility materials include the programs used for profile and event
construction, multimodal synthesis and textualization, QA generation and clue
verification, benchmark-format conversion, and dataset statistics.  They also
include the execution and evaluation pipelines for RQ1--RQ3, the
question-only and oracle settings, evidence-retrieval evaluation, token
analysis, statistical testing, and generation of the reported tables and
figures.  Experiment configurations record the evaluated backbones, memory
systems, caption and audio conditions, random seeds, and expected input and
output organization.  Checkpointing and resume support are provided for the
computationally expensive generation, memory-ingestion, and RQ3 stages.

An end-to-end reproduction first constructs and validates profiles, events,
media, and QAs; builds the benchmark histories and textualization variants;
runs the RQ1 and RQ2 memory evaluations and controls; constructs the text and
multimodal indices for the four RQ3 configurations; and finally aggregates
accuracy, EI-Gap, retrieval, token, and significance statistics.  Paths and
API credentials are machine-specific and must be configured for the target
environment, and reproduction of model-assisted stages requires access to the
corresponding model services.

\section{Additional Experimental Results}

\subsection{Statistical Significance Analysis}
\label{app:significance}

\paragraph{Scope and estimands.}
We ran a three-run robustness audit on the five profiles selected once with
sampling seed 42 (\(p0,p3,p7,p8,p16\)).  Run 1 is the corresponding subset of
the full result tree, and runs 2 and 3 use evaluation seeds 43 and 44.  The
audit covers Qwen3.6-35B-A3B for RQ1--RQ2 and both Qwen3-Omni and MiniCPM-o
for RQ3. As in the main paper,
Memory Avg.\ is the macro-average of Entity Recall, Long Pattern, and
Personalized Recommendation; Answer Refusal is not part of this estimand.
RQ1 averages the seven textual memory systems, while RQ2 averages the six
systems with complete repeated coverage.

\paragraph{Resampling and tests.}
We treat the five audited profiles as a fixed benchmark subset rather than
as a random sample from a population of profiles.  Each of 10,000 bootstrap
replicates therefore retains the same profiles and resamples questions with
replacement only within each observed profile \(\times\) task \(\times\)
evidence stratum.  Within each task, profiles with eligible questions receive
equal weight, after which the three task scores are macro-averaged.  Empty
strata in a modality-specific subset are omitted rather than assigned zero.
The resulting 95\% percentile intervals, computed with bootstrap seed
20260731, are conditional on the fixed audit profiles and exclude uncertainty
from changing profile composition.  Explicit and implicit questions are \emph{not} treated as
one-to-one counterfactual pairs: the RQ1 EI-Gap independently resamples the
two question sets within each fixed profile and stratum.  RQ2 and RQ3
comparisons, in contrast, reuse exactly the same sampled question keys in
both conditions and are therefore paired.  We report risk differences in
percentage points as effect sizes, two-sided bootstrap-tail \(p\)-values,
and Holm-adjusted \(p\)-values over the 13 prespecified contrasts in
Table~\ref{tab:significance_contrasts}; the family-wise threshold is
\(\alpha=0.05\).  For each run, we apply the same fixed-profile aggregation:
questions are averaged within profile and task, eligible profiles are equally
weighted within task, and tasks are macro-averaged.  Run SD is the sample
standard deviation of these three run-level estimates; because all runs use
the same profiles and common question keys, it measures decoding/evaluation
variation rather than profile-composition or sampling uncertainty.

\begin{table*}[!htbp]
\centering
\small
\setlength{\tabcolsep}{3.5pt}
\begin{tabular}{lllcccc}
\toprule
\textbf{Study} & \textbf{Backbone} & \textbf{Setting} &
\textbf{Explicit [95\% CI]} & \textbf{Implicit [95\% CI]} &
\textbf{EI-Gap [95\% CI]} & \textbf{Gap run SD} \\
\midrule
RQ1 & Qwen3.6 & Base
& 66.5 [61.6, 71.2] & 31.7 [27.3, 36.1] & 34.8 [28.2, 41.2] & 4.13 \\
\midrule
RQ2 image & Qwen3.6 & Brief
& 72.4 [65.6, 79.0] & 24.2 [17.8, 30.9] & 48.2 [38.6, 57.4] & 0.17 \\
& & Detailed
& 80.3 [73.9, 86.8] & 44.4 [37.3, 51.9] & 35.9 [26.1, 45.6] & 1.86 \\
RQ2 audio & Qwen3.6 & ASR
& 75.4 [67.9, 82.5] & 41.6 [27.5, 53.6] & 33.8 [19.3, 49.5] & 2.04 \\
& & Split
& 74.4 [66.8, 81.5] & 52.6 [41.5, 62.8] & 21.8 [9.1, 35.1] & 0.89 \\
\midrule
RQ3 & Qwen3-Omni & TT
& 72.7 [67.1, 78.2] & 42.5 [36.5, 49.0] & 30.2 [21.6, 38.6] & 0.75 \\
& & TM
& 79.5 [74.5, 84.3] & 46.7 [40.3, 52.8] & 32.8 [24.9, 41.0] & 0.44 \\
& & MT
& 65.1 [59.1, 71.2] & 35.1 [29.6, 40.7] & 30.0 [21.6, 38.3] & 0.60 \\
& & MM
& 75.7 [70.1, 81.2] & 44.0 [38.1, 50.0] & 31.7 [23.2, 39.7] & 0.55 \\
RQ3 & MiniCPM-o & TT
& 73.1 [67.0, 78.9] & 41.8 [35.4, 48.2] & 31.3 [22.5, 39.9] & 0.18 \\
& & TM
& 56.6 [50.3, 62.9] & 41.4 [35.1, 47.9] & 15.2 [6.2, 24.1] & 0.72 \\
& & MT
& 70.4 [64.2, 76.3] & 41.2 [34.9, 47.4] & 29.2 [20.4, 37.8] & 0.51 \\
& & MM
& 54.8 [48.4, 61.1] & 41.6 [35.3, 48.0] & 13.2 [4.0, 21.9] & 0.91 \\
\bottomrule
\end{tabular}
\caption{Three-run Memory Avg.\ accuracy and EI-Gap on the fixed five-profile
audit subset.  Values are percentages or percentage points.  In RQ3, the
first/second letter denotes Text (T) or Multimodal (M) index/use.}
\label{tab:significance_intervals}
\end{table*}

\begin{table*}[!htbp]
\centering
\small
\setlength{\tabcolsep}{4pt}
\begin{tabular}{lllrrrr}
\toprule
\textbf{Study} & \textbf{Backbone} & \textbf{Contrast and estimand} &
\textbf{Effect [95\% CI]} & \textbf{Run SD} &
\(\boldsymbol{p}\) & \(\boldsymbol{p_{\mathrm{Holm}}}\) \\
\midrule
RQ1 & Qwen3.6 & Base: Explicit \( - \) Implicit
& 34.8 [28.1, 41.3] & 4.13 & 0.0002 & 0.0026 \\
\midrule
RQ2 image & Qwen3.6 & Detailed \( - \) Brief, implicit accuracy
& 20.2 [14.7, 25.8] & 0.93 & 0.0002 & 0.0026 \\
& & Detailed \( - \) Brief, EI-Gap change
& $-$12.3 [$-$18.6, $-$6.0] & 2.02 & 0.0006 & 0.0060 \\
RQ2 audio & Qwen3.6 & Split \( - \) ASR, implicit accuracy
& 11.0 [3.7, 18.3] & 1.98 & 0.0036 & 0.0306 \\
& & Split \( - \) ASR, EI-Gap change
& $-$12.0 [$-$19.6, $-$4.0] & 1.53 & 0.0034 & 0.0306 \\
\midrule
RQ3 & Qwen3-Omni & MM-Use \( - \) Text-Use, Text-Index
& 4.1 [$-$1.0, 9.5] & 0.99 & 0.1204 & 0.7223 \\
& & MM-Use \( - \) Text-Use, MM-Index
& 8.9 [4.1, 13.9] & 1.10 & 0.0002 & 0.0026 \\
& & MM-Index \( - \) Text-Index, Text-Use
& $-$7.4 [$-$13.6, $-$1.3] & 0.28 & 0.0156 & 0.1092 \\
& & MM-Index \( - \) Text-Index, MM-Use
& $-$2.7 [$-$9.0, 3.8] & 0.93 & 0.4160 & 1.0000 \\
RQ3 & MiniCPM-o & MM-Use \( - \) Text-Use, Text-Index
& $-$0.4 [$-$7.1, 6.1] & 0.44 & 0.8969 & 1.0000 \\
& & MM-Use \( - \) Text-Use, MM-Index
& 0.4 [$-$5.6, 6.5] & 1.03 & 0.8911 & 1.0000 \\
& & MM-Index \( - \) Text-Index, Text-Use
& $-$0.6 [$-$6.3, 5.1] & 0.53 & 0.8177 & 1.0000 \\
& & MM-Index \( - \) Text-Index, MM-Use
& 0.3 [$-$6.7, 7.0] & 0.99 & 0.9573 & 1.0000 \\
\bottomrule
\end{tabular}
\caption{Prespecified significance contrasts on the fixed five-profile audit.
RQ1 compares independently resampled explicit and implicit sets; all RQ2 and
RQ3 rows are paired by question.  Profiles are not resampled.  Effects and run SDs are percentage points.
Numerical unadjusted and Holm-adjusted \(p\)-values are shown without significance
symbols.}
\label{tab:significance_contrasts}
\end{table*}

\paragraph{Findings.}
The Holm-corrected results reinforce the main-paper conclusions:
\begin{itemize}
  \item \textbf{Implicit evidence produces a significant memory gap (RQ1).}
  Explicit Memory Avg.\ exceeds implicit Memory Avg.\ by 34.8 points
  (\(p_{\mathrm{Holm}}=0.0026\)), confirming the persistent degradation under
  implicit evidence.
  \item \textbf{Detailed image captions significantly improve implicit memory
  (RQ2).}  Relative to Brief captions, Detailed captions increase implicit
  accuracy by 20.2 points (\(p_{\mathrm{Holm}}=0.0026\)) and narrow EI-Gap by
  12.3 points (\(p_{\mathrm{Holm}}=0.0060\)), supporting the main finding that
  cue-aware image textualization alleviates the textualization bottleneck.
  \item \textbf{Separating speech and background audio yields significant
  gains (RQ2).}  Split increases implicit accuracy by 11.0 points and narrows
  EI-Gap by 12.0 points; both effects remain significant after Holm correction
  (\(p_{\mathrm{Holm}}=0.0306\)).
  \item \textbf{Native multimodal evidence use significantly benefits
  Qwen3-Omni under MM-Index (RQ3).}  MM-Use increases implicit accuracy by
  8.9 points relative to Text-Use (\(p_{\mathrm{Holm}}=0.0026\)), consistent
  with the main-paper conclusion that the benefit of native multimodal access
  is backbone-dependent.
\end{itemize}

\subsection{Additional RQ2 Results}
\label{app:rq2_full}

\paragraph{Image-caption quality.}
Table~\ref{tab:rq2_image_qwen_full} gives the complete per-system results for
Qwen3.6-35B-A3B.  The caption-length intervention affects both the description
stored with an image and the descriptions supplied for image-valued answer
options.  Across memory systems, longer captions help most on Entity Recall:
the average increases from 56.4 to 71.0 for explicit questions and from 10.3
to 47.2 for implicit questions.  Long Pattern also improves monotonically on
average, whereas Personalized Recommendation changes more modestly.  Answer
Refusal behaves differently: detailed captions reduce the average explicit
score from 43.7 to 41.1 and the implicit score from 82.0 to 67.6.  Thus,
additional visual detail is useful for identifying weakly stated evidence,
but is not uniformly beneficial when the correct behavior is to reject an
unsupported premise.

\begin{table*}[!htbp]
\centering
\tiny
\setlength{\tabcolsep}{1.7pt}
\renewcommand{\arraystretch}{1.15}
\resizebox{\textwidth}{!}{%
\begin{tabular}{@{}ll*{24}{c}@{}}
\toprule
& & \multicolumn{6}{c}{Long Pattern} &
\multicolumn{6}{c}{Entity Recall} &
\multicolumn{6}{c}{Personalized Recommendation} &
\multicolumn{6}{c}{Answer Refusal} \\
\cmidrule(lr){3-8}\cmidrule(lr){9-14}\cmidrule(lr){15-20}\cmidrule(l){21-26}
Backbone & Memory &
\multicolumn{3}{c}{explicit} & \multicolumn{3}{c}{implicit} &
\multicolumn{3}{c}{explicit} & \multicolumn{3}{c}{implicit} &
\multicolumn{3}{c}{explicit} & \multicolumn{3}{c}{implicit} &
\multicolumn{3}{c}{explicit} & \multicolumn{3}{c}{implicit} \\
\cmidrule(lr){3-5}\cmidrule(lr){6-8}
\cmidrule(lr){9-11}\cmidrule(lr){12-14}
\cmidrule(lr){15-17}\cmidrule(lr){18-20}
\cmidrule(lr){21-23}\cmidrule(l){24-26}
& & br. & med. & det. & br. & med. & det.
& br. & med. & det. & br. & med. & det.
& br. & med. & det. & br. & med. & det.
& br. & med. & det. & br. & med. & det. \\
\midrule
\multirow{7}{*}{\shortstack[l]{Qwen3.6-\\35B-A3B}}
& Full Memory & 77.6&90.6&87.1&36.6&44.3&47.3&77.3&85.8&88.7&16.2&39.3&65.2&71.6&70.2&76.6&40.6&47.4&48.4&18.2&18.2&16.7&73.3&60.0&44.0\\
& NaiveRAG & 80.0&87.1&81.2&29.8&31.3&39.7&65.3&77.5&81.5&9.2&15.9&41.5&67.0&73.6&75.5&44.4&43.6&54.4&25.8&27.3&27.3&81.3&80.0&77.3\\
& A-Mem & 75.3&84.7&83.5&29.8&35.9&41.2&58.9&72.7&78.6&8.7&16.7&44.5&64.8&73.6&74.3&39.1&45.9&52.9&66.4&66.1&65.0&97.4&97.8&97.4\\
& MemoryOS & 27.1&31.8&40.0&13.0&17.6&22.9&9.6&10.1&14.1&3.2&4.0&8.5&50.0&53.1&56.1&39.8&42.9&46.9&92.7&93.2&93.6&99.3&99.7&99.7\\
& Generative Agents & 51.8&63.5&74.1&19.8&26.7&45.0&27.5&42.8&55.1&3.5&11.2&39.1&62.5&66.8&68.6&45.9&46.6&49.1&60.6&51.5&47.0&78.7&81.3&72.0\\
& Reflexion & 75.3&90.6&89.4&37.4&45.0&48.9&78.3&85.9&89.8&16.4&40.0&65.4&71.6&71.3&75.5&39.8&47.4&49.9&21.2&21.2&19.7&70.7&60.0&38.7\\
& MemGPT & 77.6&91.8&88.2&37.4&44.3&49.6&78.0&85.1&89.1&15.2&39.8&65.7&71.6&69.0&76.6&39.1&47.4&48.4&21.2&21.2&18.2&73.3&61.3&44.0\\
\midrule
& AVG. & 66.4&77.1&77.6&29.1&35.0&42.1&56.4&65.7&71.0&10.3&23.8&47.2&65.6&68.2&71.9&41.2&45.9&50.0&43.7&42.7&41.1&82.0&77.2&67.6\\
\bottomrule
\end{tabular}}
\caption{Full image-caption-quality results for Qwen3.6-35B-A3B
(accuracy, \%).  Brief (br.), medium (med.), and detailed (det.) denote the
three caption granularities.}
\label{tab:rq2_image_qwen_full}
\end{table*}

\paragraph{Audio-caption quality.}
Table~\ref{tab:rq2_audio_qwen_full} reports only questions whose supporting
clue contains audio.  ASR uses the original caption, \emph{hint} adds a
high-level acoustic hint, and \emph{split} separates speech content from
non-speech sound.  For Answer Refusal, weak/normal/strong denote the three
adversarial strengths.  Audio splitting has its clearest average effect on
implicit Long Pattern, increasing accuracy from 33.0 (ASR) to 50.6, while its
effect on explicit Long Pattern is negligible.  It also raises implicit
Personalized Recommendation from 47.1 to 51.4.  In contrast, the implicit
Answer Refusal average falls slightly as adversarial strength increases
(95.9 to 94.2).  Entity Recall has no implicit audio subset in this
intervention and is therefore reported only for explicit questions.

\begin{table*}[!htbp]
\centering
\tiny
\setlength{\tabcolsep}{1.8pt}
\renewcommand{\arraystretch}{1.15}
\resizebox{\textwidth}{!}{%
\begin{tabular}{@{}ll*{21}{c}@{}}
\toprule
& & \multicolumn{6}{c}{Long Pattern} &
\multicolumn{3}{c}{Entity Recall} &
\multicolumn{6}{c}{Personalized Recommendation} &
\multicolumn{6}{c}{Answer Refusal} \\
\cmidrule(lr){3-8}\cmidrule(lr){9-11}\cmidrule(lr){12-17}\cmidrule(l){18-23}
Backbone & Memory &
\multicolumn{3}{c}{explicit} & \multicolumn{3}{c}{implicit} &
\multicolumn{3}{c}{explicit} &
\multicolumn{3}{c}{explicit} & \multicolumn{3}{c}{implicit} &
\multicolumn{3}{c}{explicit} & \multicolumn{3}{c}{implicit} \\
\cmidrule(lr){3-5}\cmidrule(lr){6-8}\cmidrule(lr){9-11}
\cmidrule(lr){12-14}\cmidrule(lr){15-17}
\cmidrule(lr){18-20}\cmidrule(l){21-23}
& & ASR & hint & split & ASR & hint & split
& ASR & hint & split
& ASR & hint & split & ASR & hint & split
& weak & normal & strong & weak & normal & strong \\
\midrule
\multirow{7}{*}{\shortstack[l]{Qwen3.6-\\35B-A3B}}
& Full Memory &81.9&81.9&81.9&41.3&42.8&61.6&88.5&90.8&88.5&77.7&75.7&74.8&47.4&49.6&54.6&47.7&46.6&46.2&93.5&94.2&91.6\\
& NaiveRAG &81.0&80.2&80.2&32.6&33.3&51.4&88.5&85.1&82.8&68.0&71.8&73.8&47.4&46.7&50.2&61.6&60.2&62.0&96.8&97.1&94.5\\
& A-Mem &78.4&82.8&79.3&32.6&33.3&51.4&83.9&83.9&82.8&71.8&69.9&69.9&48.9&49.6&52.4&64.5&62.7&63.8&94.8&97.4&92.9\\
& MemoryOS &34.5&36.2&37.1&15.9&14.5&21.0&16.1&16.1&17.2&49.5&52.4&53.4&40.9&42.3&45.8&90.3&89.6&88.5&99.4&99.4&98.4\\
& Generative Agents &56.9&62.1&58.6&24.6&26.8&46.4&44.8&48.3&46.0&58.3&62.1&61.2&48.9&46.7&50.9&78.5&71.7&73.5&99.4&99.7&98.1\\
& Reflexion &81.0&81.9&81.9&42.0&43.5&60.9&90.8&89.7&89.7&75.7&75.7&74.8&47.4&49.6&53.1&47.3&46.6&45.5&93.9&94.2&91.9\\
& MemGPT &81.0&81.0&82.8&42.0&42.0&61.6&89.7&89.7&88.5&76.7&76.7&75.7&48.9&47.4&53.1&47.0&45.9&45.9&93.5&93.5&91.9\\
\midrule
& AVG. &70.7&72.3&71.7&33.0&33.7&50.6&71.8&71.9&70.8&68.2&69.2&69.1&47.1&47.4&51.4&62.4&60.5&60.8&95.9&96.5&94.2\\
\bottomrule
\end{tabular}}
\caption{Full audio-caption-quality results for Qwen3.6-35B-A3B
(accuracy, \%), restricted to questions whose memory clue contains audio.
Entity Recall has no implicit subset under this intervention.}
\label{tab:rq2_audio_qwen_full}
\end{table*}

\paragraph{Token statistics and accuracy--cost trade-off.}
Table~\ref{tab:rq2_token_stats} reports token counts from the reproducible
audit in \textbf{token\_analysis/token\_stats.json}.  The script samples five
profiles with seed 42 (734 questions in total) and uses
\textbf{cl100k\_base} as a tokenizer proxy.  ``Memory input'' is the complete
multi-session history written to a memory system; ``retrieved input'' is the
top-10 recalled content per question; the final two columns count the
question and its answer options, separately for text- and image-valued
questions.

\begin{table*}[!htbp]
\centering
\small
\setlength{\tabcolsep}{3.6pt}
\resizebox{\textwidth}{!}{%
\begin{tabular}{lrrrrrrrrrr}
\toprule
& \multicolumn{4}{c}{Memory input per profile} &
\multicolumn{4}{c}{Retrieved input per QA} &
\multicolumn{2}{c}{Question + options} \\
\cmidrule(lr){2-5}\cmidrule(lr){6-9}\cmidrule(l){10-11}
Caption & Text & Image cap. & Audio cap. & Total &
Text & Image cap. & Audio cap. & Total & Text QA & Image QA \\
\midrule
Brief
& 22,866 & 1,326 & 8,880 & 33,195
& $842{\pm}131$ & $108{\pm}115$ & $144{\pm}160$ & $1{,}096{\pm}161$
& $110{\pm}28$ & $182{\pm}24$ \\
Medium
& 22,866 & 4,789 & 8,880 & 36,658
& $840{\pm}132$ & $453{\pm}472$ & $145{\pm}156$ & $1{,}439{\pm}491$
& $110{\pm}28$ & $667{\pm}96$ \\
Detailed
& 22,866 & 35,966 & 8,880 & 67,834
& $836{\pm}133$ & $4{,}249{\pm}4{,}476$ & $143{\pm}149$ & $5{,}230{\pm}4{,}475$
& $110{\pm}28$ & $3{,}585{\pm}2{,}866$ \\
\bottomrule
\end{tabular}}
\caption{Token statistics for the image-caption intervention.  Memory-input
entries are means over five sampled profiles; retrieved-input and question
entries are mean $\pm$ standard deviation over questions.  The fixed text and
audio columns are unchanged by image-caption granularity.}
\label{tab:rq2_token_stats}
\end{table*}

Moving from brief to medium captions increases total stored tokens by only
10.4\% and retrieved tokens by 31.3\%.  Detailed captions are substantially
more expensive: relative to brief captions, they increase total memory input
by 104.4\% and retrieved input per question by 377.1\%.  The image-caption
component itself grows by $27.1\times$ (1,326 to 35,966 tokens per profile).
This cost increase accompanies the strongest gains on implicit Entity Recall,
but the task tables show that it does not imply a universal accuracy gain,
especially for Answer Refusal.  The large standard deviations for detailed
retrieval and image-valued questions further show that caption cost is highly
example-dependent rather than a fixed per-query overhead.

\subsection{Additional RQ3 Results}
\label{app:rq3_full}

\paragraph{Evidence-retrieval analysis.}
Table~\ref{tab:rq3_retrieval_full} gives the complete Recall@$k$ and
Precision@$k$ breakdown behind the RQ3 retrieval analysis.  The multimodal
index improves overall recall for $k\leq10$ and improves implicit recall at
every reported cutoff.  Its largest implicit-recall advantage is at $k=25$
(29.7 versus 23.1).  This gain comes with lower precision: for example,
implicit Precision@10 is 8.6 for the multimodal index versus 16.2 for the text
index.  For explicit evidence, the multimodal index has higher Recall@1--5
but lower Recall@10--25.  These patterns indicate that multimodal indexing
surfaces additional weak visual/acoustic evidence, while also introducing
more distractors as the retrieved set grows.

\begin{table*}[!htbp]
\centering
\small
\setlength{\tabcolsep}{4.0pt}
\renewcommand{\arraystretch}{1.12}
\resizebox{\textwidth}{!}{%
\begin{tabular}{@{}llrrrrrrrrrr@{}}
\toprule
& & \multicolumn{5}{c}{Recall@$k$} &
\multicolumn{5}{c}{Precision@$k$} \\
\cmidrule(lr){3-7}\cmidrule(l){8-12}
Index & Split & @1 & @3 & @5 & @10 & @25
& @1 & @3 & @5 & @10 & @25 \\
\midrule
\multirow{3}{*}{Text Index}
& Overall  & 4.5&11.6&17.2&27.4&44.6&43.3&39.7&37.3&32.4&23.9\\
& Explicit & 6.6&17.1&25.2&39.2&61.7&60.8&55.8&52.5&45.3&33.0\\
& Implicit & 1.8&4.7&7.1&12.5&23.1&21.3&19.6&18.2&16.2&12.4\\
\midrule
\multirow{3}{*}{Multimodal Index}
& Overall  & 6.2&14.6&20.1&28.6&42.6&42.4&36.0&31.0&23.1&15.4\\
& Explicit & 8.0&20.0&27.1&37.3&52.9&62.5&54.9&47.1&34.7&22.8\\
& Implicit & 3.9&7.9&11.3&17.7&29.7&17.2&12.2&10.8&8.6&6.1\\
\bottomrule
\end{tabular}}
\caption{Evidence-retrieval results (\%) for text and multimodal indices,
reported overall and by evidence explicitness.}
\label{tab:rq3_retrieval_full}
\end{table*}

\subsection{Effect of Cue Recurrence}
\label{app:recurrence}

\paragraph{Recurrence by construction.}
CUE-Mem represents a preference as a recurring property rather than as a
single isolated mention.  The event-group manifest contains 671 planned
events for 20 users.  Of these, 551 pair one explicit preference with one
implicit preference in a shared scene.  The remaining 120 are explicit-only
Relationship events, which introduce or revisit named people without
inventing an unrelated implicit cue.  Pairings vary across occurrences, so a
repeated implicit cue is not tied to one fixed explicit foreground context.

Table~\ref{tab:cue_recurrence} summarizes the resulting coverage.  Across the
manifest's 202 user-specific explicit targets, every target occurs in at least
two event groups; across 152 implicit targets, every target occurs in at least
three.  Explicit targets appear 3.32 times on average, while implicit targets
appear 3.62 times.  The higher minimum for implicit evidence is deliberate:
an implicit preference should be supported by recurring peripheral evidence,
not inferred from a single incidental object or sound.

\begin{table*}[!htbp]
\centering
\small
\setlength{\tabcolsep}{5.0pt}
\begin{tabular}{lrrrrrrrrr}
\toprule
& & \multicolumn{6}{c}{Occurrence count} & & \\
\cmidrule(lr){3-8}
Evidence & Targets & 2 & 3 & 4 & 5 & 6 & 7 & Mean & Median \\
\midrule
Explicit & 202 & 76 & 36 & 51 & 29 & 8 & 2 & 3.32 & 3 \\
Implicit & 152 & 0  & 82 & 50 & 16 & 3 & 1 & 3.62 & 3 \\
\bottomrule
\end{tabular}

\vspace{1.2ex}
\begin{tabular}{lrrrrrr}
\toprule
Evidence & Repeated targets & Mean span & Median span &
Median minimum gap & Mean adjacent gap & Span $\geq271$ days \\
\midrule
Explicit & 198 & 337.8 & 352 & 133.5 & 204.2 & 182 (91.9\%) \\
Implicit & 150 & 339.1 & 354 & 133.0 & 146.8 & 136 (90.7\%) \\
\bottomrule
\end{tabular}
\caption{Cue recurrence and temporal coverage.  The upper panel counts
user-specific preference targets in the 671-group construction manifest.
The lower panel measures days between realized occurrences in the 648
generated events; it includes targets with at least two realized
occurrences.  ``Minimum gap'' is computed within each target and then
summarized across targets.}
\label{tab:cue_recurrence}
\end{table*}

\paragraph{Temporal coverage.}
Repeated occurrences are distributed over the 2025 user history rather than
placed in a short burst.  Dates are assigned to enlarge the first-to-last
span and the gaps between adjacent occurrences of the same target, with
stronger emphasis on implicit targets; repeated explicit--implicit pairings
are also separated.  Events for the same user occupy distinct dates.  In the
648 realized events, the median first-to-last span is 352 days for explicit
targets and 354 days for implicit targets.  More than 90\% of repeated targets
span at least 271 days, and the median shortest gap is approximately 133 days
for both evidence types.  Implicit targets have a smaller mean adjacent gap
(146.8 versus 204.2 days) because they recur more often within a similarly
long annual span.

These statistics characterize a construction control, not an accuracy
ablation.  Recurrence count and temporal spacing were not independently
randomized, and no matched evaluation was run in which the same question was
answered after one versus several occurrences.  We therefore do not claim
that the observed counts causally improve detection or aggregation.
Instead, recurrence ensures that benchmark answers can be grounded in
longitudinal evidence, while the wide spacing prevents a nominally
long-memory question from being reducible to a cluster of adjacent sessions.

\begin{figure}[!htbp]
\centering
\resizebox{\columnwidth}{!}{%
\begin{minipage}{\textwidth}
\begin{mybox}
\textbf{Case R1---A profession stereotype replaces the missing private habit}
\hfill \textbf{p13 / 13-WorkAndLearning-1 / implicit}

\medskip
\textbf{Gold memory clues.}
In \emph{D02:02}, the user says, ``I froze the breakout frame and photographed
it beside the curves in my notebook.''  The shared image additionally exposes
a faded swim cap and a worn stopwatch inside the coach's bag.  \emph{D14:02}
shows another poolside bag while the dialogue concerns reapplying sunscreen;
\emph{D27:02} concerns a shell keepsake and is a weaker member of the annotated
clue set.  The private, diagnostic detail is therefore the old equipment kept
inside the bag, not merely the user's occupation.

\medskip
  \begin{minipage}[t]{0.31\linewidth}\centering
  \IfFileExists{supplementary_images/event/images/pid_0013_task_13-15-0.jpg}{%
    \includegraphics[width=\linewidth,height=3.1cm,keepaspectratio]{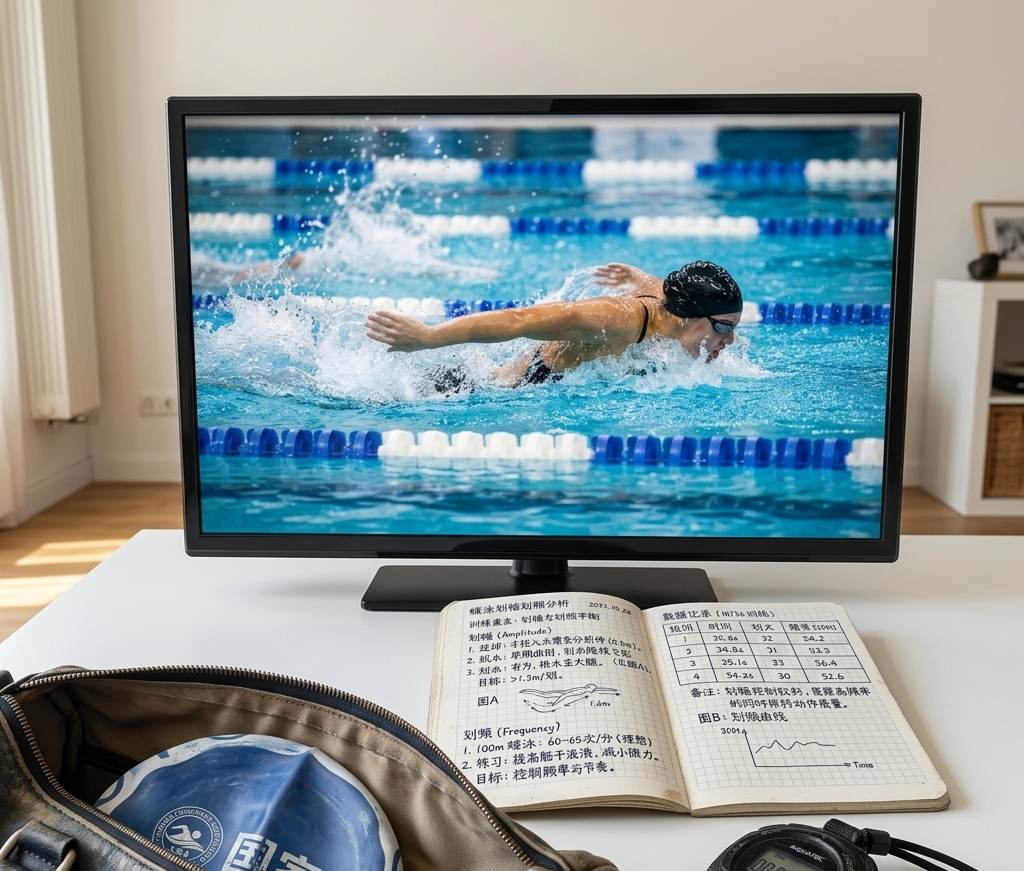}%
  }{%
    \fbox{\parbox[c][2.9cm][c]{0.88\linewidth}{\centering\scriptsize
    Original memory image\\not in transferred snapshot}}%
  }\\[0.45ex]
  \scriptsize\textbf{D02-001.png}\enspace notebook, cap, and stopwatch
  \end{minipage}
  \begin{minipage}[t]{0.31\linewidth}\centering
  \IfFileExists{supplementary_images/event/images/pid_0013_task_13-11-0.jpg}{%
    \includegraphics[width=\linewidth,height=3.1cm,keepaspectratio]{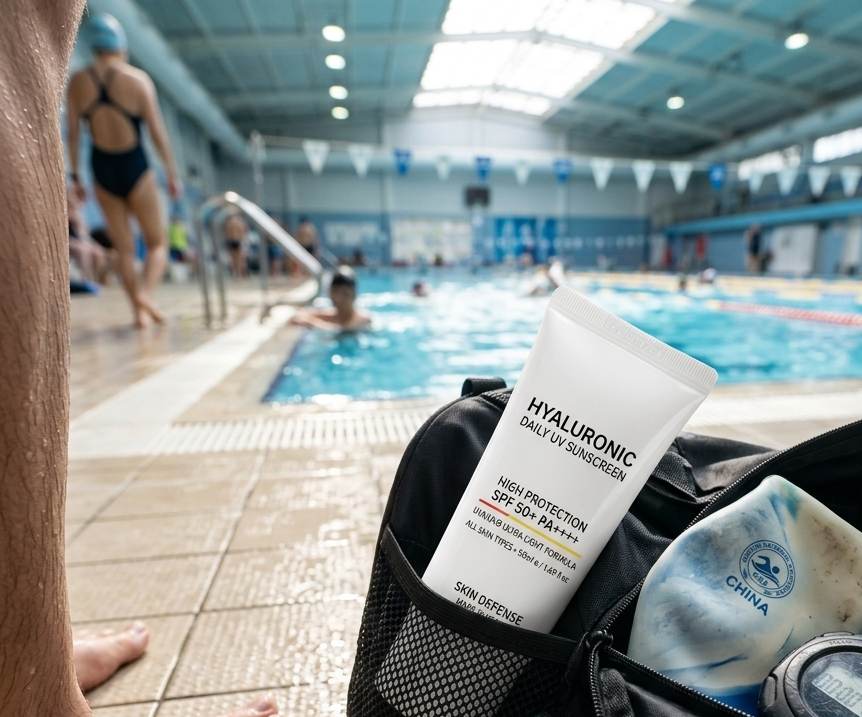}%
  }{%
    \fbox{\parbox[c][2.9cm][c]{0.88\linewidth}{\centering\scriptsize
    Original memory image\\not in transferred snapshot}}%
  }\\[0.45ex]
  \scriptsize\textbf{D14-001.png}\enspace poolside coach's bag
  \end{minipage}
  \begin{minipage}[t]{0.31\linewidth}\centering
  \IfFileExists{supplementary_images/event/images/pid_0013_task_13-19-0.jpg}{%
    \includegraphics[width=\linewidth,height=3.1cm,keepaspectratio]{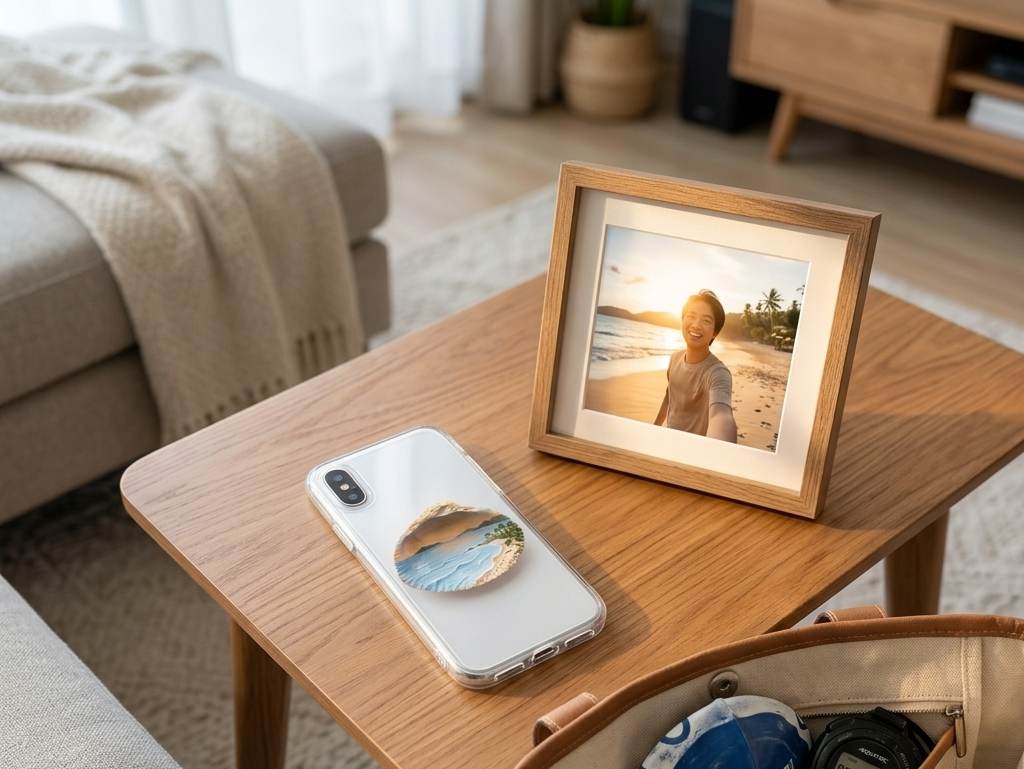}%
  }{%
    \fbox{\parbox[c][2.9cm][c]{0.88\linewidth}{\centering\scriptsize
    Original memory image\\not in transferred snapshot}}%
  }\\[0.45ex]
  \scriptsize\textbf{D27-001.png}\enspace weaker contextual clue
  \end{minipage}

\medskip
\textbf{Question.} Which image best matches the user's private habit in her
professional role?

\medskip
  \begin{minipage}[t]{0.235\linewidth}\centering
  \IfFileExists{supplementary_images/qa/pref_images/13-WorkAndLearning-1_A.jpg}{%
    \includegraphics[width=\linewidth,height=2.65cm,keepaspectratio]{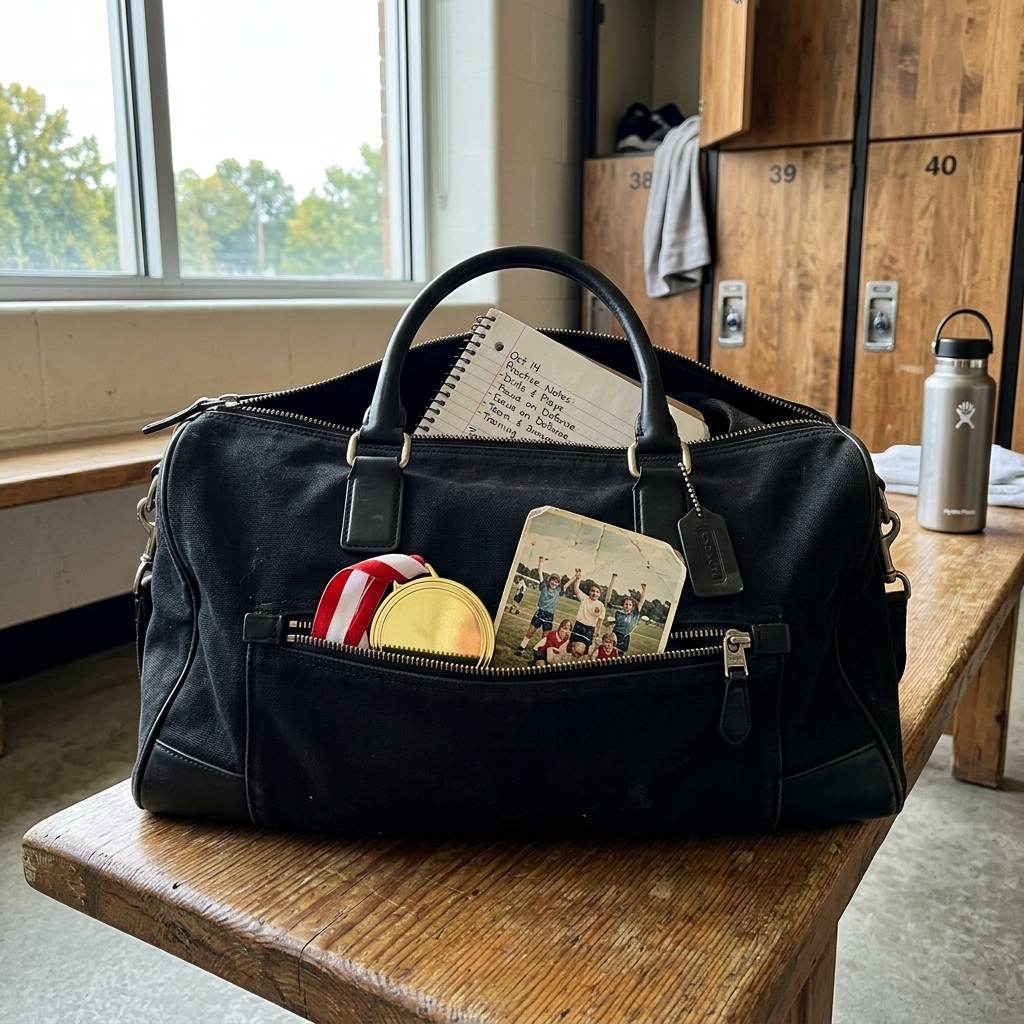}%
  }{%
    \fbox{\parbox[c][2.45cm][c]{0.88\linewidth}{\centering\scriptsize
    Option image unavailable}}%
  }\\[0.45ex]
  \scriptsize\textbf{A}\enspace medal and old photograph
  \end{minipage}
  \begin{minipage}[t]{0.235\linewidth}\centering
  \IfFileExists{supplementary_images/qa/pref_images/13-WorkAndLearning-1_B.jpg}{%
    \includegraphics[width=\linewidth,height=2.65cm,keepaspectratio]{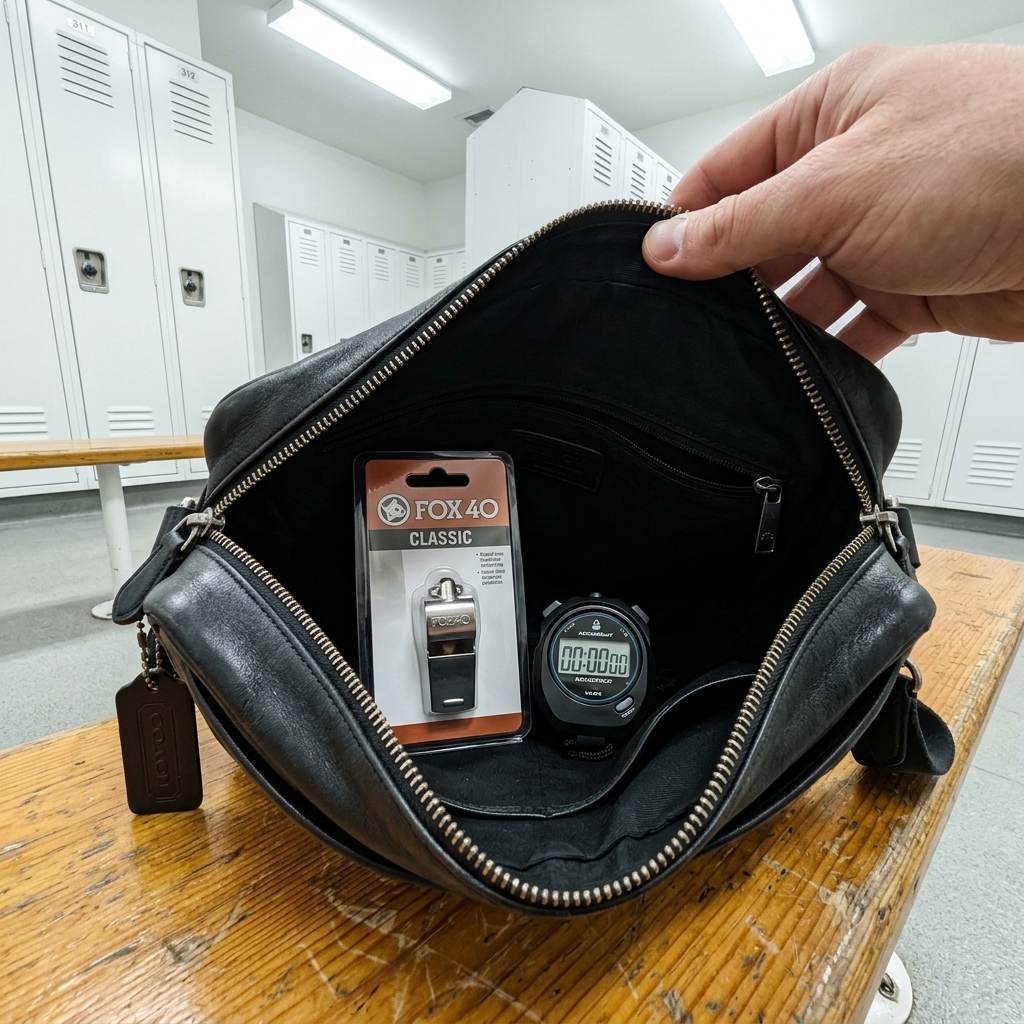}%
  }{%
    \fbox{\parbox[c][2.45cm][c]{0.88\linewidth}{\centering\scriptsize
    Option image unavailable}}%
  }\\[0.45ex]
  \scriptsize\textbf{B}\enspace new whistle and stopwatch
  \end{minipage}
  \begin{minipage}[t]{0.235\linewidth}\centering
  \IfFileExists{supplementary_images/qa/pref_images/13-WorkAndLearning-1_C.jpg}{%
    \includegraphics[width=\linewidth,height=2.65cm,keepaspectratio]{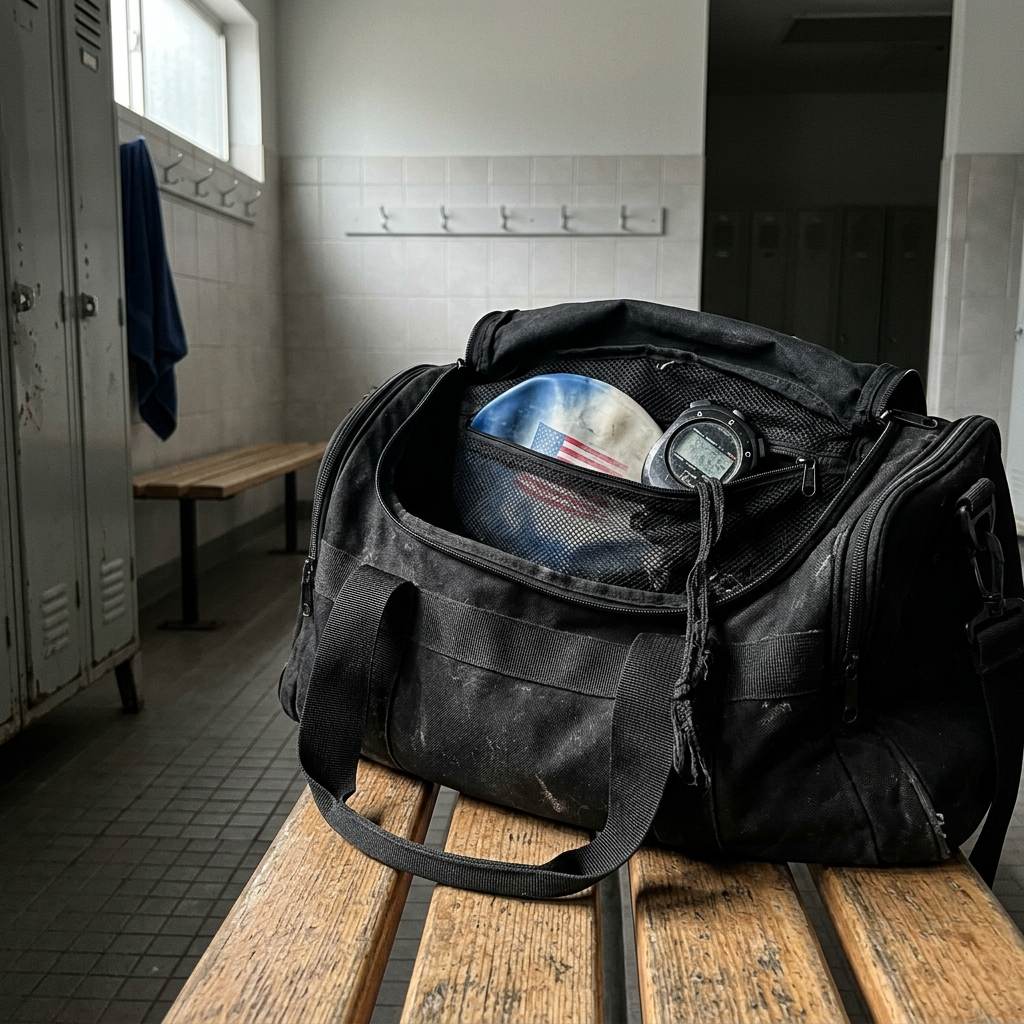}%
  }{%
    \fbox{\parbox[c][2.45cm][c]{0.88\linewidth}{\centering\scriptsize
    Option image unavailable}}%
  }\\[0.45ex]
  \scriptsize\textbf{C}\enspace old cap and worn stopwatch
  \end{minipage}
  \begin{minipage}[t]{0.235\linewidth}\centering
  \IfFileExists{supplementary_images/qa/pref_images/13-WorkAndLearning-1_D.jpg}{%
    \includegraphics[width=\linewidth,height=2.65cm,keepaspectratio]{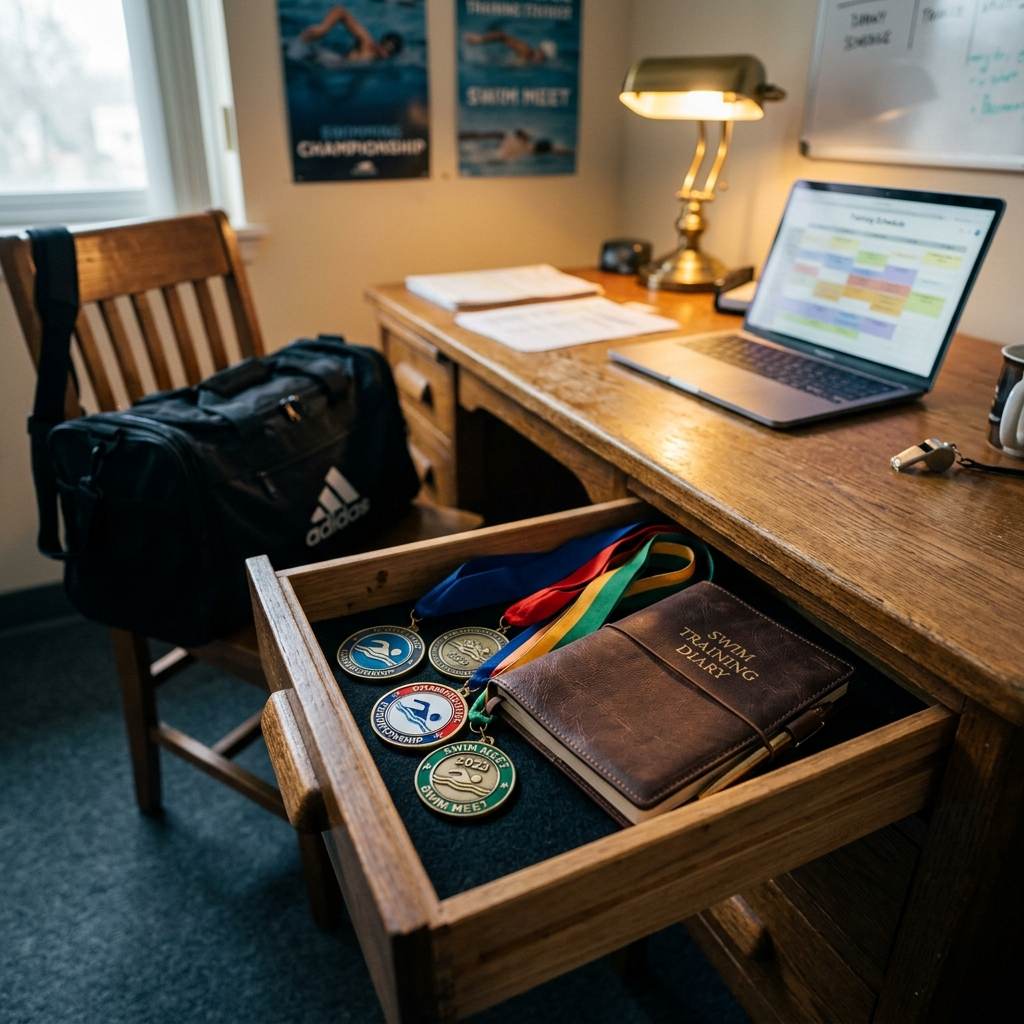}%
  }{%
    \fbox{\parbox[c][2.45cm][c]{0.88\linewidth}{\centering\scriptsize
    Option image unavailable}}%
  }\\[0.45ex]
  \scriptsize\textbf{D}\enspace medals in a desk drawer
  \end{minipage}

\medskip
\begin{tabular}{@{}p{0.485\linewidth}p{0.485\linewidth}@{}}
\textbf{Text-Index} & \textbf{Multimodal-Index} \\
Top-5: D15:02, D15:01, D20:04, D31:07, D04:06 &
Top-5: D02:02, D26:03, D31:06, D01:07, D14:02 \\
Recall@5: 0/3; prediction: \textbf{B} &
Recall@5: 2/3; prediction: \textbf{C} (gold) \\
\end{tabular}

\medskip
\textbf{Failure analysis.}
The Text-Index explanation accurately recognized a swimming coach or former
athlete, but none of the three annotated clue sessions appeared in its top
five.  It consequently substituted generic professional tools---a pristine
whistle and stopwatch---for the unobserved private habit and chose B.  The
Multimodal-Index placed \emph{D02:02} first and \emph{D14:02} fifth; once the
bag and worn training objects were visible, the answer changed to C.  This is
a retrieval miss followed by stereotype-based gap filling, rather than an
answer error in the presence of the same evidence.
\end{mybox}
\end{minipage}%
}
\caption{Retrieval failure case R1.  The side-by-side lists are the stored
top-five turn IDs from the Qwen3-Omni RQ3 runs.}
\label{fig:failure-retrieval-coach}
\end{figure}

\section{Failure and Qualitative Analysis}
\label{app:error_analysis}

\subsection{Failure Taxonomy}
We use the following qualitative taxonomy to separate failures at different
stages of a memory pipeline.  The categories are diagnostic rather than
mutually exclusive: an omitted visual detail can cause a retrieval miss, and
the resulting evidence gap can in turn invite an unsupported inference.

\paragraph{Textualization miss.}
A target-bearing visual detail or non-speech sound is absent, distorted, or
made too vague when the raw modality is converted to text.  The clue remains
visible or audible in the source memory but cannot be recovered from the
caption or transcript supplied to the memory system.

\paragraph{Cross-session association failure.}
Individually relevant occurrences are stored, but are not consolidated as
recurrences of the same entity, preference, or habit.  The model consequently
reasons from one occurrence or disconnected facts instead of using their
longitudinal agreement.

\paragraph{Retrieval miss.}
The memory contains the evidence, but none---or too little---of it appears in
the retrieved context at the evaluated cutoff.  Its direct symptom is low
Recall@K; a representation that restores the clue may also change the answer.

\paragraph{Retrieval noise or distractor dominance.}
Relevant evidence is accompanied or displaced by semantically plausible but
non-diagnostic context.  The explanation may faithfully cite retrieved
memories while supporting a stereotype or a competing preference rather than
the target.

\paragraph{Evidence-use failure.}
The required clue is present in context, but the answer model overlooks it,
combines it incorrectly, or fails to map it to an option.  Here retrieval and
the gold option agree, yet the reasoning selects an incompatible answer.

\paragraph{Unsupported inference or refusal-calibration error.}
The model supplies a specific claim when the history is insufficient, or
refuses despite adequate evidence.  A plausible narrative may be preferred
to ``not specified,'' or a supported answer rejected because its evidence is
indirect.  We do not report frequencies for these categories because the
cases below were selected for diagnosis rather than drawn as a
frequency-estimation sample.

The cases below preserve the question, media-bearing memory turns, retrieved
turn identifiers, model answer, and answer-model reasoning.  They are chosen
because the error can be localized by a controlled representation change;
they should not be read as estimates of how often each failure occurs.

\subsection{Retrieval Failure Cases}

\begin{figure*}[!htbp]
\centering
\begin{mybox}
\textbf{Case R2---A salient travel strand overwhelms a recurring visual interest}
\hfill \textbf{p15 / 15-HobbiesAndEntertainment-1 / implicit}

\medskip
\textbf{Gold memory clues.}
The foreground dialogue alternates between travel and science-fiction books:
\emph{D06} and \emph{D26} discuss maps, while \emph{D12} and \emph{D18}
discuss frequently reread novels and a shelf organized ``like an exhibition.''
Across the images, large-format photography volumes recur at the edge of the
bookshelf.  This peripheral visual strand supports an art-book fair, whereas
the map strand creates a plausible travel-bookstore distractor.

\medskip
  \begin{minipage}[t]{0.19\linewidth}\centering
  \IfFileExists{supplementary_images/event/images/pid_0015_task_15-11-0.jpg}{%
    \includegraphics[width=\linewidth,height=2.45cm,keepaspectratio]{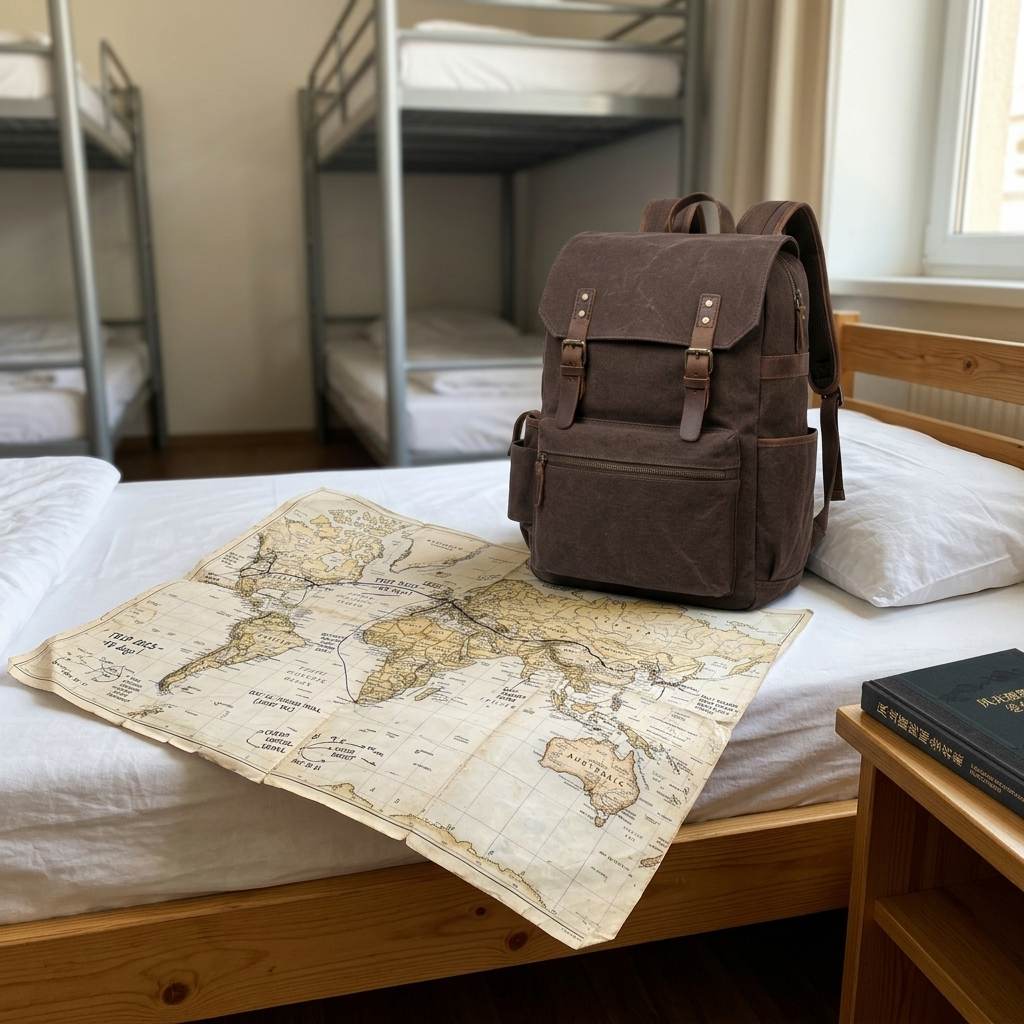}%
  }{%
    \fbox{\parbox[c][2.25cm][c]{0.86\linewidth}{\centering\scriptsize
    Memory image unavailable}}%
  }\\[0.45ex]
  \scriptsize\textbf{D06-001.png}\\[-0.2ex]backpack and map
  \end{minipage}
  \begin{minipage}[t]{0.19\linewidth}\centering
  \IfFileExists{supplementary_images/event/images/pid_0015_task_15-2-0.jpg}{%
    \includegraphics[width=\linewidth,height=2.45cm,keepaspectratio]{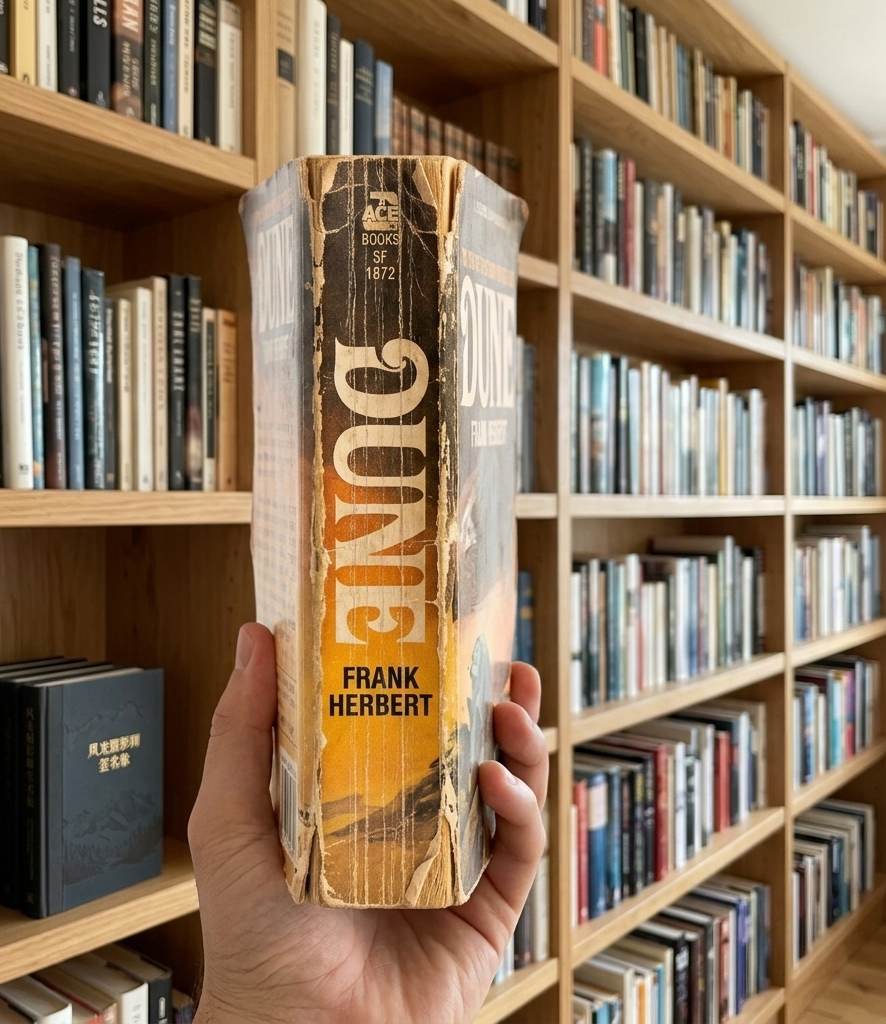}%
  }{%
    \fbox{\parbox[c][2.25cm][c]{0.86\linewidth}{\centering\scriptsize
    Memory image unavailable}}%
  }\\[0.45ex]
  \scriptsize\textbf{D12-001.png}\\[-0.2ex]worn novel; art books behind
  \end{minipage}
  \begin{minipage}[t]{0.19\linewidth}\centering
  \IfFileExists{supplementary_images/event/images/pid_0015_task_15-3-0.jpg}{%
    \includegraphics[width=\linewidth,height=2.45cm,keepaspectratio]{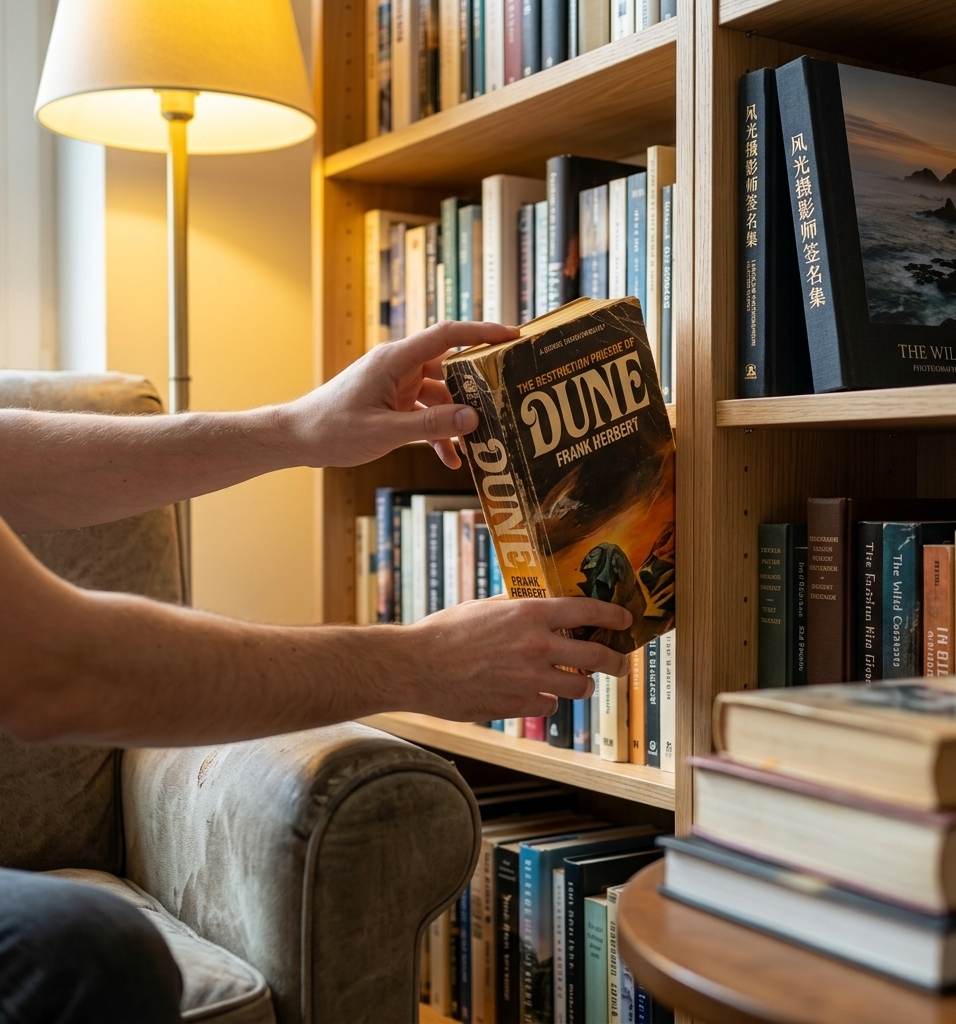}%
  }{%
    \fbox{\parbox[c][2.25cm][c]{0.86\linewidth}{\centering\scriptsize
    Memory image unavailable}}%
  }\\[0.45ex]
  \scriptsize\textbf{D18-001.png}\\[-0.2ex]curated bookshelf
  \end{minipage}
  \begin{minipage}[t]{0.19\linewidth}\centering
  \IfFileExists{supplementary_images/event/images/pid_0015_task_15-22-0.jpg}{%
    \includegraphics[width=\linewidth,height=2.45cm,keepaspectratio]{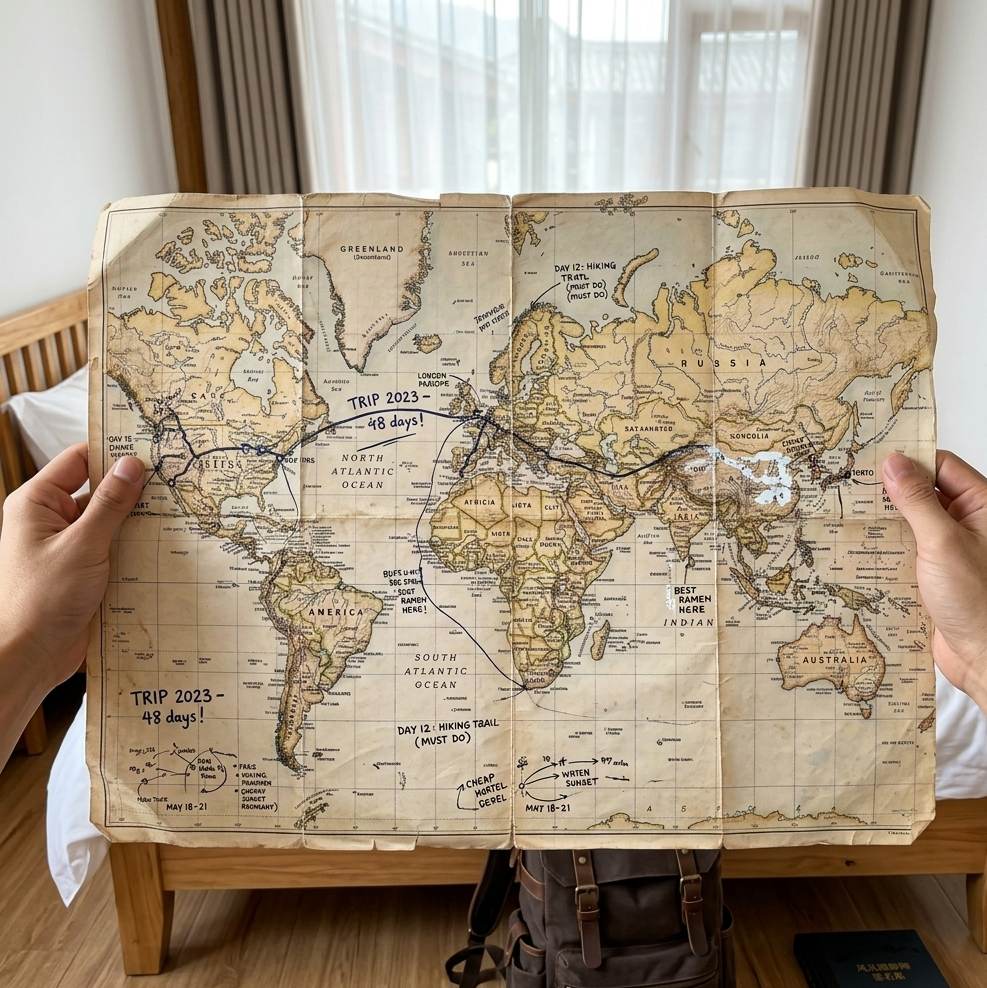}%
  }{%
    \fbox{\parbox[c][2.25cm][c]{0.86\linewidth}{\centering\scriptsize
    Memory image unavailable}}%
  }\\[0.45ex]
  \scriptsize\textbf{D26-001.png}\\[-0.2ex]annotated travel map
  \end{minipage}

\medskip
\textbf{Question.} Which pictured weekend destination would this user most
likely enjoy?

\medskip
  \begin{minipage}[t]{0.235\linewidth}\centering
  \IfFileExists{supplementary_images/qa/rec_images/15-HobbiesAndEntertainment-1_A.jpg}{%
    \includegraphics[width=\linewidth,height=2.65cm,keepaspectratio]{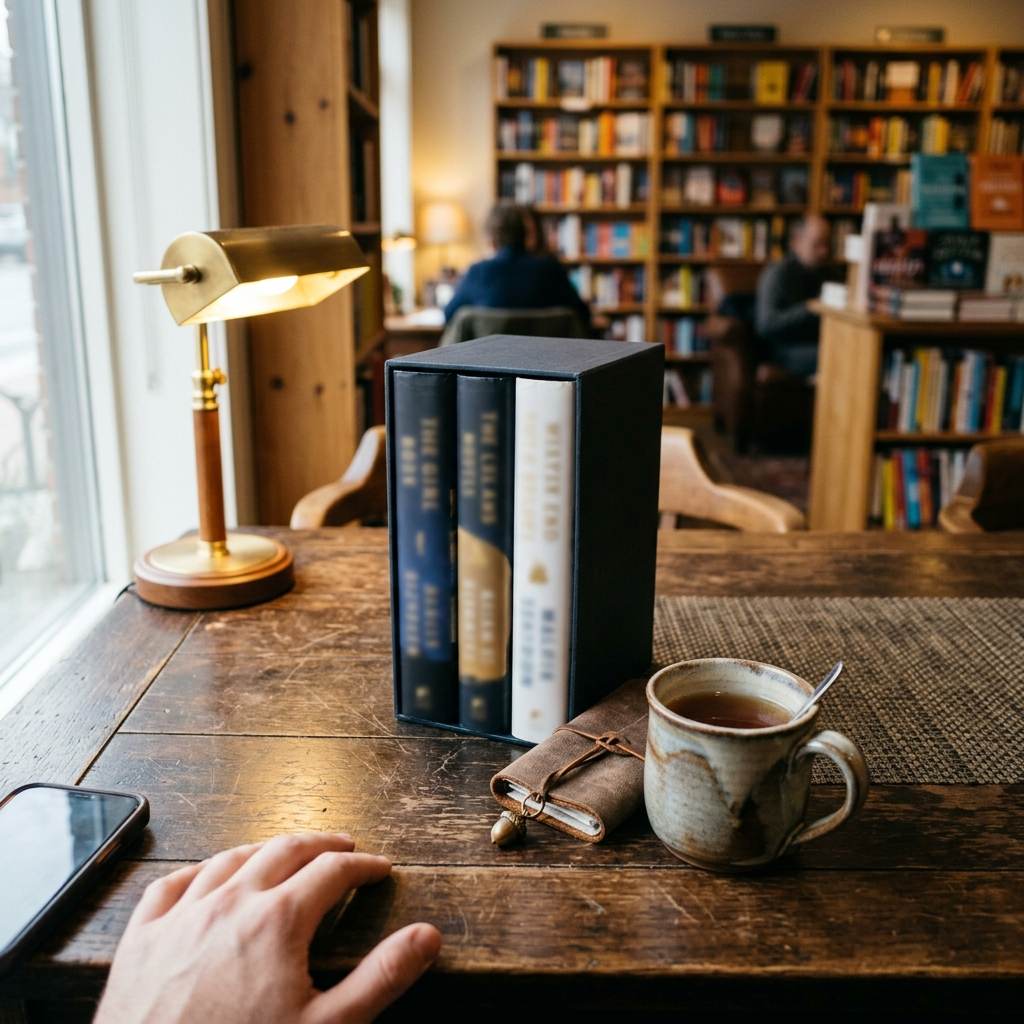}%
  }{%
    \fbox{\parbox[c][2.45cm][c]{0.88\linewidth}{\centering\scriptsize
    Option image unavailable}}%
  }\\[0.45ex]
  \scriptsize\textbf{A}\enspace general bookstore
  \end{minipage}
  \begin{minipage}[t]{0.235\linewidth}\centering
  \IfFileExists{supplementary_images/qa/rec_images/15-HobbiesAndEntertainment-1_B.jpg}{%
    \includegraphics[width=\linewidth,height=2.65cm,keepaspectratio]{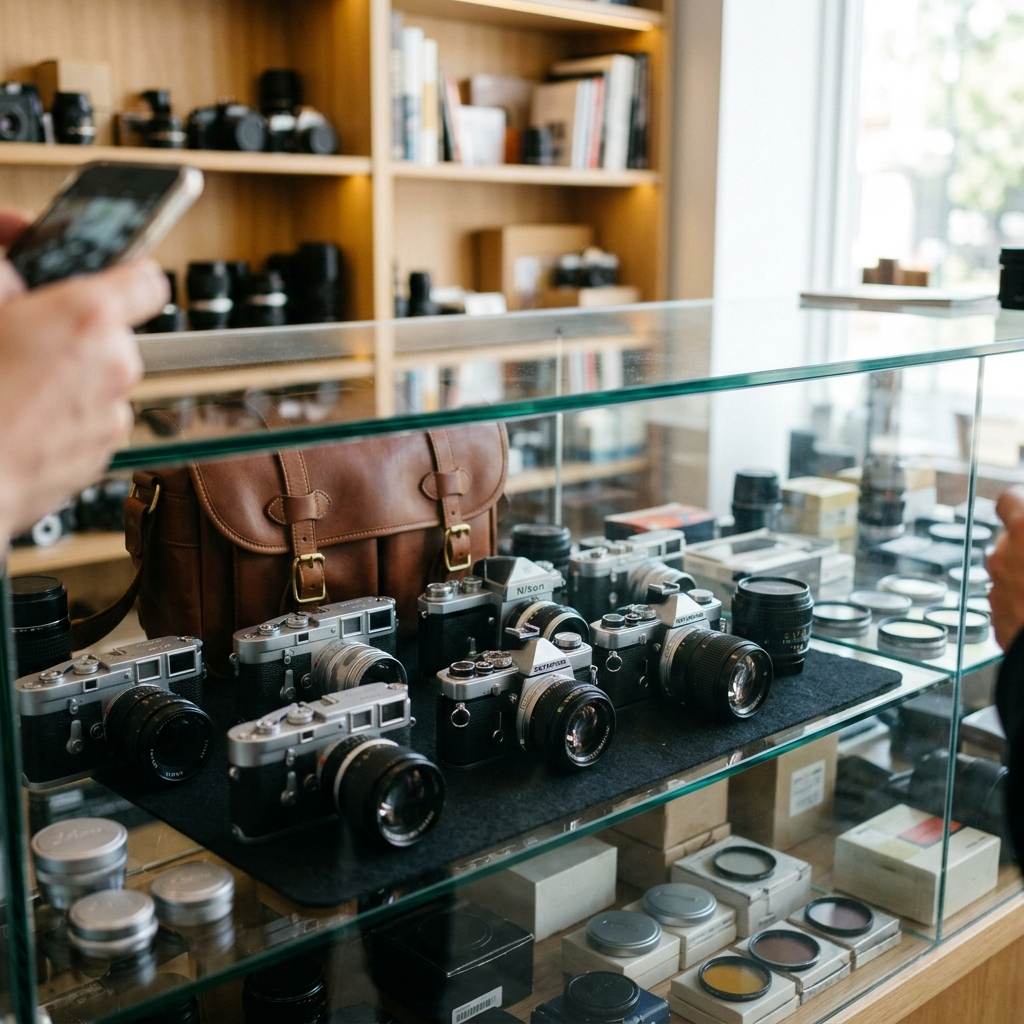}%
  }{%
    \fbox{\parbox[c][2.45cm][c]{0.88\linewidth}{\centering\scriptsize
    Option image unavailable}}%
  }\\[0.45ex]
  \scriptsize\textbf{B}\enspace camera-equipment shop
  \end{minipage}
  \begin{minipage}[t]{0.235\linewidth}\centering
  \IfFileExists{supplementary_images/qa/rec_images/15-HobbiesAndEntertainment-1_C.jpg}{%
    \includegraphics[width=\linewidth,height=2.65cm,keepaspectratio]{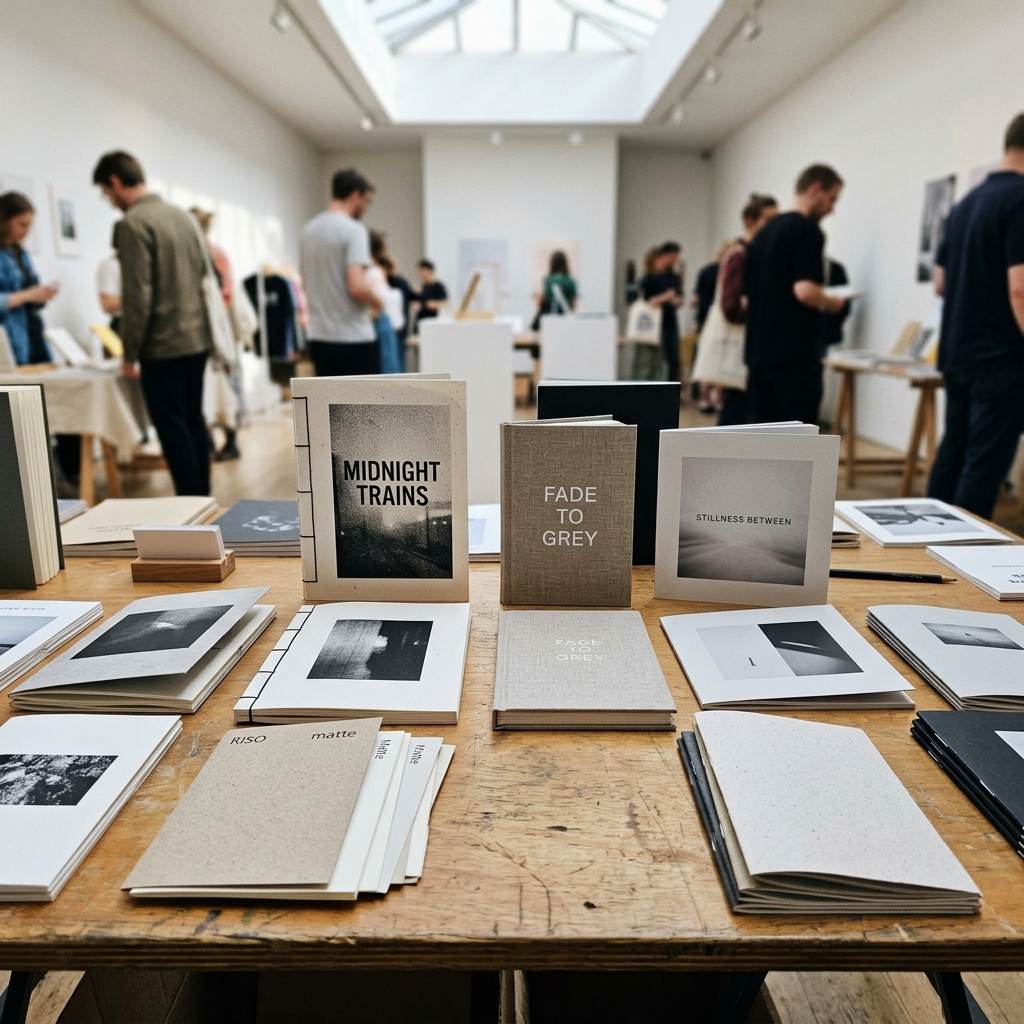}%
  }{%
    \fbox{\parbox[c][2.45cm][c]{0.88\linewidth}{\centering\scriptsize
    Option image unavailable}}%
  }\\[0.45ex]
  \scriptsize\textbf{C}\enspace independent art-book fair
  \end{minipage}
  \begin{minipage}[t]{0.235\linewidth}\centering
  \IfFileExists{supplementary_images/qa/rec_images/15-HobbiesAndEntertainment-1_D.jpg}{%
    \includegraphics[width=\linewidth,height=2.65cm,keepaspectratio]{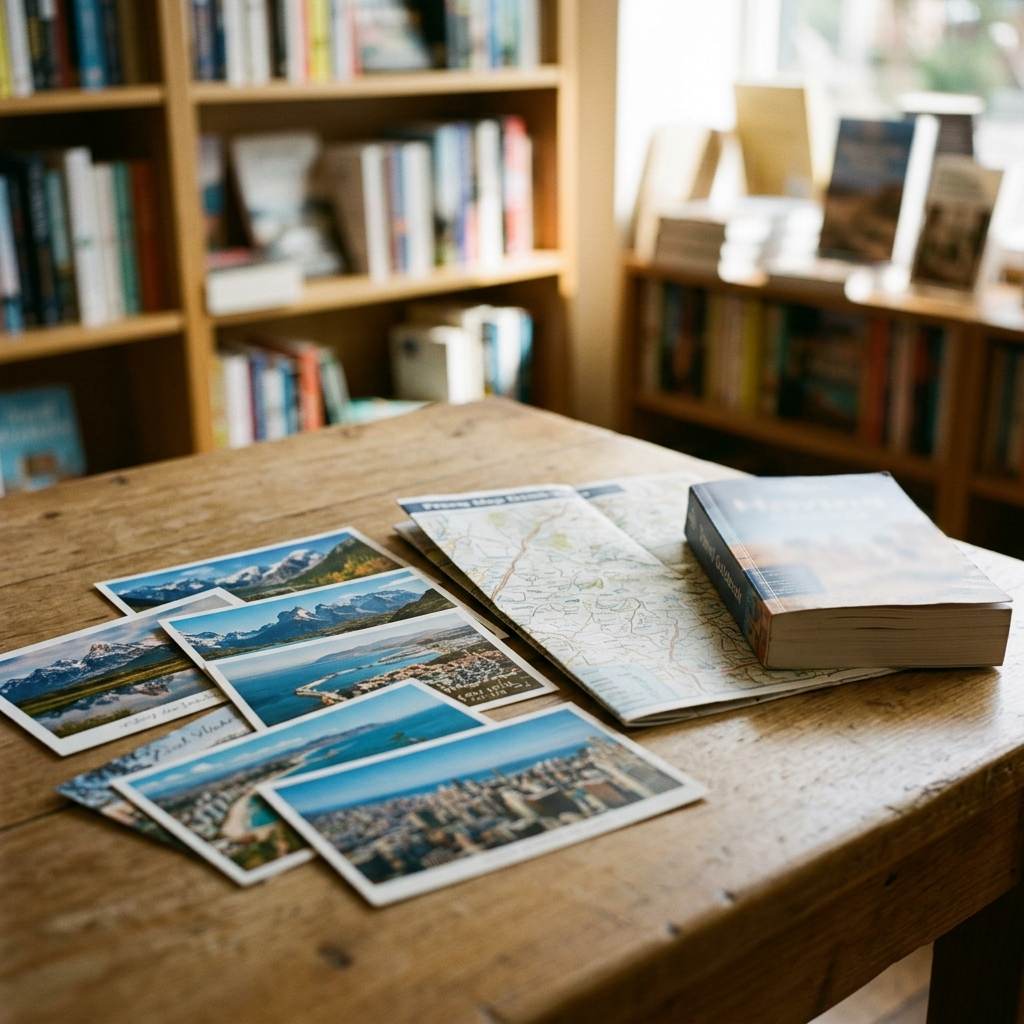}%
  }{%
    \fbox{\parbox[c][2.45cm][c]{0.88\linewidth}{\centering\scriptsize
    Option image unavailable}}%
  }\\[0.45ex]
  \scriptsize\textbf{D}\enspace travel bookstore
  \end{minipage}

\medskip
\begin{tabular}{@{}p{0.485\linewidth}p{0.485\linewidth}@{}}
\textbf{Text-Index} & \textbf{Multimodal-Index} \\
Top-5: D15:07, D15:06, D07:07, D10:10, D19:06 &
Top-5: D07:00, D18:03, D15:07, D12:00, D26:00 \\
Recall@5: 0/4; prediction: \textbf{D} &
Recall@5: 3/4; prediction: \textbf{C} (gold) \\
\end{tabular}

\medskip
\textbf{Failure analysis.}
The Text-Index retrieved no annotated clue in its top five.  Its reasoning
then emphasized hand-drawn routes and nonstandard travel, selecting D---a
coherent recommendation for a real but competing preference.  The
Multimodal-Index recovered \emph{D18:03}, \emph{D12:00}, and \emph{D26:00};
its explanation combined the curated shelf, deep reading, and visual-art
interest and selected C.  The contrast illustrates distractor dominance:
plausibility is insufficient when retrieval covers only one of several
longitudinal preference strands.
\end{mybox}
\caption{Retrieval failure case R2.  Multimodal retrieval restores the
peripheral bookshelf evidence needed to distinguish two plausible venues.}
\label{fig:failure-retrieval-artbooks}
\end{figure*}

\subsection{Textualization Failure Cases}

\begin{figure*}[!htbp]
\centering
\begin{mybox}
\textbf{Case T1---A peripheral magazine identity disappears in a brief caption}
\hfill \textbf{p0 / 0-Items-6-img\_identify / LTMemory / implicit}

\smallskip
\textbf{Memory clues.}
The same dog-eared issue of \emph{Wallpaper*} appears incidentally in five
sessions.  The conversations concern a mood board (\emph{D06}), a colleague's
analysis (\emph{D12}), cats (\emph{D17}, \emph{D21}), and route planning
(\emph{D28}); none names the magazine.  Thus its identity must survive image
textualization.

\smallskip
  \begin{minipage}[t]{0.19\linewidth}\centering
  \IfFileExists{supplementary_images/event/images/pid_0000_task_0-13-0.jpg}{%
    \includegraphics[width=\linewidth,height=1.6cm,keepaspectratio]{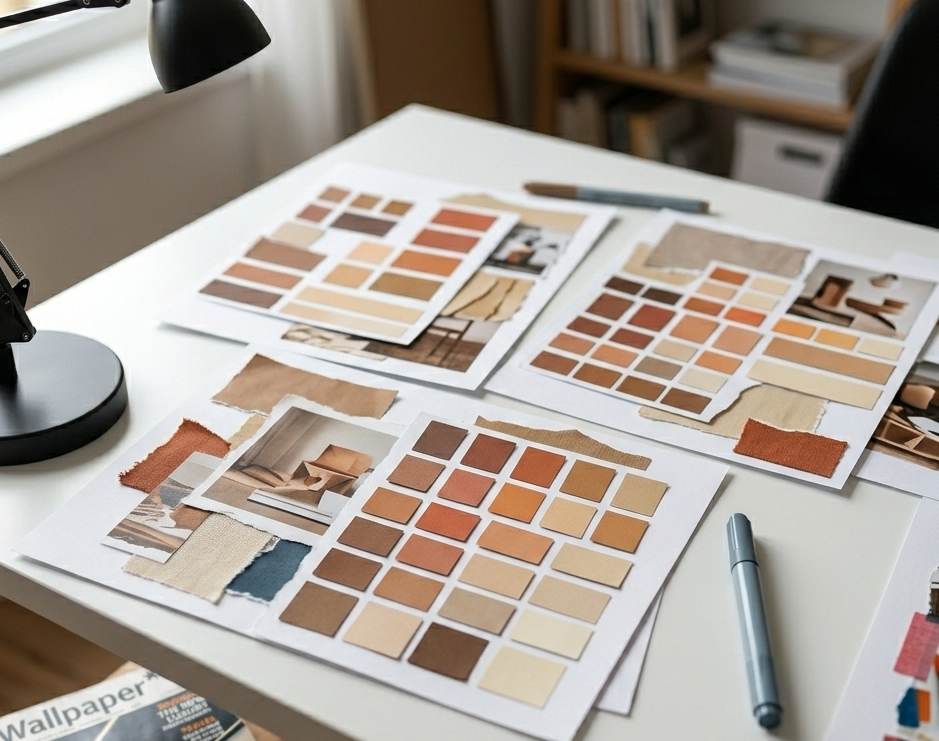}%
  }{%
    \fbox{\parbox[c][2.25cm][c]{0.86\linewidth}{\centering\scriptsize
    Memory image unavailable}}%
  }\\[0.45ex]
  \scriptsize\textbf{D06-001.png}\\[-0.2ex]mood-board desk
  \end{minipage}
  \begin{minipage}[t]{0.19\linewidth}\centering
  \IfFileExists{supplementary_images/event/images/pid_0000_task_0-20-0.jpg}{%
    \includegraphics[width=\linewidth,height=1.6cm,keepaspectratio]{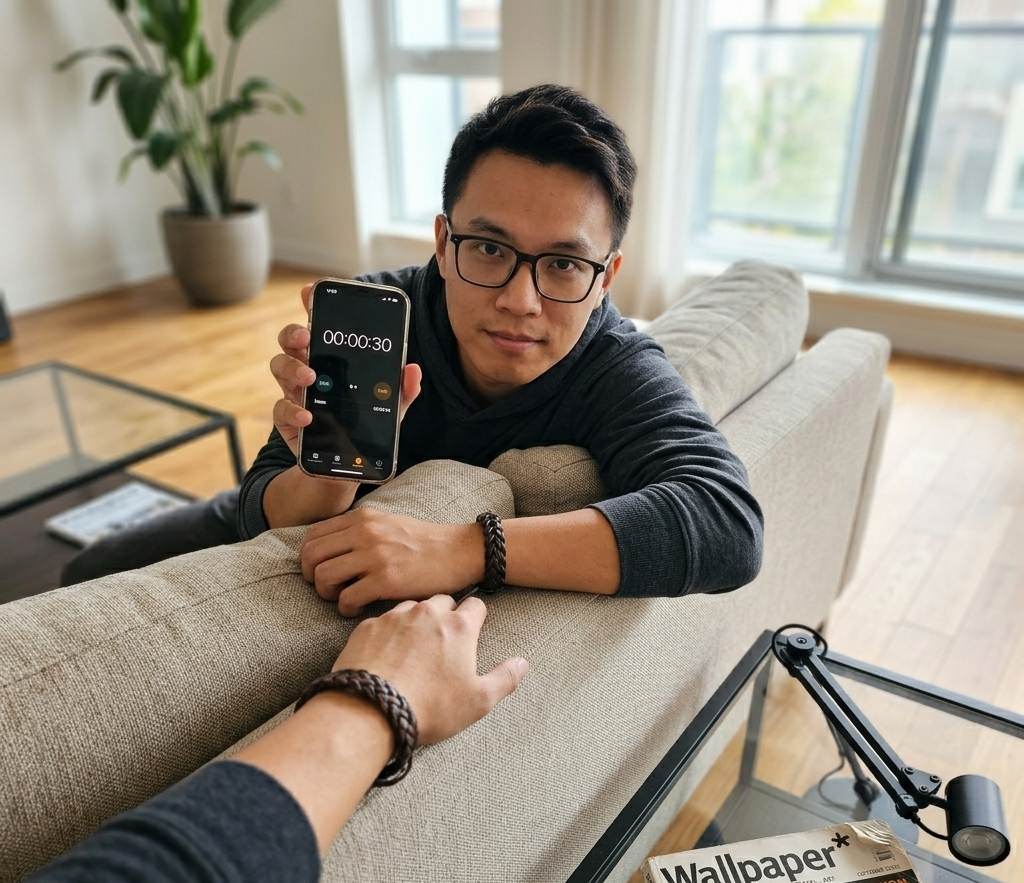}%
  }{%
    \fbox{\parbox[c][2.25cm][c]{0.86\linewidth}{\centering\scriptsize
    Memory image unavailable}}%
  }\\[0.45ex]
  \scriptsize\textbf{D12-001.png}\\[-0.2ex]living-room table
  \end{minipage}
  \begin{minipage}[t]{0.19\linewidth}\centering
  \IfFileExists{supplementary_images/event/images/pid_0000_task_0-6-0.jpg}{%
    \includegraphics[width=\linewidth,height=1.6cm,keepaspectratio]{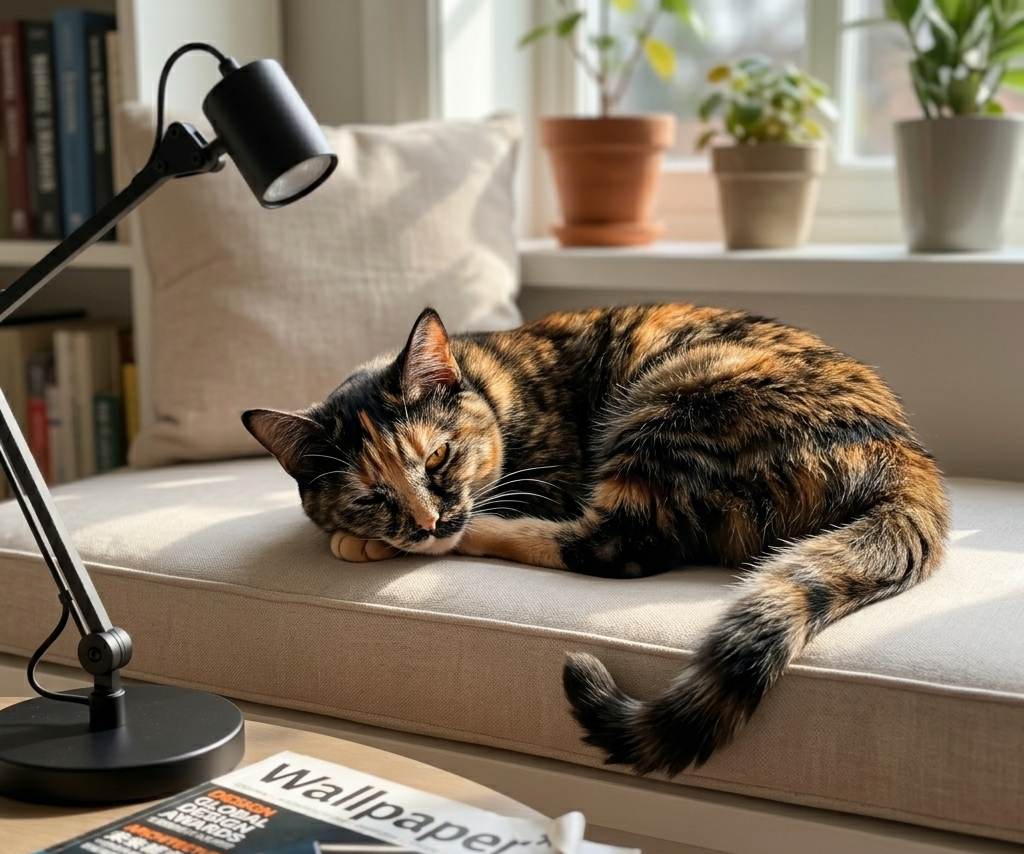}%
  }{%
    \fbox{\parbox[c][2.25cm][c]{0.86\linewidth}{\centering\scriptsize
    Memory image unavailable}}%
  }\\[0.45ex]
  \scriptsize\textbf{D17-001.png}\\[-0.2ex]beside resting cat
  \end{minipage}
  \begin{minipage}[t]{0.19\linewidth}\centering
  \IfFileExists{supplementary_images/event/images/pid_0000_task_0-2-0.jpg}{%
    \includegraphics[width=\linewidth,height=1.6cm,keepaspectratio]{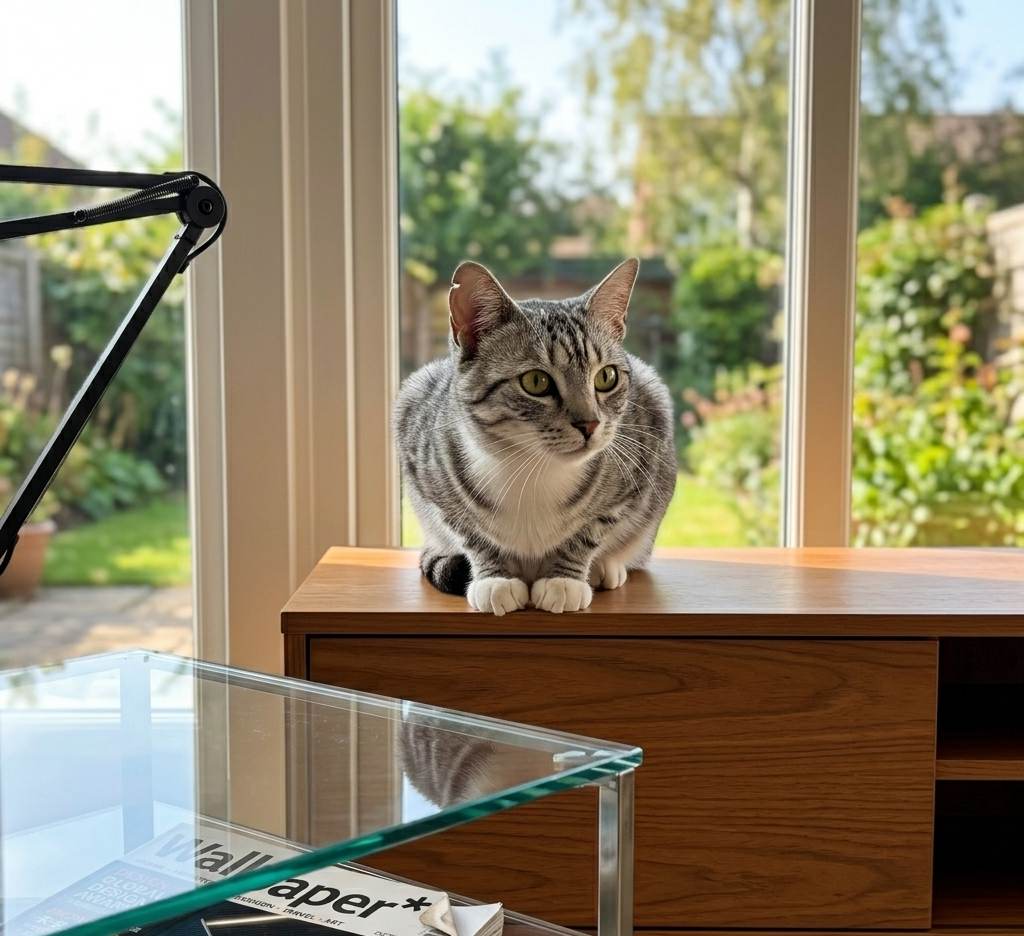}%
  }{%
    \fbox{\parbox[c][2.25cm][c]{0.86\linewidth}{\centering\scriptsize
    Memory image unavailable}}%
  }\\[0.45ex]
  \scriptsize\textbf{D21-001.png}\\[-0.2ex]foreground table
  \end{minipage}
  \begin{minipage}[t]{0.19\linewidth}\centering
  \IfFileExists{supplementary_images/event/images/pid_0000_task_0-18-0.jpg}{%
    \includegraphics[width=\linewidth,height=1.6cm,keepaspectratio]{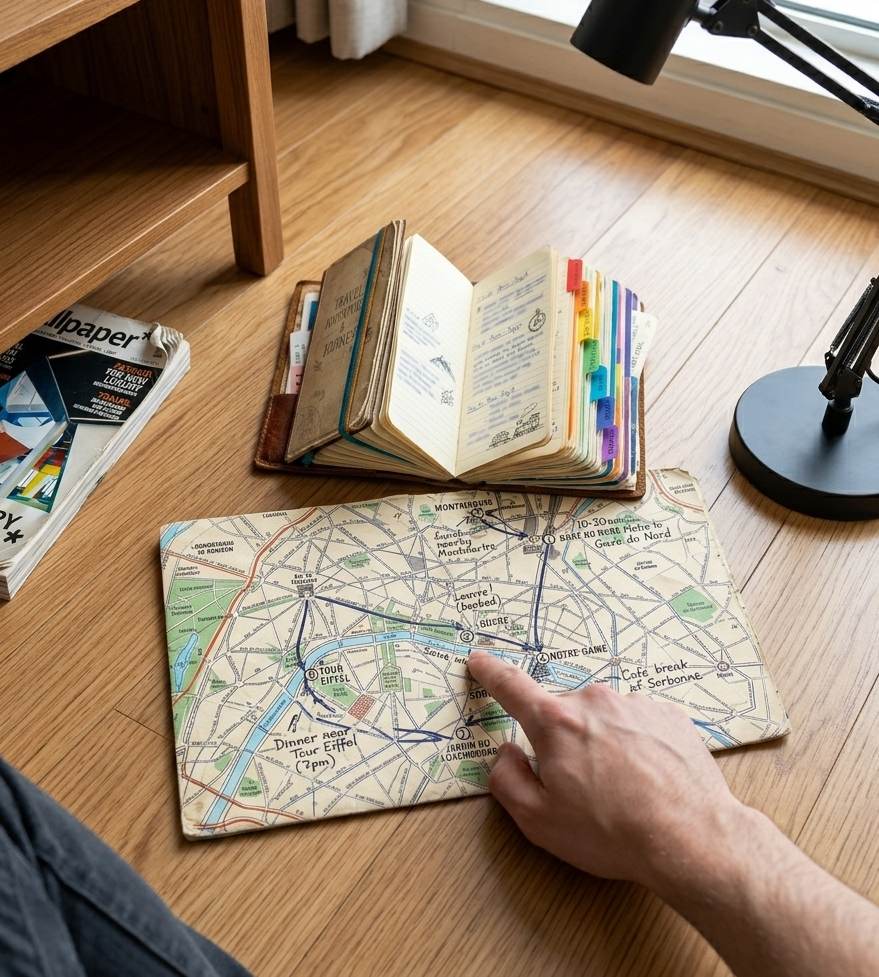}%
  }{%
    \fbox{\parbox[c][2.25cm][c]{0.86\linewidth}{\centering\scriptsize
    Memory image unavailable}}%
  }\\[0.45ex]
  \scriptsize\textbf{D28-001.png}\\[-0.2ex]route-planning floor
  \end{minipage}

\smallskip
\begin{tabular}{@{}p{0.485\linewidth}p{0.485\linewidth}@{}}
\textbf{Brief caption of D06-001} & \textbf{Detailed caption of D06-001} \\
``Color sample cards, fabric samples, and a pen are arranged on a white
desk.'' &
``At the lower-left edge is part of a magazine whose cover reads
\emph{Wallpaper*}.'' \\
Target status: \textbf{omitted} & Target status: \textbf{retained} \\
\end{tabular}

\smallskip
\textbf{Question.} Which image is the user's magazine?

\smallskip
  \begin{minipage}[t]{0.235\linewidth}\centering
  \IfFileExists{supplementary_images/qa/entity_images/0-Items-6-img_identify_A.jpg}{%
    \includegraphics[width=\linewidth,height=1.8cm,keepaspectratio]{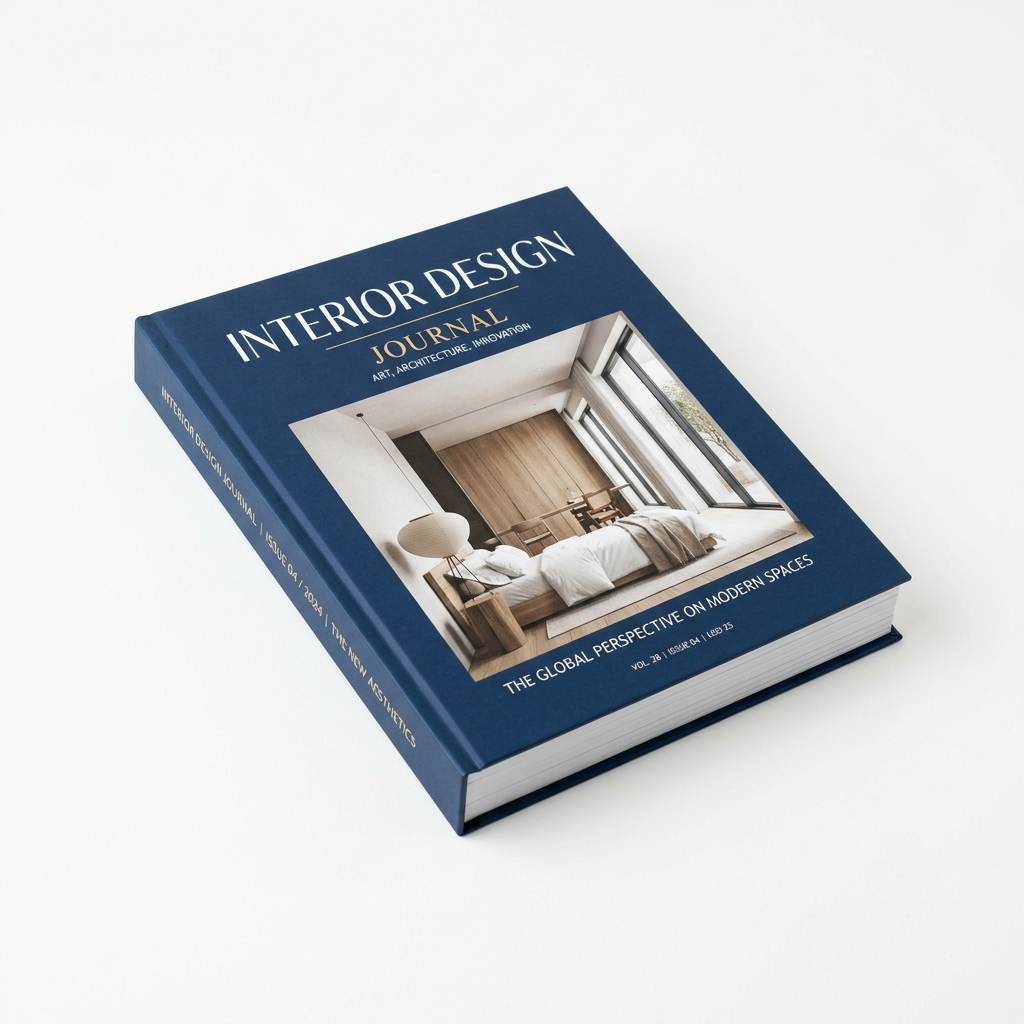}%
  }{%
    \fbox{\parbox[c][2.45cm][c]{0.88\linewidth}{\centering\scriptsize
    Option image unavailable}}%
  }\\[0.45ex]
  \scriptsize\textbf{A}\enspace blue design journal
  \end{minipage}
  \begin{minipage}[t]{0.235\linewidth}\centering
  \IfFileExists{supplementary_images/profile/generated_portraits/profile_0_items_wallpaper_magazine_12.jpg}{%
    \includegraphics[width=\linewidth,height=1.8cm,keepaspectratio]{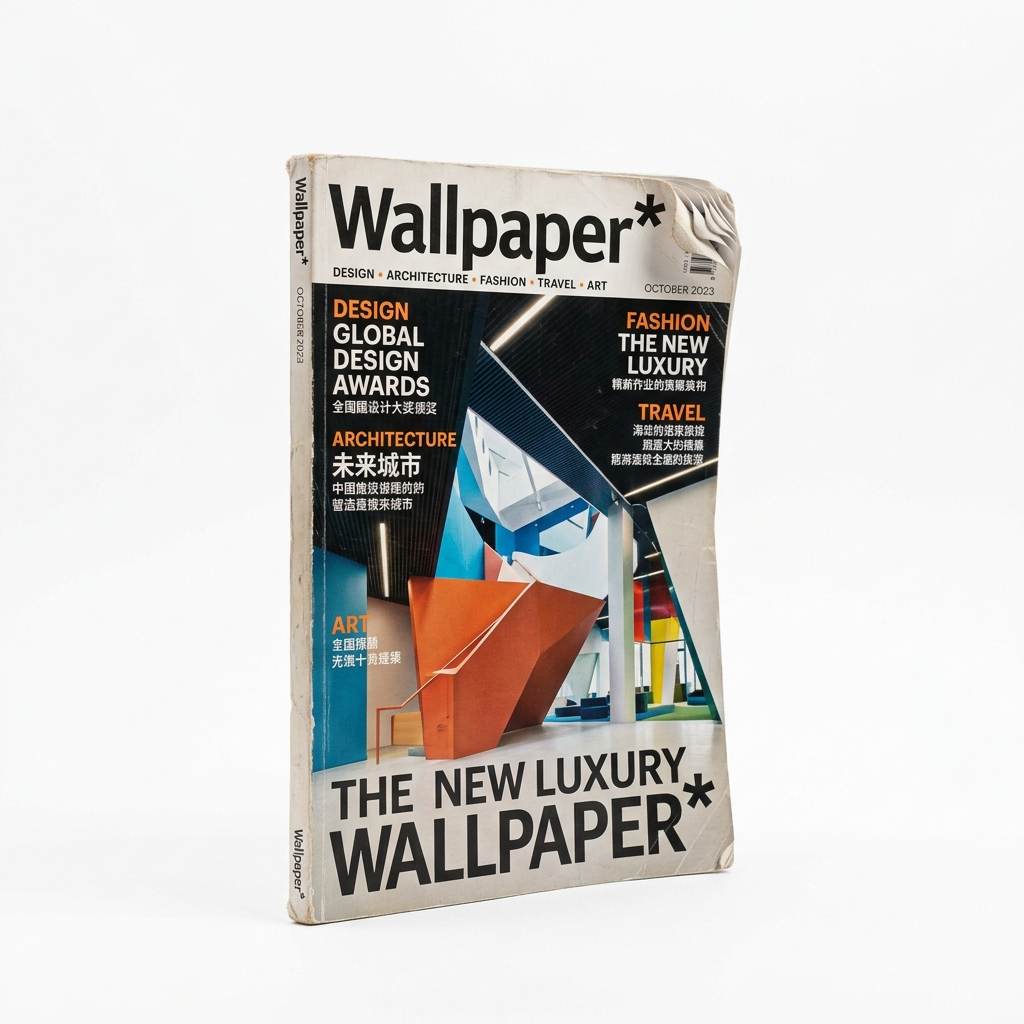}%
  }{%
    \fbox{\parbox[c][2.45cm][c]{0.88\linewidth}{\centering\scriptsize
    Option image unavailable}}%
  }\\[0.45ex]
  \scriptsize\textbf{B}\enspace dog-eared \emph{Wallpaper*}
  \end{minipage}
  \begin{minipage}[t]{0.235\linewidth}\centering
  \IfFileExists{supplementary_images/qa/entity_images/0-Items-6-img_identify_C.jpg}{%
    \includegraphics[width=\linewidth,height=1.8cm,keepaspectratio]{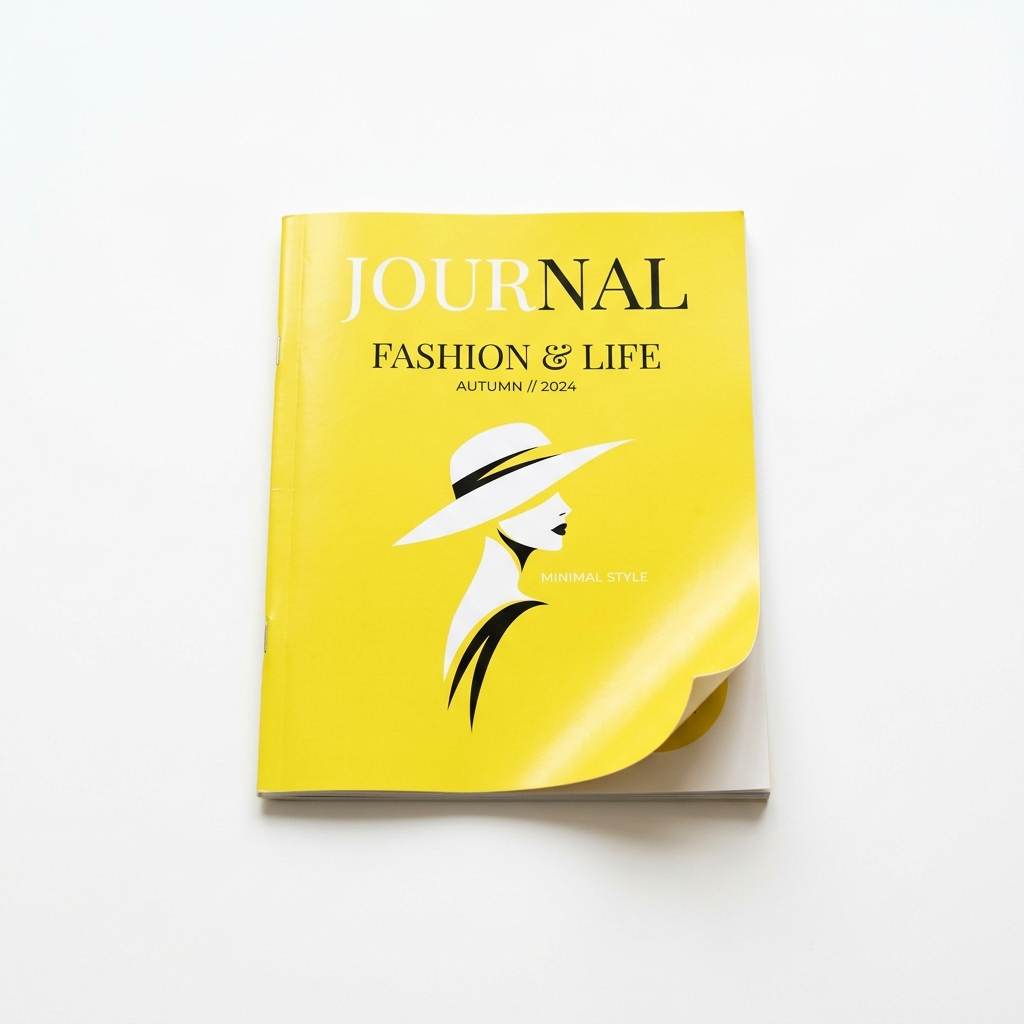}%
  }{%
    \fbox{\parbox[c][2.45cm][c]{0.88\linewidth}{\centering\scriptsize
    Option image unavailable}}%
  }\\[0.45ex]
  \scriptsize\textbf{C}\enspace yellow fashion magazine
  \end{minipage}
  \begin{minipage}[t]{0.235\linewidth}\centering
  \IfFileExists{supplementary_images/qa/entity_images/0-Items-6-img_identify_D.jpg}{%
    \includegraphics[width=\linewidth,height=1.8cm,keepaspectratio]{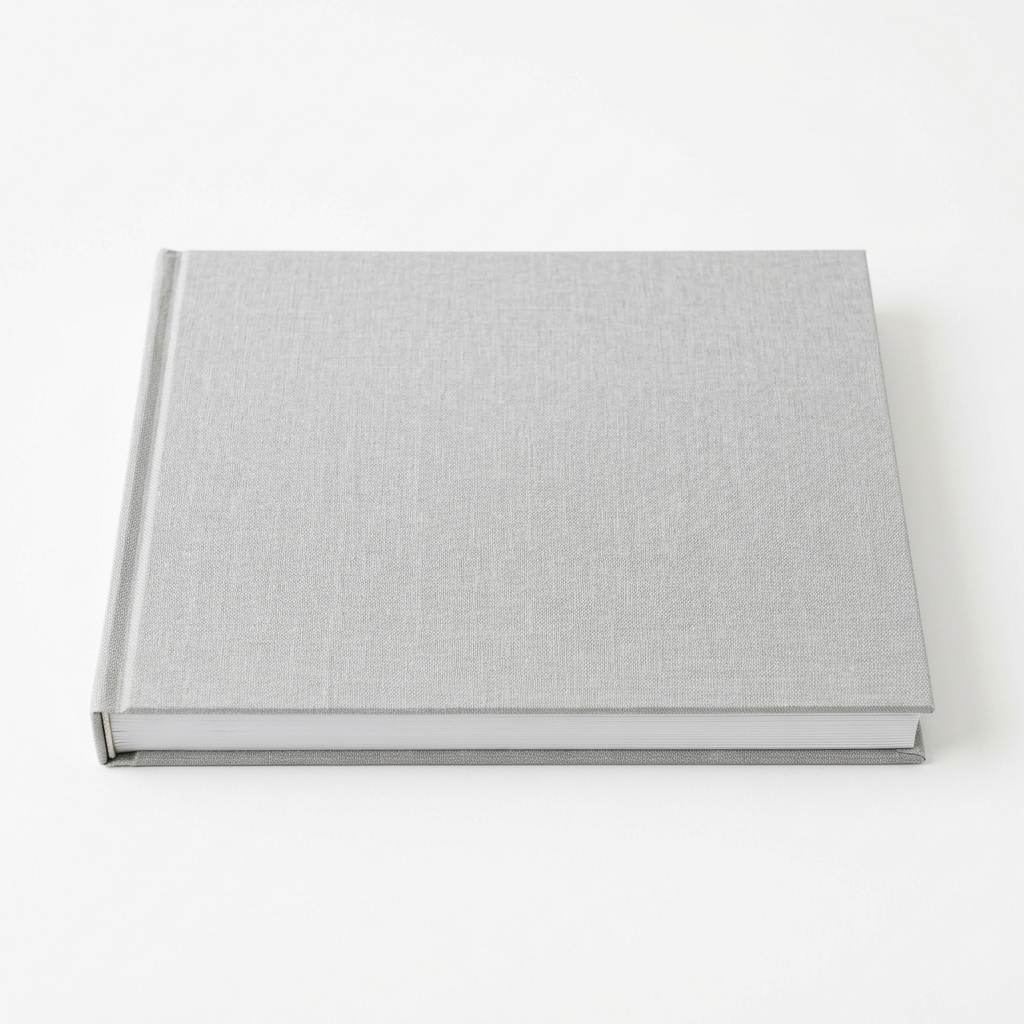}%
  }{%
    \fbox{\parbox[c][2.45cm][c]{0.88\linewidth}{\centering\scriptsize
    Option image unavailable}}%
  }\\[0.45ex]
  \scriptsize\textbf{D}\enspace gray art book
  \end{minipage}

\smallskip
\textbf{Observed answer change.}
With brief captions, Qwen3.6-35B-A3B selected \textbf{D}; its reasoning
explicitly noted that the memory captions supplied neither a cover color nor
a title and then guessed among underdetermined options.  With detailed
captions, it selected \textbf{B} (gold) and cited the visible
``\emph{Wallpaper*}'' title in \emph{D06}.  The raw image is unchanged, so the
failure is localized to an encoding omission.  Detailed prose helps here not
because length is intrinsically beneficial, but because it preserves the
small, peripheral identifier required by the question.
\end{mybox}
\caption{Image-textualization failure.  A single retained edge detail changes
an underdetermined entity match into a supported answer.}
\label{fig:failure-textualization-image}
\end{figure*}

\begin{figure*}[!htbp]
\centering
\begin{mybox}
\textbf{Case T2---Speech-only transcription removes the target audio channel}
\hfill \textbf{p0 / 0-FoodAndDrink-1 / LTMemory / implicit}

\smallskip
\textbf{Memory clues and foreground content.}
Cooking sounds recur behind unrelated or only loosely related conversations.
For example, \emph{D25-004--006.wav} discusses blackout curtains, earplugs,
and window noise while hot-oil sizzling and metal-on-wok impacts continue in
the background.  \emph{D19-003--006.wav} discusses how a colleague organizes
the user's bookshelf; its foreground image likewise depicts the colleague and
books, not cooking.  In \emph{D09-003--006.wav}, the user mentions a hot meal
while the background progresses from food entering hot oil, through wok--spatula
impacts, to the low hum of an exhaust hood.

\smallskip
  \begin{minipage}[t]{0.31\linewidth}\vspace{0pt}\centering
  \IfFileExists{supplementary_images/event/images/pid_0000_task_0-23-0.jpg}{%
    \includegraphics[width=\linewidth,height=1.5cm,keepaspectratio]{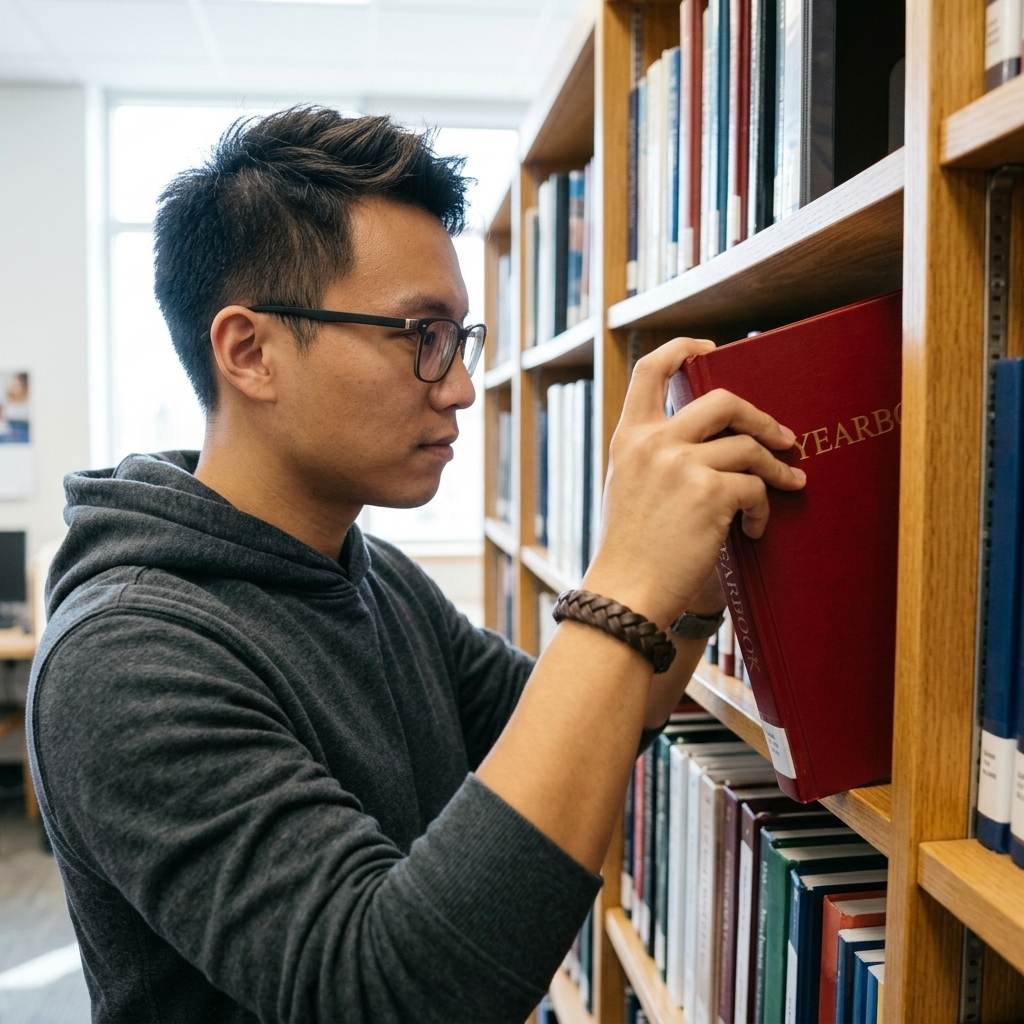}%
  }{%
    \fbox{\parbox[c][2.9cm][c]{0.88\linewidth}{\centering\scriptsize
    Original memory image\\not in transferred snapshot}}%
  }\\[0.45ex]
  \scriptsize\textbf{D19-001.png}\enspace foreground: colleague organizing books
  \end{minipage}
\hfill
\fbox{\parbox[t]{0.61\linewidth}{\footnotesize
\textbf{Representative mixed-waveform content}\par
\emph{Foreground speech (D25):} ``The blackout lining is the key 
\ldots{} I may replace the window seal to reduce wind noise.''\par
\emph{Background (D25):} sustained hot-oil sizzling; frequent metal spatula
impacts and scraping against a wok.\par
\emph{Background (D09):} food hits hot oil $\rightarrow$ wok--spatula impacts
$\rightarrow$ continuous exhaust-hood hum.}}

\smallskip
\begin{tabular}{@{}p{0.485\linewidth}p{0.485\linewidth}@{}}
\textbf{ASR-only textualization} & \textbf{Speech/background Split} \\
Retains the spoken discussion but not a description of the non-speech
channel. &
Stores foreground speech and an explicit ambient-sound description in
separate fields. \\
Model reasoning cites slow eating, texture, and ``refined'' food preferences;
prediction: \textbf{D}, low-temperature cooker. &
Model reasoning cites hot oil, wok--spatula impacts, and exhaust-hood hum;
prediction: \textbf{C}, high-suction exhaust hood (gold). \\
\end{tabular}

\smallskip
\textbf{Question.} Which kitchen appliance best fits the user's needs?

\smallskip
\begin{tabular}{@{}p{0.48\linewidth}p{0.48\linewidth}@{}}
A. Smart electric stew pot & B. Multifunction breakfast maker \\
C. High-suction exhaust hood & D. Low-temperature cooker \\
\end{tabular}

\smallskip
\textbf{Failure analysis.}
The ASR answer is plausible but rests on memories unrelated to the annotated
ambient clue.  Once the background channel is textualized separately, the
reasoning shifts to the repeated high-heat cooking and ventilation evidence
and reaches C.  This case distinguishes speech recognition quality from
multimodal memory fidelity: a correct transcript can still be an incomplete
representation of an audio memory when the target is non-speech.
\end{mybox}
\caption{Audio-textualization failure.  The same Qwen3.6-35B-A3B and memory
system change answer when the non-speech channel is retained.}
\label{fig:failure-textualization-audio}
\end{figure*}

\section{Ethical Considerations}
\label{app:ethics}

\paragraph{Synthetic identities and source media.}
CUE-Mem contains synthetic users and generated multi-session histories rather
than conversations collected from real users.  Persona seeds from ALOE and
PersonaHub are expanded into new profiles, and the names, recurring identity
assets, dialogue, portraits, scene images, and user speech are synthesized for
the benchmark.  In particular, each voice is produced from a generated voice
design and anchor, rather than by cloning a recording of a real participant.
This construction substantially reduces direct exposure of private
conversation histories, faces, and voices.  It does not guarantee that a
generated face or profile can never resemble a real person by coincidence,
so synthetic identities should not be interpreted as depictions of actual
individuals.

\paragraph{Privacy implications of implicit memory.}
The benchmark deliberately tests whether systems can infer stable user
information from incidental objects in shared images and non-speech sounds in
voice messages.  The same capability could be intrusive outside this
controlled setting: household composition, routines, interests, and living
conditions may be inferred even when a user did not state them or make them
the subject of an exchange.  Recurrence does not by itself make such an
inference accurate or consented to.  Accordingly, strong benchmark
performance should not be treated as permission to retain or profile
background information from real interactions.  A deployed system using this
capability would need explicit user control over collection, retention, and
deletion, as well as a conservative threshold for turning weak cues into
persistent memories.

\paragraph{Stereotypes and uncertain inference.}
Profiles and events are generated by language and media models from persona
seeds.  They may reproduce culturally specific or stereotyped associations
between age, gender, occupation, household setting, appearance, voice, and
lifestyle.  Moreover, a peripheral object or ambient sound often admits more
than one explanation.  CUE-Mem reduces ambiguity through repeated clues,
automatic consistency checks, three-annotator review, and an Answer Refusal
task, but these controls validate benchmark support; they do not establish
that comparable inferences about real people would be fair.  The failure cases
above further show that a model can replace missing evidence with an
occupation stereotype or a plausible but competing preference.

\paragraph{Generated-media stewardship.}
Consistent portraits and voices are necessary for cross-session Entity Recall,
but the same assets could be detached from their benchmark context and used
to fabricate statements by a synthetic identity.  Any future release should
therefore retain provenance linking profiles, generated assets, and source
sound collections, together with the applicable redistribution conditions.
These considerations apply to the synthetic assets themselves; CUE-Mem does
not contain a mechanism for identifying real people from images or voices.

\section{Limitations}
\label{app:limitations}

\paragraph{Synthetic and controlled histories.}
The benchmark's controllability is also its principal limitation.  Its
profiles, dialogues, images, and speech may contain generation artifacts,
overly coherent personalities, or repeated visual and acoustic patterns that
differ from naturally occurring user histories.  Target preferences are
scheduled to recur over a one-year timeline and remain stable enough to
support a single gold memory.  Consequently, CUE-Mem does not evaluate
preference drift, contradictory disclosures, corrections, or forgetting over
longer interaction.  The recurrence analysis in
Appendix~D.4 describes this construction control and is not a causal ablation
of how repetition affects accuracy.

\paragraph{Coverage.}
CUE-Mem contains 20 synthetic users and Chinese multi-session conversations
covering the six profile domains used in construction.  Its multimodal scope
is text, shared images, and voice messages with environmental audio.  Implicit
targets are realized as peripheral visual objects or non-speech acoustic
events; unstated information inferred only from text is intentionally outside
the operational definition.  Results therefore should not be assumed to
transfer to other languages, cultural settings, cue distributions, or broader
media settings without additional evaluation.

\paragraph{Unequal cue difficulty.}
Visual implicitness is controlled through peripheral placement and
text--image similarity checks, whereas acoustic implicitness is controlled by
mixing an audible background event behind speech and excluding it from the
transcript.  These are modality-specific constructions, not perceptually
matched difficulty levels.  Cue size, occlusion, loudness, semantic rarity,
and interference vary within each modality, and the current benchmark does
not provide a calibrated human salience score for every clue.  Explicit--
implicit comparisons should therefore be interpreted at the aggregate
condition level rather than as paired counterfactual effects.

\paragraph{Evaluation scope and model dependence.}
All four tasks use four-way single-answer questions.  This enables consistent
scoring and controlled Answer Refusal lures, but does not measure how systems
form, revise, explain, or abstain from memories in open-ended conversation.
Performance also depends on the evaluated answer backbones, seven memory
systems, captioning granularities, audio textualizations, embedding model, and
retrieval cutoffs.  Detailed captions demonstrate one preservation--cost
tradeoff, while the RQ3 experiments test one text index and one multimodal
index; neither exhausts possible native multimodal memory designs.  Finally,
the repeated-run significance analysis uses a fixed five-profile sample, so
its confidence intervals quantify uncertainty for that audit rather than for
all possible users, models, and deployment environments.